\documentclass{article}

\usepackage{PRIMEarxiv}

\usepackage[utf8]{inputenc} 
\usepackage[T1]{fontenc}    
\usepackage{hyperref}       
\usepackage{url}            
\usepackage{booktabs}       
\usepackage{amsfonts}       
\usepackage{nicefrac}       
\usepackage{microtype}      
\usepackage{xcolor}         
\usepackage{amsmath,cleveref}
\usepackage{amssymb}
\usepackage{mathtools}
\usepackage{amsthm}
\usepackage[ruled,vlined]{algorithm2e}
\usepackage{subcaption}
\usepackage{graphics,graphicx}
\usepackage{fancyhdr}       

\usepackage[round]{natbib}

\newcommand\TT[1]{\boldsymbol{\mathcal{#1}}}                
\newcommand\RdTT{\mathbb{R}^{p_1 \times \dots \times p_d}}  
\newcommand\argmax[1]{\underset{#1}{\text{argmax }}}        
\newcommand\argmin[1]{\underset{#1}{\text{argmin }}}        

\newtheorem{proposition}{Proposition}

\fancypagestyle{firststyle}
{
   \fancyhf{}
   \renewcommand{\footruleskip}{2.2pt}
   \renewcommand{\footrulewidth}{0.6pt}
   \fancyfoot[L]{
    \footnotesize ${}^1$ Université Paris-Saclay, CNRS, CentraleSupélec, Laboratoire des signaux et systèmes, 91190, Gif-sur-Yvette, France. \\
    ${}^2$ Sorbonne Université, Institut du Cerveau - Paris Brain Institute - ICM, Inserm, CNRS, APHP, Hôpital Pitié Salpétrière Univ. Hosp., DMU Neuroscience 6, Paris, France. \\
    ${}^3$ Centre National de Recherche en Génomique Humaine, Institut François Jacob, CEA, Université Paris-Saclay, 91057, Évry, France. \\~\\
    Corresponding author: \texttt{arthur.tenenhaus@centralesupelec.fr}\\~\\
    Preprint. Under review.
   }
}

\newcommand\blfootnote[1]{%
  \begingroup
  \renewcommand\thefootnote{}\footnote{#1}%
  \addtocounter{footnote}{-1}%
  \endgroup
}

\title{Tensor Generalized Canonical Correlation Analysis}

\author{%
  Fabien Girka${}^{1, 2}$\\
  \And
  Arnaud Gloaguen${}^{3}$\\
  \AND
  Laurent Le Brusquet${}^{1}$\\
  \And
  Violetta Zujovic${}^{2}$\\
  \And
  Arthur Tenenhaus${}^{1, 2}$\\
}

\begin{document}

\maketitle

\begin{abstract}
Regularized Generalized Canonical Correlation Analysis (RGCCA) is a general statistical framework for multi-block data analysis. RGCCA enables deciphering relationships between several sets of variables and subsumes many well-known multivariate analysis methods as special cases. However, RGCCA only deals with vector-valued blocks, disregarding their possible higher-order structures. This paper presents Tensor GCCA (TGCCA), a new method for analyzing higher-order tensors with canonical vectors admitting an orthogonal rank-R CP decomposition. Moreover, two algorithms for TGCCA, based on whether a separable covariance structure is imposed or not, are presented along with convergence guarantees. The efficiency and usefulness of TGCCA are evaluated on simulated and real data and compared favorably to state-of-the-art approaches.
\end{abstract}

\begin{figure}[b]
    \noindent\makebox[\linewidth]{\rule{\columnwidth}{0.4pt}}
    \noindent\footnotesize
            \begin{tabular}{l}
                ${}^1$ Université Paris-Saclay, CNRS, CentraleSupélec, Laboratoire des signaux et systèmes, 91190, Gif-sur-Yvette, France. \\
                ${}^2$ Sorbonne Université, Institut du Cerveau - Paris Brain Institute - ICM, Inserm, CNRS, APHP, Hôpital Pitié Salpétrière \\
                Univ. Hosp., DMU Neuroscience 6, Paris, France. \\
                ${}^3$ Centre National de Recherche en Génomique Humaine, Institut François Jacob, CEA, Université Paris-Saclay, 91057, \\
                Évry, France. \\~\\
                Corresponding author: \texttt{arthur.tenenhaus@centralesupelec.fr}\\~\\
                Preprint. Under review.
            \end{tabular}
    \vspace{-50pt}
\end{figure}



\section{Introduction}
The study of a given phenomenon under multiple views can hopefully reveal a more significant part of the mechanisms at stake rather than considering each view separately.
In order to design a study under such a paradigm, measurements are usually acquired through different modalities resulting in multimodal/multi-view/multiblock/multi-source data.
One statistical framework suited explicitly for the joint analysis of such multi-source data is Regularized Generalized Canonical Correlation Analysis (RGCCA) \citep{Tenenhaus2011, Tenenhaus2017}. RGCCA is designed with a flexible yet simple algorithmic framework that encompasses a large number of well-known multi-block component methods. This includes Canonical Correlation Analysis \citep{HOTELLING1936}, Partial Least Squares Regression \citep{Wold1983}, many variants of correlation/covariance-based multiblock component methods (e.g., Carroll's GCCA \citep{Carroll1968}, MAXVAR \citep{Kettenring1971},  MAXBET \citep{Van1984}, Multiple Co-Inertia Analysis \citep{Chessel1996}, to cite a few). See \cite{Tenenhaus2017} for a detailed overview.

However, RGCCA can only treat vector-valued data (a.k.a first-order tensors) at the population level, whereas sometimes their natural structure is of higher order.
This is the case, for example, in electroencephalography (EEG) (time $\times$ channels and sometimes $\times$ frequencies), social network (users $\times$ channels $\times$ servers), process analysis (monitored variables $\times$ time $\times$ batches),
or text-mining (concepts $\times$ documents $\times$ languages) data. Taking into account such underlying tensor structure is not only a way to properly analyze the data but it can also improve the interpretation of the results, provide more robust estimators, associated with faster algorithms.

Tensor versions of vector-valued methods have been developed for a wide variety of problems \citep{acar2011allatonce, Zhou2013, Papalexakis2017}. These methods usually perform tensor analysis by imposing a tensor factorization model on the estimated vectors associated with each variable. Among the most known models are the CANDECOMP/PARAFAC (CP) decomposition \citep{Hars1970, Carroll1970} and the Tucker decomposition \citep{Tuck1963a, Tucker1964}.
\cite{TaeKyunKim2009, 10.5555/2540128.2540346, Gloaguen2020, Chen} have proposed extensions of CCA where the canonical vectors are constrained to follow a CP decomposition model  of rank-1. In \cite{Min}, the imposed CP decomposition is of rank-R.

Moreover, since the variables are supposed to bear a natural tensor structure, a central question for Tensor-based CCA methods is the resulting structure of the covariance matrices associated with each block. \cite{Min} addressed this question by proposing a separable structure to the block covariance matrices.

A first attempt to propose a tensor version of RGCCA was made in \cite{Gloaguen2020} with Multiway GCCA (MGCCA), but limited to matrix-valued data (a.k.a second-order tensor) at the population level, and where canonical vectors are modeled with a rank-1 CP decomposition, together with a separable structure for covariance matrices. In the line of \cite{Gloaguen2020}, we propose Tensor Generalized Canonical Correlation Analysis (TGCCA), a new tensor version of RGCCA, by enforcing an orthogonal rank-R CP decomposition to the canonical vectors and relaxing the separable assumption. TGCCA can (i) handle an arbitrary number of blocks, (ii) handle  tensor-valued data of any order, (iii) extract, from each block, canonical vectors modeled with an orthogonal rank-R CP decomposition, (iv) handle separable and non-separable covariance structure. To the extent of our knowledge, no tensor CCA method that gathers all these four properties has yet been proposed. Finally, algorithms designed to solve the TGCCA optimization problem are provided with theoretical convergence guarantees and experimental validation.

The remainder of the paper is organized as follows. In Section \ref{RGCCA-section}, we describe the RGCCA problem. Section \ref{pop_TGCCA}
presents the TGCCA optimization problem with and without the separable assumptions. The two strategies rely on the same master algorithm but with different core updates presented in Section \ref{Updates}. In Section \ref{Simulations}, we conduct numerical experiments to illustrate the benefits of the proposed methods compared to existing ones. We further evaluate our method on real data in Section \ref{Real-data}. Finally, we discuss the limitations of our approach and perspectives in Section \ref{Discussion}.

Our code is freely available on github\footnote{\url{https://github.com/GFabien/TGCCA-supplementary-material}}.

\section{RGCCA at the population level}
\label{RGCCA-section}
We consider $L$ random vectors $\mathbf{x}_1, \dots, \mathbf{x}_l, \dots, \mathbf{x}_L$. We assume that each random vector  $\mathbf{x}_l \in \mathbb{R}^{p_l}$ has a zero mean and a covariance matrix $\boldsymbol \Sigma_{ll}$. Let $\boldsymbol \Sigma_{lk} = \mathbb{E}[\mathbf x_l\mathbf x_k^\top]$ be the cross-covariance matrix between $\mathbf x_l$ and $\mathbf x_k$. Let $\mathbf w_l \in \mathbb{R}^{p_l}$ be the non-random canonical vector associated with the block $\mathbf{x}_l$. The objective of RGCCA is to find composite random variables $y_l =  \mathbf w_l^\top \mathbf{x}_l$ associated with each block that summarizes the relevant information between and within the blocks.
RGCCA at the population level is defined as the following optimization problem:
\begin{align}
    \label{RGCCA_optim}
    \underset{\mathbf w_1, \ldots, \mathbf w_L}{\text{maximize}} \sum_{l,k=1}^L c_{lk} \text{g}\left(\mathbf w_l^\top \mathbf \Sigma_{lk} \mathbf w_k\right) \quad \\
    \text{s.t.} \quad \mathbf w_l^\top \mathbf M_l \mathbf w_l = 1,  ~ l=1, \ldots, L. \nonumber
\end{align}
\begin{itemize}
\item The function $\text{g}$ is any continuously differentiable convex function. Its derivative is noted $\text{g}'$. If $c_{ll} \neq 0$ for some $l$ the constraint $\text{g}'(x)\ge 0$ for $x \geq 0$ must be added in order to guarantee the objective function to be multi-convex (i.e., convex with respect to each $\mathbf w_l$ while holding all others fixed).

\item The design matrix $\mathbf C = \lbrace c_{lk}\rbrace$ is a symmetric $L \times L$ matrix of non-negative elements describing the network of connections between blocks that the user wants to consider. Usually, $c_{lk} = 1$ between two connected blocks and $0$ otherwise.

\item Each block regularization matrix $\mathbf M_l \in \mathbb{R}^{p_l \times p_l}$ is symmetric positive-definite.
\end{itemize}

Many correlation and covariance-based component methods, including CCA, fall under this general formulation. See \cite{Tenenhaus2017} for a detailed overview.

For higher-order blocks, the RGCCA notations introduced above need to be extended. The following subsection reviews the notations and basic tensor operations needed for presenting TGCCA. We follow the terminology and notation introduced in \cite{KoBa09}.

\section{Population TGCCA}
\label{pop_TGCCA}

\subsection{Notations}
Scalars, vectors, matrices and higher order tensors are represented by $x$, $\mathbf x$, $\mathbf X$ and  $\TT{X}$, respectively. The shorthand $[n]$ will be used to denote the index set $\{1, \dots, n\}$. Let $\TT{X}$ be a tensor of order $d$, it means there exists $\{p_1, \dots, p_d \} \in \mathbb{N}^d$ such that $\TT{X} \in \RdTT$ i.e. $\TT{X}$ has $d$ modes, where the $m^\text{th}$ mode is of dimension $p_m$. Tensor elements can be described by $x_{i_1 \dots i_d}$ with $i_m \in [p_m]$ for $m \in [d]$.

Tensor fibers are the extension of matrix rows and columns: mode-$m$ fibers are vectors of $p_m$ elements obtained by fixing all indices except the $m^\text{th}$, leading to $\mathbf{x}_{i_1 \dots i_{m-1} . i_{m+1} \ldots i_d}$. Tensors can be matricized or unfolded along a given mode. The mode-$m$ matricization of tensor $\TT{X} \in \RdTT$ is denoted $\mathbf{X}_{(m)}$ and is of dimension $p_m \times \prod_{j \neq m} p_j$. This operation arranges the mode-$m$ fibers of $\TT{X}$ in a matrix. As matrices, tensors can be vectorized. For a matrix, $\mathbf{A} = \begin{bmatrix} \mathbf{a}_1 & \dots & \mathbf{a}_p \end{bmatrix}$, the vectorized version of $\mathbf{A}$ is $\mathbf{a} = \text{Vec}(\mathbf{A}) = \begin{bmatrix} \mathbf{a}_1^\top & \dots & \mathbf{a}_p^\top \end{bmatrix}^\top$ where $\top$ denotes the transpose operator. Therefore we can define the mode-$m$ vectorization of $\TT{X}$ by the vectorization of its mode-$m$ matricization.


We use the symbols $\circ$ for the outer product and $\otimes$ for the Kronecker product.

Finally, a tensor $\TT{X} \in \RdTT$ is said to be of rank one if there exists $d$ vectors $\mathbf{w}_1, \dots, \mathbf{w}_d \in \RdTT$ of unit norm and a scalar $\lambda$ such that $\TT{X} = \lambda ~ \mathbf{w}_1 \circ \dots \circ \mathbf{w}_d$. As for matrices, we can talk about rank-$R$ tensors if they cannot be expressed as the sums of less than $R$ rank-one tensors: $\TT{X} = \sum_{r = 1}^R \lambda^{(r)} \mathbf{w}_1^{(r)} \circ \dots \circ \mathbf{w}_d^{(r)}$. 
This decomposition, called CANDECOMP/PARAFAC (CP) \citep{Carroll1970, Hars1970} will be denoted $\mathbf x = [\![ \boldsymbol \lambda; \mathbf W_1, \dots, \mathbf W_d ]\!]$. It implicitly defines the quantities $\mathbf w^{(r)} = \mathbf{w}_d^{(r)} \otimes \dots \otimes \mathbf{w}_1^{(r)}$, $\mathbf W = \begin{bmatrix} \mathbf w^{(1)} & \dots & \mathbf w^{(R)}\end{bmatrix}$, and $\boldsymbol \lambda = \begin{bmatrix} \lambda^{(1)} & \dots & \lambda^{(R)}\end{bmatrix}^\top$, such that \\
\begin{equation*}
    \mathbf{x} = \sum_{r = 1}^R \lambda^{(r)} \mathbf{w}_d^{(r)} \otimes \dots \otimes \mathbf{w}_1^{(r)} = \sum_{r = 1}^R \lambda^{(r)} \mathbf{w}^{(r)} = \mathbf W \boldsymbol \lambda.
\end{equation*}

We talk about orthogonal rank when the factors $\mathbf w^{(r)}$ are orthogonal. If the factors $\mathbf w_m^{(r)}$ are orthogonal for every $m \in [d]$, we then talk about completely orthogonal rank \citep{Kolda2001}.

In the following sections, we consider $L$ random tensors $\TT{X}_1, \ldots, \TT{X}_l, \ldots, \TT{X}_L$. Each random tensor  $\TT{X}_l \in \mathbb{R}^{p_{l,1}\times \ldots \times p_{l,d_l}}$ is of order $d_l$ and the dimension of the $m^\text{th}$ mode of $\TT{X}_l$ is equal to $p_{l,m}$. We denote the mode-1 vectorization of  $\TT{X}_l$ by $\mathbf x_l$. We assume that the random vector $\mathbf x_l$ has a zero mean and a covariance matrix $\boldsymbol \Sigma_{ll}$. Let $\boldsymbol \Sigma_{lk} = \mathbb{E}[\mathbf x_l\mathbf x_k^\top]$ be the cross-covariance matrix between $\mathbf x_l$ and $\mathbf x_k$.
We note $\mathbf w_l$ an unknown non-random canonical vector of dimension $p_l = \prod_{m=1}^{d_l} p_{l,m}$.

\subsection{MGCCA optimization problem}
As the proposed work extends MGCCA, we first introduce its optimization problem. In the case of MGCCA, $\forall l \in [L], d_l = 2$. Hence, the following optimization problem given in Equation (2.2) of \cite{Gloaguen2020}:
\begin{align}
    \label{MGCCA_optim}
    &\underset{\mathbf w_1, \ldots, \mathbf w_L}{\text{maximize}} \sum_{l,k=1}^L c_{lk} \text{g}\left(\mathbf w_l^\top \mathbf \Sigma_{lk} \mathbf w_k\right) \quad \\
    &\text{s.t.} \quad \mathbf w_l^\top \mathbf M_l \mathbf w_l = 1, \text{ and } \mathbf w_l = \mathbf w_{l, 2} \otimes \mathbf w_{l, 1}, ~ l \in [L]. \nonumber
\end{align}
\cite{Gloaguen2020} make the additional assumption that the matrices $\mathbf M_l \in \mathbb{R}^{p_l \times p_l}$ can be written as the Kronecker product of two matrices $\mathbf M_{l, 1} \in \mathbb{R}^{p_{l, 1} \times p_{l, 1}}$ and $\mathbf M_{l, 2} \in \mathbb{R}^{p_{l, 2} \times p_{l, 2}}$: $\mathbf M_l = \mathbf M_{l, 2} \otimes \mathbf M_{l, 1}$. We qualify matrices with such a structure as separable matrices. Therefore, the change of variables $\mathbf v_l = \mathbf M_l^{\frac{1}{2}} \mathbf w_l$ leads to this new set of constraints for \eqref{MGCCA_optim} (Equation (2.6) in \cite{Gloaguen2020}):
\begin{equation}
    \label{MGCCA_constraints}
    \mathbf v_l^\top \mathbf v_l = 1 \text{ and } \mathbf v_l = \mathbf v_{l, 2} \otimes \mathbf v_{l, 1}, ~ l \in [L].
\end{equation}
Thus, MGCCA aims to maximize the criterion of RGCCA under the assumption that the canonical vectors admit a CP decomposition of rank 1 and that the regularization matrices $\mathbf M_l$ are separable.

\subsection{TCCA optimization problem}
Another close related work is TCCA from \cite{Min}. They maximize the CCA criterion: $ \mathbf w_1^\top \mathbf \Sigma_{12} \mathbf w_2$ under the constraints that $\mathbf w_l^\top \mathbf \Sigma_{ll} \mathbf w_l = 1$ and that the canonical vectors admit rank-$R_l$ CP decompositions:
\begin{align}
    \label{TCCA_optim}
    \mathbf w_l = [\![ \boldsymbol \lambda; \mathbf W_{l, 1}, \dots, \mathbf W_{l, d} ]\!], 
    \mathbf W_{l, m} \in &\mathbb{R}^{p_{l,m} \times R_l}, ~ l \in [2].
\end{align}
With such structures, canonical vectors can describe more complex interactions than those extracted with MGCCA while keeping a low degree of freedom compared to RGCCA.

\subsection{TGCCA optimization problem}
We now introduce the TGCCA optimization problem. Like MGCCA, we want to maximize the flexible criterion of RGCCA. Like TCCA, we want to consider rank-$R_l$ CP decompositions of $d_l^\text{th}$-order tensors.
Hence, a natural optimization problem that generalizes both MGCCA and TCCA consists in maximizing the criterion of \eqref{RGCCA_optim} under the following constraints:
\begin{align}
    \label{dream_TGCCA_optim}
    &\mathbf w_l^\top \mathbf M_l \mathbf w_l = 1, \text{ and } \mathbf w_l = [\![ \boldsymbol \lambda_l; \mathbf W_{l, 1}, \dots, \mathbf W_{l, d} ]\!], \\
    &\mathbf W_{l, m} \in \mathbb{R}^{p_{l,m} \times R_l}, ~ l \in [L]. \nonumber
\end{align}
To ensure that the solution of \eqref{dream_TGCCA_optim} does not degenerate in practice, we add orthogonality constraints between the rank-1 factors $\mathbf w_l^{(r)}$ of the CP decomposition. This additional constraint aims to prevent collinearity between factors. We can consider orthogonality constraints of the type $\mathbf w_l^{(r)\top} \mathbf K_l \mathbf w_l^{(s)} = \delta_{rs}$ where $\delta$ is the Kronecker delta, and $\mathbf K_l \in \mathbb{R}^{p_l \times p_l}$ is any symmetric positive-definite matrix.

\textbf{Separable TGCCA.} Under the assumption that matrices $\mathbf M_l$ are separable: $\mathbf M_l = \mathbf M_{ld} \otimes \ldots \otimes \mathbf M_{l1}$ with $\mathbf M_{lm} \in \mathbb{R}^{p_{lm} \times p_{lm}}$ for $m \in [d]$, we choose $\forall l \in [L]$, $\mathbf K_l = \mathbf M_l$. Therefore, applying the change of variables $\mathbf{v}_l = \mathbf{M}_l^\frac{1}{2} \mathbf{w}_l$ and $\mathbf{Q}_{lk} = \mathbf{M}_l^{-\frac{1}{2}} \boldsymbol{\Sigma}_{lk} \mathbf{M}_k^{-\frac{1}{2}}$ leads to the separable TGCCA optimization problem:
\begin{align}
    \label{separable_TGCCA_optim}
    &\underset{\mathbf v_1, \ldots, \mathbf v_L}{\text{maximize}} \sum_{l,k=1}^L c_{lk} \text{g}\left(\mathbf v_l^\top \mathbf Q_{lk} \mathbf v_k\right) \quad \text{s.t.} \quad \mathbf v_l^\top \mathbf v_l = 1,\\
    &\mathbf v_l = [\![ \boldsymbol \lambda_l; \mathbf V_{l, 1}, \dots, \mathbf V_{l, d} ]\!], ~ \mathbf V_{l, m} \in \mathbb{R}^{p_{l,m} \times R_l}, \nonumber \\
    &\text{and } \mathbf V_l^\top \mathbf V_l = \mathbf I_{R_l} ~ l \in [L]. \nonumber
\end{align}
As $\mathbf v_l = \mathbf V_l \boldsymbol \lambda_l$, the constraints of \eqref{separable_TGCCA_optim} imply that $\boldsymbol \lambda_l^\top \boldsymbol \lambda_l = 1$. Setting $R_l = 1$ and $d_l = 2$, we see that we get the MGCCA problem with the constraints described in \eqref{MGCCA_constraints}. 

\textbf{Non-separable TGCCA.} In the general case (i.e., no separability assumption on $\mathbf M_l$), the problem is hard to solve (see Appendix \ref{Appendix-nsTGCCA} for a detailed discussion), so we propose to study a relaxed version of \eqref{dream_TGCCA_optim} with orthogonality constraints. Indeed, we authorize $\mathbf w_l^\top \mathbf M_l \mathbf w_l \leq 1$ and impose $\mathbf W_l^\top \mathbf W_l = \mathbf I_{R_l}$. The constraints on $\boldsymbol \lambda_l$ and $\mathbf W_l$ are entangled since $\mathbf w_l^\top \mathbf M_l \mathbf w_l = \boldsymbol \lambda_l^\top \mathbf W_l^\top \mathbf M_l \mathbf W_l \boldsymbol \lambda_l$ and $\mathbf W_l^\top \mathbf M_l \mathbf W_l$ does not simplify for an arbitrary orthonormal matrix $\mathbf W_l$. To disentangle them, we can observe the following fact:
\begin{align*}
    \mathbf w_l^\top \mathbf M_l \mathbf w_l &\leq \Vert \mathbf w_l \Vert_2 \Vert \mathbf M_l \mathbf w_l \Vert_2, \\
    &\leq \Vert \mathbf M_l \Vert_2 \Vert \mathbf w_l \Vert_2^2, \\
    &\leq \Vert \mathbf M_l \Vert_2 \mathbf w_l^\top \mathbf w_l \\
    &= \Vert \mathbf M_l \Vert_2 \boldsymbol \lambda_l^\top \mathbf W_l^\top \mathbf W_l \boldsymbol \lambda_l = \Vert \mathbf M_l \Vert_2 \Vert \boldsymbol \lambda_l \Vert_2^2.
\end{align*}
The Cauchy-Schwarz inequality was used to get the second line and $\Vert \mathbf . \Vert_2$ is the matrix norm such that $\Vert \mathbf A \Vert_2 = \sup_{\mathbf u, \Vert \mathbf u \Vert_2 = 1} \Vert \mathbf A \mathbf u \Vert_2$.
Finally, imposing $\Vert \boldsymbol \lambda_l \Vert_2 \leq \Vert \mathbf M_l \Vert_2^{-\frac{1}{2}}$ makes sure that the relaxed constraint is satisfied, whatever the choice of $\mathbf W_l$. Consequently, we define the non-separable TGCCA optimization problem as:
\begin{align}
    \label{non-separable_TGCCA_optim}
    &\underset{\mathbf w_1, \ldots, \mathbf w_L}{\text{maximize}} \sum_{l,k=1}^L c_{lk} \text{g}\left(\mathbf w_l^\top \mathbf \Sigma_{lk} \mathbf w_k\right) \\
    &\text{s.t. } \mathbf w_l = [\![ \boldsymbol \lambda; \mathbf W_{l, 1}, \dots, \mathbf W_{l, d_l} ]\!], 
    \mathbf W_{l, m} \in \mathbb{R}^{p_{l,m} \times R_l}, \nonumber\\
    &\mathbf W_l^\top \mathbf W_l = \mathbf I_{R_l}, \text{ and } \Vert \boldsymbol \lambda_l \Vert_2 \leq \Vert \mathbf M_l \Vert_2^{-\frac{1}{2}} ~ l \in [L]. \nonumber
\end{align}
This last formulation hides the impact of the structure of $\mathbf M_l$ as it appears only through its matrix norm, which is its highest singular value. Therefore, we will take into account $\mathbf M_l$ in the optimization scheme instead (see calculations in Appendix \ref{Appendix-nsTGCCA}).

In a nutshell, both the separable and non-separable TGCCA optimization problems boil down to maximizing the criterion of RGCCA on compact sets $\Omega = \Omega_1 \times \dots \times \Omega_L$, where $\Omega_l$ is the constraint space of $\mathbf w_l$ or $\mathbf v_l$ (the proof that $\Omega$ is compact can be found in Appendix \ref{Appendix-Compactness}).
Both problems look for solutions that admit CP decompositions of orthogonal ranks $R_l$.

\subsection{An algorithm for maximizing a multi-convex continuously differentiable function}

Given a multi-convex continuously differentiable function $f(\mathbf w_1, \ldots,\mathbf w_L):\mathbb{R}^{p_1}\times \ldots \times \mathbb{R}^{p_L} \xrightarrow{}\mathbb{R}$ and compact sets $\Omega_l \subset \mathbb{R}^{p_l}, l \in [L]$, we are interested in the following general optimization problem:
\begin{equation}
\underset{\mathbf w_1, \ldots,\mathbf w_L}{\text{maximize}} ~ f(\mathbf w_1, \ldots,\mathbf w_L) \quad \text{s.t.} \quad \mathbf w_l \in \Omega_l, ~ l \in [L].
\label{optim_general}
\end{equation}

In \cite{Tenenhaus2017}, a master algorithm for maximizing a continuously differentiable multi-convex function, under the constraint that each $\mathbf{w}_l$ belongs to a compact set $\Omega_l$, is proposed.
As problems \eqref{separable_TGCCA_optim} and \eqref{non-separable_TGCCA_optim} fall under this configuration, this master algorithm can be used for TGCCA. This algorithm is based on a Block Coordinate Ascent (BCA) strategy \citep{deLeeuw1994},
which consists in updating sequentially only one canonical vector $\mathbf w_l \in \mathbb{R}^{p_l}$ (while keeping all the others fixed) in a way that increases the objective function. This is where the multi-convexity of the function $f$ comes in hand:
as a convex function lies above its linear approximation at $\mathbf{w}_l$, for any $\Tilde{\mathbf{w}}_l \in \Omega_l$, the following inequality holds:
\begin{align}
    \label{linear_approx}
    f(\mathbf{w}_1, \dots, &\mathbf{w}_{l - 1}, \Tilde{\mathbf{w}}_l, \mathbf{w}_{l + 1}, \dots, \mathbf{w}_L) \geq \\
    &f(\mathbf{w}) + \nabla_l f (\mathbf{w})^\top (\Tilde{\mathbf{w}}_l - \mathbf{w}_l) = \text{l}_l(\Tilde{\mathbf{w}}_l, \mathbf{w}), \nonumber
\end{align}
where $\mathbf{w} = (\mathbf{w}_1, \dots, \mathbf{w}_L)$ and $\nabla_l f (\mathbf{w})$ is the partial gradient of $f$ with respect to $\mathbf{w}_l$:
\begin{equation*}
    \nabla_l f (\mathbf{w}) = 2 \sum_{k = 1}^L c_{lk} \text{g}' (\mathbf{w}_l^\top \boldsymbol{\Sigma}_{lk} \mathbf{w}_k) \boldsymbol{\Sigma}_{lk} \mathbf{w}_k.
\end{equation*}
Maximizing $\text{l}_l(\Tilde{\mathbf{w}}_l, \mathbf{w})$ defined in \eqref{linear_approx} over $\Tilde{\mathbf{w}}_l$ is then equivalent to find
\begin{equation}
    \argmax{\Tilde{\mathbf{w}}_l \in \Omega_l} \nabla_l f (\mathbf{w})^\top \Tilde{\mathbf{w}}_l = \text{r}_l (\mathbf{w}).
    \label{update}
\end{equation}

On the one hand, from the definitions of $r_l(\mathbf w)$ and $\text{l}_l (\mathbf{w}_l, \mathbf{w})$ in respectively \eqref{update} and \eqref{linear_approx}, $ \text{l}_l (r_l(\mathbf{w}), \mathbf{w}) \geq \text{l}_l (\mathbf{w}_l, \mathbf{w}) = f(\mathbf{w})$. On the other hand, according to \eqref{linear_approx}, $f(\mathbf{w}_1, \dots, \mathbf{w}_{l - 1}, \text{r}_l (\mathbf{w}), \mathbf{w}_{l + 1}, \dots, \mathbf{w}_L) \geq \text{l}_l (\text{r}_l (\mathbf{w}), \mathbf{w})$. Thus, the proposed update \eqref{update} increases the value of the objective function. This fact remains true even if $\text{r}_l (\mathbf{w})$ is no longer the maximizer of $\nabla_l f (\mathbf{w})^\top \Tilde{\mathbf{w}}_l$, as long as $\nabla_l f (\mathbf{w})^\top \text{r}_l (\mathbf{w}) \geq \nabla_l f (\mathbf{w})^\top \mathbf{w}_l$.

\begin{algorithm}[ht]
    \SetAlgoLined
    \textbf{Result:} $\textbf{w}_1^s, \dots , \textbf{w}_L^s$ (approximate solution of \eqref{optim_general}) \\
    
    \textbf{Initialization:} $\textbf{w}_l^0 \in \Omega_l$, $l = 1, \dots, L$, $\varepsilon$; \\
    
    s = 0; \\
    
    \Repeat{$f(\mathbf{w}_1^{s + 1}, \dots, \mathbf{w}_L^{s + 1}) - f(\mathbf{w}_1^s, \dots, \mathbf{w}_L^s) < \varepsilon$}{
        \For{$l = 1$ \KwTo $L$}{
            $\mathbf{w}_l^{s + 1} = \text{r}_l(\mathbf{w}_1^{s + 1}, \dots, \mathbf{w}_{l-1}^{s + 1}, \mathbf{w}_l^s, \dots, \mathbf{w}_L^s)$
        }
        
        $s = s + 1$;
     }
    
    \caption{Master algorithm for optimization problem (\ref{optim_general}).}
    \label{general_algo}
    
\end{algorithm}

The master algorithm, without details on how to compute update \eqref{update}, can be found in Algorithm \ref{general_algo}.
For studying the convergence properties of Algorithm \ref{general_algo}, it is useful to introduce some additional notations. Let $c_l~:~\Omega\mapsto\Omega$ be an operator defined as $c_l(\mathbf{w}) = \left(\mathbf{w}_1, \ldots,  \mathbf{w}_{l-1}, \text{r}_l(\mathbf{w}), \mathbf{w}_{l+1}, \ldots,  \mathbf{w}_L\right)$ and $c~:~\Omega\mapsto\Omega$ be defined as $c = c_L\circ  \ldots \circ c_1$, where $\circ$ stands for the function composition.\\\\
We consider the sequence $\lbrace \mathbf{w}^s = \left(\mathbf{w}_1^{s}, \ldots, \mathbf{w}_L^{s} \right) \rbrace_{s=0}^\infty$ generated by Algorithm \ref{general_algo}. Using the operator $c$, the "for loop" inside Algorithm \ref{general_algo} can be replaced by the following recurrence relation:
\begin{equation}
\mathbf{w}^{s+1} = c(\mathbf{w}^{s}).
\label{c_stationary_eq}
\end{equation}

To study the convergence properties of Algorithm \ref{general_algo}, we will consider the infinite sequence $\left\lbrace \mathbf{w}^s \right\rbrace_{s=0}^\infty$ generated by (\ref{c_stationary_eq}). The convergence properties of Algorithm \ref{general_algo} are summarized in the next proposition.
\begin{proposition}
Let $\left\lbrace \mathbf{w}^s\right\rbrace_{s=0}^{\infty}$ be any sequence generated by the recurrence relation $\mathbf{w}^{s+1} = c(\mathbf{w}^s)$ with $\mathbf{w}^0 \in \Omega$. Then, the following properties hold:
\begin{itemize}
    \item The sequence $\{f (\mathbf{w}^s )\}_{s=0}^\infty$ is monotonically increasing and therefore convergent as $f$ is bounded on $\Omega$. This result implies the monotonic convergence of Algorithm \ref{general_algo}.
    \item If the infinite sequence $\{f (\mathbf{w}^s )\}_{s=0}^\infty$ involves a finite number of distinct terms, then the last distinct point satisfies $c(\mathbf{w}^s) = \mathbf{w}^s$ and therefore is a stationary point of problem \eqref{optim_general}.
    \item The limit of any convergent subsequence of $\{ \mathbf{w}^s \}_{s=0}^\infty$ is a fixed point of $c$.
    \item $\lim_{s \rightarrow \infty} f(\mathbf{w}^s ) = f (\mathbf{w}^\star )$, where $\mathbf{w}^\star$ is a fixed point of $c$.
    \item The sequence $\{ \mathbf{w}^s \}_{s=0}^\infty$ is asymptotically regular: $\lim_{s \rightarrow \infty} \sum_{l=1}^L \|\mathbf{w}_l^{s + 1} - \mathbf{w}_l^s\| = 0$. This result implies that if the threshold $\epsilon$ in Algorithm \ref{general_algo} is made sufficiently small, the output of Algorithm \ref{general_algo} will be as close as wanted to a stationary point of \eqref{optim_general}.
    \item If the equation $\mathbf{w} = c(\mathbf{w})$ has a finite number of solutions, then the sequence $\{\mathbf{w}^s \}_{s=0}^\infty$ converges to one of them.
\end{itemize}
\end{proposition}
The TGCCA algorithm inherits from the convergence properties of the algorithm from \cite{Tenenhaus2017} as long as the solution found for \eqref{update} exists and is unique.

\section{Updates for TGCCA}
\label{Updates}
Let assume that we have $n$ observations of the $L$ tensor blocks:  $(\TT{X}_{1, i}, \dots, \TT{X}_{L, i})_{i \in [n]}$. Our goal is to estimate $\mathbf{w}_1, \dots, \mathbf{w}_L$ solution of \eqref{separable_TGCCA_optim} and \eqref{non-separable_TGCCA_optim} where $\boldsymbol{\Sigma}_{lk}$ and $\mathbf{M}_l$ are replaced with their estimates, respectively $\hat{\boldsymbol{\Sigma}}_{lk}$ and $\hat{\mathbf{M}}_l$. In subsection \ref{sample_sep_TGCCA}, we focus on the separable case and the special case of matrix blocks is detailed in subsection \ref{update_Matrix}. The non-separable case is presented in Appendix \ref{Appendix-nsTGCCA}. The main difference with the separable case lies in the update of $\boldsymbol \lambda_l$.

\subsection{Update for sample separable TGCCA}
\label{sample_sep_TGCCA}
In case regularization matrices have a separable structure, our goal is to find a solution of \eqref{update} where $\mathbf{Q}_{lk}$ is replaced with its estimated counterpart. In practice, $\hat{\mathbf{M}}_l$ and $\hat{\boldsymbol{\Sigma}}_{lk}$ are estimated and 
$\Hat{\mathbf Q}_{lk} = \hat{\mathbf{M}}_l^{-\frac{1}{2}} \hat{\boldsymbol{\Sigma}}_{lk} \hat{\mathbf{M}}_k^{-\frac{1}{2}}$.

We can note that $\mathbf v_l^{(r)}$ is a linear function of $\mathbf v_{l,m}^{(r)}$. Indeed, $\mathbf{v}_{l}^{(r)} = \mathbf{V}_{l, (-m)}^{(r)} \mathbf{v}_{l,m}^{(r)}$, where $\mathbf{V}_{l, (-m)}^{(r)} = \left( \mathbf{v}_{l,d_l}^{(r)} \otimes \dots \otimes \mathbf{v}_{l, m+1}^{(r)} \otimes \mathbf{I}_{p_{l,m}} \otimes \mathbf{v}_{l, m-1}^{(r)} \otimes \dots \otimes \mathbf{v}_{l,1}^{(r)} \right) \in \mathbb{R}^{p_l \times p_{l,m}}$.
We can also observe that
\begin{align}
    \label{sep_observation}
    &\nabla_l f(\mathbf{v})^\top \mathbf{v}_l = \sum_{r = 1}^R \lambda_l^{(r)} \nabla_l f(\mathbf{v})^\top \mathbf{V}_{l, (-m)}^{(r)} \mathbf{v}_{l,m}^{(r)} \nonumber \\
    &\quad \quad \quad \quad ~ ~ ~ = \text{Tr}(\mathbf{F}^\top \mathbf{V}_{l,m}),  \\
    &\text{with} \quad \mathbf{F} = \begin{bmatrix} \mathbf{f}^{(1)} \dots \mathbf{f}^{(R_l)} \end{bmatrix} \nonumber \\ 
    &\text{and} \quad \mathbf{f}^{(r)} = \lambda_l^{(r)} \mathbf{V}_{l,(-m)}^{(r) \top} \nabla_l f(\mathbf{v}) \in \mathbb{R}^{p_{l,m}}. \nonumber
\end{align}
The interest in formulation \eqref{sep_observation} is the separation of an entire mode, embodied by $\mathbf V_{l,m}$, from all the others.
As a consequence, we propose to use BCA here to alternate between $\boldsymbol{\lambda}_l$ and $\mathbf{V}_{l,m}$ for $m \in [d_l]$. Hence we define $\Omega_{l,m}^v = \{ \mathbf{V}_{l,m} \in \mathbb{R}^{p_{l,m} \times R_l}; \quad \mathbf{V}_{l,m}^\top \mathbf{V}_{l,m} = \mathbf{I}_{R_l} \}$ the feasible set for $\mathbf{V}_{l,m}$ and the intermediate update
\begin{equation}
   \text{r}_{l,m}(\mathbf{v}) = \argmax{\mathbf{V}_{l,m} \in \Omega_{l,m}^v} \text{Tr} (\mathbf{F}^\top \mathbf{V}_{l,m}) = \mathbf{S} \mathbf{T}^\top,
   \label{r_m}
\end{equation}
where $\mathbf{S}$ and $\mathbf{T}$ are respectively the left and right singular vectors of the rank-$R_l$ Singular Value Decomposition (SVD) of $\mathbf{F}$ (orthogonal Procrustes problem, \citep{Everson97orthogonal}). It is worth noting here that the update for mode $m$ is unique if $\mathbf{F} \in \mathbb{R}^{p_{l,m} \times R_l}$ has full rank.

By choosing this update for every mode $m$, we implicitly imposed a completely orthogonal rank to the vector $\mathbf v_l$. In fact, imposing orthogonality constraints on one mode is enough. In this case, this update should be chosen for a given mode $m$, and the update for every other mode $q$ would directly be $\text{r}_{l,q}(\mathbf{v}) = \begin{bmatrix}\frac{\mathbf{f}^{(1)}}{\Vert \mathbf{f}^{(1)} \Vert_2} \dots \frac{\mathbf{f}^{(R_l)}}{\Vert \mathbf{f}^{(R_l)} \Vert_2} \end{bmatrix}$.

From a tensor point of view, $\mathbf f^{(r)} / \lambda_l^{(r)}$ can be seen as the result of the mode products between the folded version of $\nabla_l f(\mathbf v)$: $\TT{F} \in \mathbb{R}^{p_{l,1} \times \dots \times p_{l, d_l}}$ and all the $\mathbf v_{l, q}^{(r)}$ for $q \in [d_l] \backslash \{m\}$. Therefore, the matrices $\mathbf V_{l, (-m)}^{(r)}$ do not need to be computed in practice, limiting the algorithm's complexity.

From Cauchy-Schwartz, the update on $\boldsymbol \lambda_l$ is:
\begin{equation}
    \text{r}_{l,\lambda}(\mathbf{v}) = \argmax{\boldsymbol \lambda_l, \Vert \boldsymbol \lambda_l \Vert_2 = 1} \nabla_l f(\mathbf{v})^\top \mathbf V_l \boldsymbol \lambda_l = \frac{\mathbf{V}_l^\top \nabla_l f(\mathbf{v})}{\| \mathbf{V}_l^\top \nabla_l f(\mathbf{v}) \|_2}
    \label{r_lambda}.
\end{equation}

Updating each $\mathbf V_{l,m}$ using \eqref{r_m} and then updating $\boldsymbol \lambda_l$ using \eqref{r_lambda} yields a unique update for \eqref{update} that increases the value of the objective function.

\subsection{Special case of matrix blocks}
\label{update_Matrix}
When $d_l = 2$, i.e. block $l$ has intrinsically a matrix structure, we can make an observation similar to \eqref{sep_observation}. Indeed, let $\mathbf{F} \in \mathbb{R}^{p_{l,1} \times p_{l,2}}$ be a reshaped version of $\nabla_l f(\mathbf{v})$ and $\boldsymbol{\Lambda}_l = \text{diag}(\boldsymbol{\lambda}_l)$,
\begin{equation*}
    \sum_{r = 1}^{R_l} \lambda_l^{(r)} \nabla_l f(\mathbf{v}) ^\top (\mathbf{v}_{l,2}^{(r)} \otimes \mathbf{v}_{l,1}^{(r)}) = \text{Tr}(\boldsymbol{\Lambda}_l \mathbf{V}_{l,1}^\top \mathbf{F} \mathbf{V}_{l,2}).
\end{equation*}
A closed form solution of $\argmax{\boldsymbol{\Lambda}_l, \mathbf{V}_{l,1}, \mathbf{V}_{l,2}} \text{Tr}(\boldsymbol{\Lambda}_l \mathbf{V}_{l,1}^\top \mathbf{F} \mathbf{V}_{l,2})$ cannot be found but applying one iteration of the tandem algorithm from \cite{Everson97orthogonal} increases the criterion and gives
\begin{equation*}
    \text{r}_{l,1}(\mathbf{v}) = \mathbf S, \quad \text{r}_{l,2}(\mathbf{v}) = \mathbf T \quad \text{and} \quad \text{r}_{l, \lambda}(\mathbf{v}) = \frac{\boldsymbol \delta}{\Vert \boldsymbol \delta \Vert_2},
\end{equation*}
where $\mathbf{S} ~ \text{diag}(\boldsymbol{\delta}) ~ \mathbf{T}^\top$ is the rank-$R_l$ SVD of $\mathbf{F}$.
If the $R_l^\text{th}$ singular value of $\mathbf{F}$ is not degenerated, then the proposed update is unique. This is equivalent to alternate between $(\mathbf{V}_{l, 1}, \mathbf{V}_{l, 2})$ and $\boldsymbol{\lambda}_l$. If $R_l = 1$, we retrieve the update of MGCCA (Equation (2.10) in \cite{Gloaguen2020}). 

\section{Numerical experiments}
\label{Simulations}

\subsection{Methods}
To evaluate the quality of the estimates provided by TGCCA, we generate blocks using the data model presented in Appendix \ref{simulation-appendix}, based on the probabilistic TCCA model of \cite{Min}.
Our simulations aim to assess the ability of TGCCA and state-of-the-art approaches to recover the canonical vectors used to generate the data. The cosine between the true canonical vectors and the estimated ones is used as an indicator of quality: $ \alpha_l = \frac{|\mathbf w_l^\top \Hat{\mathbf w}_l| }{\Vert\mathbf w_l\Vert_2 \Vert \Hat{\mathbf w}_l\Vert_2} $. In our comparisons, we include MGCCA \citep{Gloaguen2020}, Tensor CCA (TCCA) \citep{Min}, and two-dimensional CCA (2DCCA) \citep{Chen}. 
Finally, RGCCA and the per block SVD are considered as baselines. If appropriate, the method's rank is added as a suffix and the separable assumption as a prefix with ”sp”. As shown in Section \ref{update_Matrix}, spTGCCA1 is equivalent to MGCCA, so only MGCCA will be reported in the tables.


\subsection{Results}

\begin{table}
  \caption{Cosine between the true and the estimated canonical vectors as well as computation times. Median and quantiles (2.5\% and 97.5\%) are reported.}
    \label{tab:results}
  \centering
  \resizebox{0.65\columnwidth}{!}{%
\begin{tabular}{llll} 
\toprule 
Model & Gas & Cross (small) & Computation time \\ 
\midrule 
2DCCA1 & 0.30 (0.01, 0.89) & 0.43 (0.16, 0.85) & 3.09 (2.76, 4.50) \\ 
TCCA1 & 0.89 (0.22, 0.90) & 0.85 (0.32, 0.86) & 7.72 (7.38, 9.17) \\ 
TGCCA1 & 0.89 (0.87, 0.90) & 0.85 (0.83, 0.86) & 8.70 (8.50, 10.26) \\ 
spTCCA1 & 0.89 (0.22, 0.90) & 0.85 (0.32, 0.86) & 7.43 (7.24, 8.04) \\ 
MGCCA & 0.89 (0.87, 0.90) & 0.86 (0.83, 0.86) & 4.98 (4.66, 5.16) \\ 
\\
2DCCA3 & 0.04 (0.01, 0.21) & 0.13 (0.05, 0.31) & 1.50 (1.41, 3.09) \\ 
TCCA3 & 0.89 (0.87, 0.90) & 0.85 (0.83, 0.86) & 7.93 (7.73, 8.92) \\ 
TGCCA3 & 0.91 (0.78, 0.94) & 0.92 (0.79, 0.96) & 10.87 (10.11, 13.93) \\ 
spTCCA3 & 0.89 (0.87, 0.90) & 0.85 (0.83, 0.86) & 7.32 (7.22, 7.51) \\ 
spTGCCA3 & 0.92 (0.82, 0.94) & 0.93 (0.83, 0.96) & 5.62 (5.44, 6.31) \\ 
\\
RGCCA & 0.17 (0.05, 0.26) & 0.11 (0.06, 0.20) & 13.12 (12.67, 14.07) \\ 
SVD & 0.00 (0.00, 0.01) & 0.01 (0.00, 0.03) & 5.78 (5.44, 6.07) \\ 
\bottomrule 
\end{tabular}
}%
\end{table}

Table \ref{tab:results} reports the results of the compared methods.
The median and the quantiles (2.5\% and 97.5\%) on the different 
data folds of the cosines and computation times (in seconds) are reported.
TCCA and TGCCA are run 5 times per fold, and the model with the best criterion is kept each time while 2DCCA is run only once using a so-called "effective" strategy for the choice of the starting point \citep{Chen}.

Firstly, it appears that all algorithms solving the rank-1 TCCA problem behave similarly. Only TGCCA takes advantage of the use of higher rank.
Both TGCCA and TCCA seek canonical vectors with the same underlying rank-$R$ CP structure. The main difference between the two models relies on the fact that TCCA does not impose orthogonal rank-1 factors. In our experiments, TCCA3 extracted canonical vectors
$\{ \hat{\mathbf w}_l^{(r)} \}_{r \in [3]}$
that are almost collinear. This degenerate solution of TCCA3 justifies the orthogonality constraints in TGCCA.

2DCCA aims to extract $R$ rank-$1$ canonical vectors with orthogonal canonical components $\mathbf X_{l,(1)} \mathbf w_l$. A deflation strategy is used to ensure orthogonality between components.
As 2DCCA3 is designed to find 3 distinct canonical components of rank 1 with the same weights, the reconstructed vectors $\hat{\mathbf w}_l = \sum_{r=1}^{R_l} \hat{\mathbf w}_{l, d_l}^{(r)} \otimes \dots \otimes \hat{\mathbf w}_{l, 1}^{(r)}$ do not correspond to the true canonical vectors.

Looking at the tables in Appendix \ref{simulation-appendix}, it is worth noticing that RGCCA and SVD need a higher signal-to-noise ratio (SNR) to perform equally well as TGCCA3 and spTGCCA3. 


We perform an additional experiment with 3D canonical vectors. The conclusions are similar, except for RGCCA, which performs much better due to changes in the noise generation process. (see Table \ref{tab:results_4D} and Appendix \ref{Appendix-4D}). 2DCCA was excluded because we did not implement a higher-order version algorithm. 
\begin{table}
  \caption{Cosine between the true and the estimated canonical vectors as well as computation times. Median and quantiles (2.5\% and 97.5\%) are reported (3D settings).}
    \label{tab:results_4D}
  \centering
  \resizebox{0.7\columnwidth}{!}{%
\begin{tabular}{llll} 
\toprule 
Model & Cross & Cross (small) 3D & Computation time \\ 
\midrule 
TCCA1 & 0.88 (0.87, 0.89) & 0.88 (0.85, 0.88) & 17.31 (16.74, 18.94) \\ 
TGCCA1 & 0.88 (0.87, 0.89) & 0.88 (0.85, 0.88) & 14.29 (13.24, 21.84) \\ 
spTCCA1 & 0.88 (0.87, 0.89) & 0.88 (0.85, 0.88) & 15.70 (15.32, 23.05) \\ 
spTGCCA1 & 0.88 (0.87, 0.89) & 0.87 (0.87, 0.88) & 15.90 (15.10, 24.15) \\ 
\\
TCCA3 & 0.88 (0.87, 0.89) & 0.88 (0.85, 0.88) & 23.89 (19.64, 26.21) \\ 
TGCCA3 & 0.98 (0.96, 1.00) & 0.88 (0.82, 0.88) & 21.58 (20.46, 25.45) \\ 
spTCCA3 & 0.88 (0.87, 0.89) & 0.88 (0.85, 0.88) & 17.96 (15.65, 20.41) \\ 
spTGCCA3 & 0.98 (0.96, 0.99) & 0.99 (0.95, 0.99) & 49.52 (27.35, 113.45) \\ 
\\
RGCCA & 0.97 (0.92, 0.98) & 0.98 (0.84, 0.98) & 115.32 (113.21, 116.23) \\ 
SVD & 0.01 (0.00, 0.03) & 0.01 (0.00, 0.03) & 1.62 (1.54, 1.73) \\ 
\bottomrule 
\end{tabular}
}%
\end{table}
\vspace{-2mm}

\section{Evaluation on real data}
\label{Real-data}

\begin{table*}[t]
\caption{Cosine between the true and the estimated concentrations with computation times (mean and standard deviation). The cosine for the best model over the 100 runs is reported between parentheses.}
    \label{tab:acar}
  \centering
  \resizebox{.8\columnwidth}{!}{%
    \begin{tabular}{llll} 
        \toprule 
        Chemical & TGCCA & CMTF & ACMTF \\ 
        \midrule 
        Val-Tyr-Val & 0.961 $\pm$ 0.006 (0.962) & 0.625 $\pm$ 0.228 (0.999) & 0.608 $\pm$ 0.204 (0.949) \\ 
        Trp-Gly & 0.907 $\pm$ 0.026 (0.920) & 0.505 $\pm$ 0.260 (0.978) & 0.538 $\pm$ 0.225 (0.947) \\ 
        Phe & 0.834 $\pm$ 0.096 (0.870) & 0.556 $\pm$ 0.341 (0.981) & 0.637 $\pm$ 0.300 (0.703) \\ 
        Malto & 0.998 $\pm$ 0.000 (0.998) & 0.995 $\pm$ 0.002 (0.992) & 0.995 $\pm$ 0.002 (0.991) \\ 
        Propanol & 0.998 $\pm$ 0.000 (0.998) & 0.533 $\pm$ 0.319 (0.997) & 0.518 $\pm$ 0.295 (0.991) \\ \\
        Computation time & 3.926 $\pm$ 0.209 & 93.723 $\pm$ 43.599 & 111.782 $\pm$ 45.052 \\
        \bottomrule
    \end{tabular}
  }%
\end{table*}
\subsection{Chemical mixtures dataset}
In this section, we evaluate the ability of TGCCA to retrieve the signatures/patterns related to different chemicals in given mixtures from the dataset publicly available at \href{http://www.models.life.ku.dk/joda/prototype}{http://www.models.life.ku.dk/joda/prototype} \citep{Acar2014}. In this dataset, a set of 28 mixtures with known chemical composition was measured using different analytical techniques, i.e., NMR (Nuclear Magnetic Resonance) spectroscopy and LC-MS (Liquid Chromatography-Mass Spectrometry), resulting in two blocks of data. The first one is a third-order tensor of dimensions $28 \times 13324 \times 8$, and the second one is a matrix of dimensions $28 \times 168$. In this dataset, 4 of the chemicals can be identified in both blocks, while the last one does not appear in the matrix block \citep{Acar2014}. We choose to extract 1 component of rank 2 and 3 components of rank 1 using the same deflation strategy for orthogonal components as in \cite{Gloaguen2020}. We use $g : x \rightarrow x^2$ and choose to take identity matrices for $\mathbf M_l$, so we use spTGCCA in this Section.

We compare our approach with CMTF and Advanced CMTF (ACMTF) methods \citep{acar2011allatonce, Acar2014}. CMTF aims at solving the following optimization problem:
\begin{equation}
    \label{CMTF}
    \argmin{\mathbf A, \mathbf B, \mathbf C, \mathbf V, \boldsymbol \lambda, \boldsymbol \sigma} \Vert \TT{X} - [\![\boldsymbol \lambda; \mathbf A, \mathbf B, \mathbf C]\!] \Vert_F^2 + \Vert \mathbf Y - \mathbf A \Sigma \mathbf V^\top \Vert_F^2
\end{equation}
with $\boldsymbol \Sigma = \text{diag}(\boldsymbol \sigma)$, $\mathbf A$, $\mathbf B$, $\mathbf C$ and $\mathbf V$ being matrices with $R$ columns.
Hence, CMTF looks for the best rank-$R$ decompositions of both the tensor and matrix blocks, with the $\mathbf A$ matrix being shared between both blocks. ACMTF allows both shared and unshared components between blocks by adding penalties to enforce sparsity on the coefficients $\boldsymbol \lambda$ and $\boldsymbol \sigma$. Doing so, the model becomes able to select different columns of the matrix $\mathbf A$ in each block. This is particularly interesting with this dataset since one of the chemicals does not appear in the matrix block. Implementations of CMTF and ACMTF were used from the \href{http://www.models.life.ku.dk/~acare/CMTF_Toolbox.html}{MATLAB CMTF Toolbox v1.1, 2014}.

As in \cite{Acar2014}, blocks are divided by their Frobenius norms. An additional centering step is performed before applying TGCCA. We run 100 times each model with random initial points and report the cosines (mean and standard deviation) between the true concentrations of the chemicals in the mixture and the estimated ones in Table \ref{tab:acar}. The computation times (in seconds) are also reported.

It is known that the (A)CMTF algorithm is sensitive to initialization.
On the opposite, TGCCA seems really stable in this experiment while being much faster than CMTF. We can also observe that, for TGCCA, the cosines are very high and the standard deviations very low for the first two extracted chemicals (Malto and Propanol). In contrast, the cosines decrease, and the standard deviations increase for the next ones.
This is expected because the deflation strategy imposes that the vectors of concentrations of the different chemicals are orthogonal, which is not the case for this dataset.
Hence, the next components cannot be recovered perfectly.
As this limitation does not apply to CMTF and ACMTF, it is not surprising that the best model over the 100 runs for CMTF (reported in parenthesis in Table \ref{tab:acar}) reaches higher cosines than the best model for TGCCA for 3 of the 5 chemicals.

\subsection{The Multi-PIE Face dataset}
\begin{figure*}
    \centering
    \includegraphics[width=0.95\linewidth]{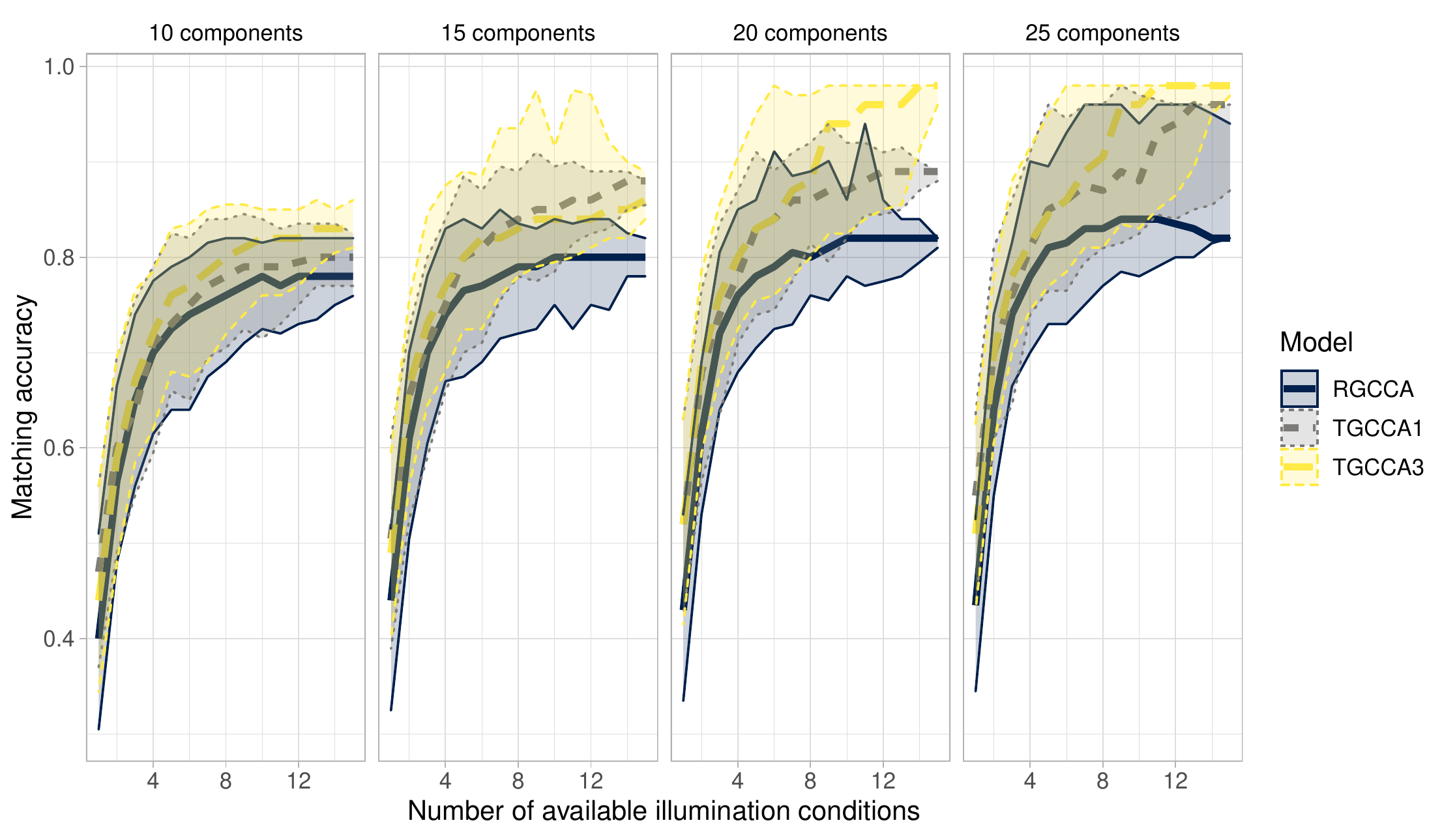}
    \caption{Matching accuracy on the test set for different models and different number of components. Experiments are repeated 100 times. Ribbons contain 95\% of the points around the median for each model.}
    \label{fig:matching_accuracy}
\end{figure*}

To further evaluate our model, we perform an analysis similar to the one performed in \cite{10.5555/2540128.2540346} using faces from the Multi-PIE Face dataset \citep{Multi-PIE}. We select images (of dimensions $64 \times 64$) from 100 subjects in two different views and 15 different illumination conditions and organize them in two $4^\text{th}$-order tensor blocks of dimensions $100 \times 64 \times 64 \times 15$ corresponding to the two views. The goal is to learn a common latent subspace between the two blocks and then use this latent representation to match subjects across the two views. We use 100 new subjects to evaluate the matching performances. Each new subject comes with images in the two views, and we vary the number of illumination conditions from 1 to 15. Therefore, we want to pair tensors of dimensions $64 \times 64 \times i$ where $i \in [15]$. Since the latent subspace is learned using all illumination conditions, if $i < 15$, there are missing slices in the tensors. This problem can be overcome by imputing the missing slices to the means of the training subjects (see Appendix \ref{Appendix-matching faces} for a justification).
The illumination conditions are randomly sampled, so they may differ across views for a given subject. 
We suppose that the illumination condition for each image is unknown. Therefore, we train a Linear Discriminant Analysis (LDA) classifier on the $100 \times 15 \times 2 = 3000$ available images of the training set to predict the illumination condition of each image. These images were downsampled and vectorized to the dimension $16 \times 16 = 256$ to reduce the number of variables while keeping enough information to perform the classification task.

Once the illumination conditions have been predicted and the tensors completed with missing slices, they are projected on the learned subspace. 
Subjects are then paired across the views by solving Integer Linear Programming to find the assignment that maximizes the sum of cosines between paired projections. Finally, the metric of interest is the accuracy of the matching.

We compare here RGCCA and TGCCA of ranks 1 and 3. As they are $64 \times 64 \times 15 = 61440$ variables for 100 subjects, we do not try to estimate the covariance matrices but instead, use the identity matrices as the regularization matrices $\mathbf M_l$. Results based on the number of available illumination conditions per subject are shown in Figure \ref{fig:matching_accuracy}. As illumination conditions are randomly sampled, we repeat the experiments 100 times. 
Rank-3 TGCCA seems to perform slightly better than rank-1 TGCCA, itself performing slightly better than RGCCA, and the matching accuracy increases with the number of available illumination conditions.
Furthermore, it is important to note that RGCCA must estimate 61440 parameters while rank-1 TGCCA estimates $(64 + 64 + 15) \times 1 = 143$ parameters and rank-3 TGCCA estimates $(64 + 64 + 15) \times 3 = 429$ parameters per component and block. Figure \ref{fig:canonical_vectors} shows the obtained canonical vectors with the three methods. While rank-1 and rank-3 TGCCA focus on capturing what varies between subjects, RGCCA produces canonical vectors that are much closer to actual faces. Therefore, RGCCA is probably more prone to overfitting than the TGCCA models, which may explain their better results here. 
Figure \ref{fig:matching_accuracy} also shows that by increasing the number of components, the matching accuracy increases more for TGCCA models than RGCCA.

\section{Conclusion and Discussion}
\label{Discussion}
We have proposed Tensor Generalized Canonical Correlation Analysis as a general framework for analyzing several higher-order tensors and matrices jointly. TGCCA relies on the RGCCA framework on which we imposed that an orthogonal rank-R CP decomposition models the canonical vectors.
Both separable and non-separable TGCCA give promising results, while separable TGCCA seems faster.
Convergence of our algorithms is guaranteed, up to uniqueness conditions that can be monitored at run time.

If orthogonality constraints allow interesting results,
they may not coincide with the true underlying structure of the
data. A solution could be to impose angle constraints like in \cite{CMTF-angle-constraints}.

We now introduce some points for further investigations.
Without noise, overestimating the rank would lead to have extra factors with zero weights. On real data, both the noise and the relevant information would be estimated by additional factors. Thus, a procedure to find the best rank is of great interest and under investigation.


TGCCA extracts the canonical vector for each block sequentially. Extracting the next canonical vectors can be done using a deflation strategy (see \cite{Gloaguen2020} for details). This approach imposes the canonical components to be orthogonal, which is not the case in the real dataset we studied. A procedure that extracts all sets of canonical vectors simultaneously is currently under investigation.



To fully benefit from the $L \geq 2$ setting, an interesting research line would be defining higher-order correlations \citep{7123622, cor_theory, wang2020measures} and incorporating them in the criterion to optimize.

\section*{Acknowledgments}
This work is supported by a public grant overseen by the French National Research Agency (ANR) through the program UDOPIA, project funded by the ANR-20-THIA-0013-01. We would also like to thank Eun Jeong Min and Hua Zhou for sharing their implementation of TCCA with us.

\bibliography{egbib.bib}

\newpage
\appendix
\onecolumn
\section{Non-separable TGCCA}
\label{Appendix-nsTGCCA}
We detail in this Section the choice of the formulation of the non-separable TGCCA optimization problem \eqref{non-separable_TGCCA_optim} and how we propose to tackle this problem. 

\subsection{Formulation of non-separable TGCCA}
As stated in Section \ref{pop_TGCCA}, we seek to solve 
\begin{align}
    \label{dream_objective}
    \underset{\mathbf w_1, \ldots, \mathbf w_L}{\text{maximize}} &\sum_{l,k=1}^L c_{lk} \text{g}\left(\mathbf w_l^\top \mathbf \Sigma_{lk} \mathbf w_k\right) \\
    \text{s.t. } \mathbf w_l^\top \mathbf M_l \mathbf w_l = 1, \mathbf w_l = [\![ \boldsymbol \lambda; \mathbf W_{l, 1}, \dots, &\mathbf W_{l, d} ]\!], ~ \mathbf W_{l, m} \in \mathbb{R}^{p_{l,m} \times R_l}, \text{ and } \mathbf W_l^\top \mathbf K_l \mathbf W_l = \mathbf I_{R_l} ~ l \in [L]. \nonumber
\end{align}
The two natural choices for $\mathbf K_l$ are $\mathbf K_l = \mathbf M_l$ and $\mathbf K_l = \mathbf I_{p_l}$.

\begin{figure}
\centering
\begin{subfigure}{0.3\columnwidth}
  \centering
  \includegraphics[width=1\linewidth]{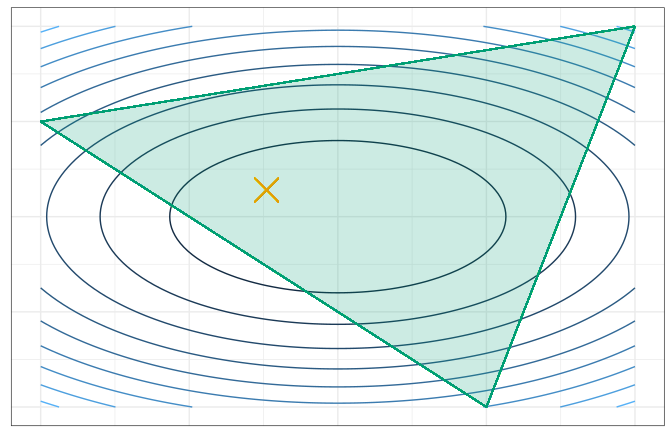}
  \caption{Initial point}
\end{subfigure}%
\begin{subfigure}{0.3\columnwidth}
  \centering
  \includegraphics[width=1\linewidth]{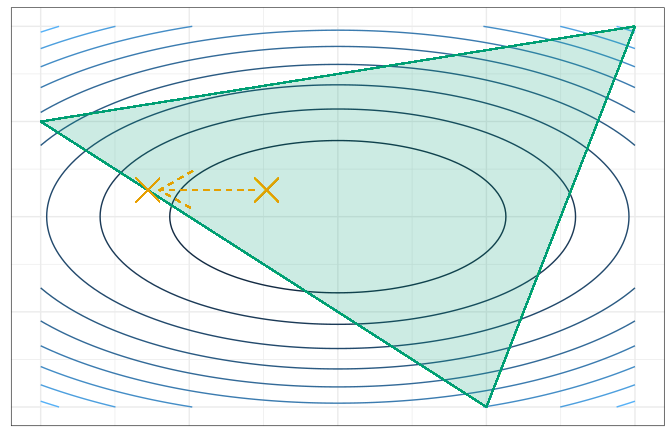}
  \caption{1st iteration}
\end{subfigure}%
\begin{subfigure}{0.3\columnwidth}
  \centering
  \includegraphics[width=1\linewidth]{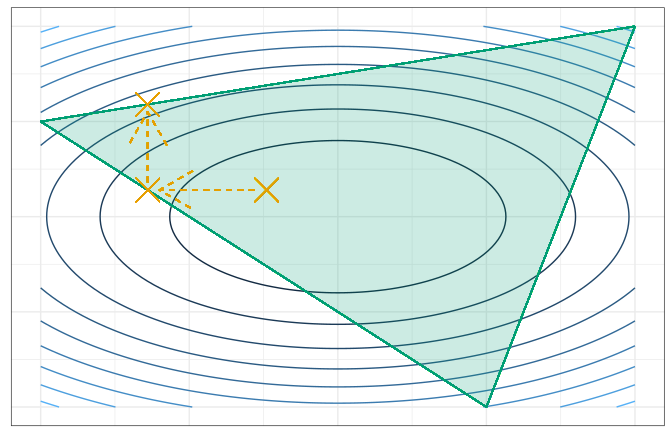}
  \caption{2nd iteration}
\end{subfigure}
\caption{Illustration of the BCA strategy to maximize a convex function under entangled constraints. The concentric curves represent the levels of the function and the green triangle is the feasible set. After the second update, the BCA algorithm is stuck and cannot move anymore but the point is not a local maximum.}
\label{fig:entangled_constraints}
\end{figure}

In the first case, $\mathbf w_l^\top \mathbf M_l \mathbf w_l = \boldsymbol \lambda_l^\top \mathbf W_l^\top \mathbf M_l \mathbf W_l \boldsymbol \lambda_l = \boldsymbol \lambda_l^\top \boldsymbol \lambda_l$. Therefore, the constraint $\mathbf w_l^\top \mathbf M_l \mathbf w_l = 1$ becomes $\Vert \boldsymbol \lambda \Vert_2 = 1$. According to \eqref{update}, we have to find $\mathbf w_l$ that allows to increase the value of the objective function. Using the same trick as in Section \ref{sample_sep_TGCCA}, we can observe that $\mathbf w_l^{(r)} = \mathbf W_{l, (-m)}^{(r)} \mathbf w_{l, m}^{(r)}$ with $\mathbf{W}_{l, (-m)}^{(r)} = \left( \mathbf{w}_{l,d_l}^{(r)} \otimes \dots \otimes \mathbf{w}_{l, m+1}^{(r)} \otimes \mathbf{I}_{p_{l,m}} \otimes \mathbf{w}_{l, m-1}^{(r)} \otimes \dots \otimes \mathbf{w}_{l,1}^{(r)} \right) \in \mathbb{R}^{p_l \times p_{l,m}}$ and that:
\begin{align}
    \nabla_l f(\mathbf w)^\top \mathbf w_l = \sum_{r = 1}^R \lambda_l^{(r)} &\nabla_l f(\mathbf w)^\top \mathbf W_{l, (-m)}^{(r)} \mathbf w_{l,m}^{(r)} = \text{Tr}(\mathbf{F}^\top \mathbf W_{l,m}),
    \label{non-sep_observation} \\
    \text{with} \quad \mathbf{F} = \begin{bmatrix} \mathbf{f}^{(1)} & \dots & \mathbf{f}^{(R_l)} \end{bmatrix} \quad
    &\text{and} \quad \mathbf{f}^{(r)} = \lambda_l^{(r)} \mathbf W_{l,(-m)}^{(r) \top} \nabla_l f(\mathbf w) \in \mathbb{R}^{p_{l,m}}. \nonumber
\end{align}
We can traduce the orthogonality constraint into the following constraints:
\begin{equation*}
    \left\{
    \begin{array}{ll}
            \mathbf w_{l,m}^{(r) \top} \mathbf W_{l, (-m)}^{(r) \top} \mathbf M_l \mathbf W_{l, (-m)}^{(r)} \mathbf w_{l,m}^{(r)} = 1 \\
            \mathbf w_{l,m}^{(r) \top} \mathbf W_{l, (-m)}^{(r) \top} \mathbf M_l \mathbf w_l^{(s)} = 0 \quad \text{for } r \neq s.
        \end{array}
    \right.
\end{equation*}
Unfortunately, it is not possible to find a matrix $\mathbf M_{l, m}$ such that $\mathbf W_{l,m}^\top \mathbf M_{l,m} \mathbf W_{l,m} = \mathbf I_{R_l}$. Therefore, we cannot end up with a problem of the following kind:
\begin{equation*}
    \argmax{\mathbf W_{l, m}} \text{Tr} (\mathbf{F}^\top \mathbf W_{l,m}) \quad \text{s.t.} \quad \mathbf W_{l,m}^\top \mathbf M_{l,m} \mathbf W_{l,m} = \mathbf I_{R_l}.
\end{equation*}
In this case, we did not find a better solution than to solve for each $\mathbf w_{l,m}^{(r)}$ in turn the optimization problems: 
\begin{equation}
\label{bad_TGCCA_optim}
    \argmax{\mathbf w_{l,m}^{(r)}} \mathbf f^{(r)\top} \mathbf w_{l, m}^{(r)} \quad \text{s.t.} \quad \left\{
    \begin{array}{ll}
            \mathbf w_{l,m}^{(r) \top} \mathbf W_{l, (-m)}^{(r) \top} \mathbf M_l \mathbf W_{l, (-m)}^{(r)} \mathbf w_{l,m}^{(r)} = 1 \\
            \mathbf w_{l,m}^{(r) \top} \mathbf W_{l, (-m)}^{(r) \top} \mathbf M_l \mathbf w_l^{(s)} = 0 \quad \text{for } r \neq s.
        \end{array}
    \right.
\end{equation}
By writing the Lagrangian of \eqref{bad_TGCCA_optim}, it is possible to find a closed form solution.
The issue is that the constraints are no longer separated between the different $\mathbf w_{l,m}^{(r)}$, which leads the optimization scheme to get stuck in uninteresting points. See Figure \ref{fig:entangled_constraints} for an illustration of this phenomenon in a toy case. This is why we looked for another choice of the matrix $\mathbf K_l$ and decided to try with $\mathbf K_l = \mathbf I_{R_l}$.

\subsection{Update for non-separable TGCCA}
Choosing $\mathbf K_l = \mathbf I_{R_l}$ leads to the optimization problem introduced in equation \eqref{non-separable_TGCCA_optim}:
\begin{align*}
    \underset{\mathbf w_1, \ldots, \mathbf w_L}{\text{maximize}} &\sum_{l,k=1}^L c_{lk} \text{g}\left(\mathbf w_l^\top \mathbf \Sigma_{lk} \mathbf w_k\right) \\
    \text{s.t. } \mathbf w_l = [\![ \boldsymbol \lambda; \mathbf W_{l, 1}, \dots, \mathbf W_{l, d_l} ]\!], ~ \mathbf W_{l, m} \in &\mathbb{R}^{p_{l,m} \times R_l}, ~ \mathbf W_l^\top \mathbf W_l = \mathbf I_{R_l}, ~ \Vert \boldsymbol \lambda_l \Vert_2 \leq \Vert \mathbf M_l \Vert_2^{-\frac{1}{2}} ~ l \in [L]. 
\end{align*}
As shown in \eqref{non-sep_observation}, for each $\mathbf W_{l,m}$, we have 
\begin{equation}
    \text{r}_{l,m}(\mathbf w) =  \argmax{\mathbf W_{l,m}, \mathbf W_{l,m}^\top \mathbf W_{l,m} = \mathbf I_{R_l}} \text{Tr} (\mathbf F^\top \mathbf W_{l,m}) = \mathbf S \mathbf T^\top,
\end{equation}
where $\mathbf S$ and $\mathbf T$ are respectively the left and right singular vectors of the rank-$R_l$ SVD of $\mathbf F$. Hence we get the same kind of solution as \ref{r_m}.

What is specific to this non-separable case is the update of $\boldsymbol \lambda_l$. To simplify the notations, we will note $\mathbf u_l = \mathbf W_l^\top \nabla_l f(\mathbf w)$. We could try to do like before and search for 
\begin{equation*}
    \boldsymbol \lambda_l^\text{opt} = \argmax{\boldsymbol \lambda_l, \Vert \boldsymbol \lambda_l \Vert_2 \leq \Vert \mathbf M_l \Vert_2^{-\frac{1}{2}}} \mathbf u_l^\top \boldsymbol \lambda_l = \frac{\mathbf u_l}{\Vert \mathbf M_l \Vert_2^{-\frac{1}{2}}\Vert \mathbf u_l \Vert_2}.
\end{equation*}
However, this solution implies that we do not take into account the structure of $\mathbf M_l$ in our optimization scheme. As our first goal was to solve $\eqref{dream_objective}$, for a fixed $\mathbf W_l$, we consider:
\begin{equation*}
    \boldsymbol \lambda_l^\text{ref} = \argmax{\boldsymbol \lambda_l,  \boldsymbol \lambda_l^\top \mathbf W_l^\top \mathbf M_l \mathbf W_l \boldsymbol \lambda_l = 1 } \mathbf u_l^\top \boldsymbol \lambda_l = \frac{\left( \mathbf W_l^\top \mathbf M_l \mathbf W_l \right)^{-1} \mathbf u_l}{\sqrt{\mathbf u_l^\top \left( \mathbf W_l^\top \mathbf M_l \mathbf W_l \right)^{-1} \mathbf u_l}}.
\end{equation*}
To ensure that $\boldsymbol \lambda_l^{\text{ref}\top} \mathbf W_l^\top \mathbf M_l \mathbf W_l \boldsymbol \lambda_l^\text{ref}$ remains below 1 after the update of $\mathbf W_l$, we showed in Section \ref{pop_TGCCA} that we need to normalize $\boldsymbol \lambda_l^\text{ref}$ so that $\Vert \boldsymbol \lambda_l^\text{ref} \Vert_2 \leq \Vert \mathbf M_l \Vert_2^{-\frac{1}{2}}$. Unfortunately, there is no guarantee that this new point would still allow to increase the value of the objective function. 

Let $\boldsymbol \lambda_l^\text{prev}$ be the value of $\boldsymbol \lambda_l$ before the update. We define the following ball and hyperplane:
\begin{align*}
    \mathcal{B}_\alpha &= \{ \boldsymbol \lambda_l \in \mathbb{R}^{R_l}; ~ \boldsymbol \lambda_l^\top \boldsymbol \lambda_l \leq \alpha \} \\
    \mathcal{H}_\epsilon &= \{ \boldsymbol \lambda_l \in \mathbb{R}^{R_l}; ~ \mathbf u_l^\top \boldsymbol \lambda_l \geq \epsilon \}
\end{align*}
We can aim for a compromise between $\boldsymbol \lambda_l^\text{ref}$ and $\boldsymbol \lambda_l^\text{opt}$ by projecting $\boldsymbol \lambda_l^\text{ref}$ on the intersection of the ball $\mathcal{B}_\alpha$ and $\mathcal{H}_\epsilon$ with $\alpha = \Vert \mathbf M_l \Vert_2^{-1}$ and $\epsilon \in \left[\mathbf u_l^\top \boldsymbol \lambda_l^\text{prev}, \mathbf u_l^\top \boldsymbol \lambda_l^\text{opt} \right]$. We arbitrarily choose $\epsilon = \frac{1}{2} \left(\mathbf u_l^\top \boldsymbol \lambda_l^\text{prev} + \mathbf u_l^\top \boldsymbol \lambda_l^\text{opt} \right)$. From the definitions of $\boldsymbol \lambda_l^\text{prev}$ and $\boldsymbol \lambda_l^\text{opt}$ we are sure that $\epsilon \geq 0$. 

Once we have computed $\boldsymbol \lambda_l^\text{ref}$, two cases arise: either $\frac{\mathbf u_l^\top \boldsymbol \lambda_l^\text{ref}}{\Vert \mathbf M_l \Vert_2^{-\frac{1}{2}}\Vert \boldsymbol \lambda_l^\text{ref} \Vert_2} \geq \epsilon$ or $\frac{\mathbf u_l^\top \boldsymbol \lambda_l^\text{ref}}{\Vert \mathbf M_l \Vert_2^{-\frac{1}{2}}\Vert \boldsymbol \lambda_l^\text{ref} \Vert_2} < \epsilon$.
In the first case, we can take
\begin{equation}
    \label{r_lambda_non_sep1}
    \text{r}_{l, \boldsymbol \lambda} (\mathbf w) = \frac{\boldsymbol \lambda_l^\text{ref}}{\Vert \mathbf M_l \Vert_2^{-\frac{1}{2}}\Vert \boldsymbol \lambda_l^\text{ref} \Vert_2}.
\end{equation}
In the second case, we have to find the projection $\boldsymbol \lambda_l^\text{ref}$ on the intersection of the frontiers of $\mathcal{B}_\alpha$ and $\mathcal{H}_\epsilon$. This is equivalent to solve
\begin{equation}
    \label{projection_optim}
    \argmin{\boldsymbol \lambda_l} \frac{1}{2}\Vert \boldsymbol \lambda_l - \boldsymbol \lambda_l^\text{ref} \Vert_2^2 \quad \text{s.t.} \quad \boldsymbol \lambda_l^\top \boldsymbol \lambda_l = \alpha \text{ and } \mathbf u_l^\top \boldsymbol \lambda_l = \epsilon
\end{equation}
The Lagrangian associated with optimization problem \eqref{projection_optim} is
\begin{equation*}
    \mathcal{L}(\boldsymbol \lambda_l, \mu, \nu) = \frac{1}{2}\Vert \boldsymbol \lambda_l - \boldsymbol \lambda_l^\text{ref} \Vert_2^2 + \frac{1}{2}\mu(\boldsymbol \lambda_l^\top \boldsymbol \lambda_l - \alpha) + \nu(\mathbf u_l^\top \boldsymbol \lambda_l - \epsilon),
\end{equation*}
where $\mu, \nu \in \mathbb{R}$ are the Lagrange multipliers.
Cancelling the derivative of the Lagrangian function with respect to $\boldsymbol \lambda_l$ yields the following stationary equation:
\begin{equation}
    \label{projection_s1}
    (1 + \hat{\mu}) \hat{\boldsymbol \lambda}_l - \boldsymbol \lambda_l^\text{ref} + \hat{\nu} \mathbf u_l = 0
\end{equation}
We can already notice that, if $(1 + \hat{\mu}) = 0$, $\boldsymbol \lambda_l^\text{ref}$ is collinear with $\mathbf u_l$, so $\frac{\boldsymbol \lambda_l^\text{ref}}{\Vert \mathbf M_l \Vert_2^{-\frac{1}{2}}\Vert \boldsymbol \lambda_l^\text{ref} \Vert_2} = \boldsymbol \lambda_l^\text{opt}$. Thus, the optimal point we seek is $\hat{\boldsymbol \lambda}_l = \boldsymbol \lambda_l^\text{opt}$. We suppose now that $(1 + \hat{\mu}) \neq 0$.

Left multiplying \eqref{projection_s1} by $\mathbf u_l^\top$, it comes
\begin{align}
    &\epsilon (1 + \hat{\mu}) - \mathbf u_l^\top \boldsymbol \lambda_l^\text{ref} + \hat{\nu} \mathbf u_l^\top \mathbf u_l = 0 \nonumber \\
    \Rightarrow \quad &\hat{\nu} = \frac{\mathbf u_l^\top \boldsymbol \lambda_l^\text{ref} - \epsilon (1 + \hat{\mu})}{\mathbf u_l^\top \mathbf u_l}.
    \label{projection_s2}
\end{align}
Left multiplying \eqref{projection_s1} by $\hat{\boldsymbol \lambda}_l^\top$, we get
\begin{align*}
    &\alpha(1 + \hat{\mu}) - \hat{\boldsymbol \lambda}_l^\top \boldsymbol \lambda_l^\text{ref} + \hat{\nu} \epsilon = 0 \\
    \Rightarrow \quad &\alpha (1 + \hat{\mu})^2 + \hat{\nu} \epsilon (1 + \hat{\mu}) + (\hat{\nu} \mathbf u_l - \boldsymbol \lambda_l^\text{ref})^\top \boldsymbol \lambda_l^\text{ref} = 0 \text{ multiplying by } (1 + \hat{\mu}) \text{ and reinjecting } \eqref{projection_s1} \\
    \Rightarrow \quad &\left(\alpha - \frac{\epsilon^2}{\mathbf u_l^\top \mathbf u_l}\right)(1 + \hat{\mu})^2 
    - \boldsymbol \lambda_l^{\text{ref} \top} \boldsymbol \lambda_l^\text{ref} + \frac{\left(\mathbf u_l^\top \boldsymbol \lambda_l^\text{ref}\right)^2}{\mathbf u_l^\top \mathbf u_l} = 0 \text{ reinjecting } \eqref{projection_s2} 
\end{align*}
We observe that $(1 + \hat{\mu})$ satisfies a binomial equation and that 
\begin{align*}
    \boldsymbol \lambda_l^{\text{ref} \top} \boldsymbol \lambda_l^\text{ref} - \frac{\left(\mathbf u_l^\top \boldsymbol \lambda_l^\text{ref}\right)^2}{\mathbf u_l^\top \mathbf u_l} &= \boldsymbol \lambda_l^{\text{ref}\top} \left( \mathbf I_{R_l} - \mathbf u_l (\mathbf u_l^\top \mathbf u_l)^{-1} \mathbf u_l^\top \right) \boldsymbol \lambda_l^\text{ref} \\
    &= \boldsymbol \lambda_l^{\text{ref}\top} \mathbf P_{\mathbf u_l^\perp} \boldsymbol \lambda_l^\text{ref}\\
    &= \Vert \mathbf P_{\mathbf u_l^\perp} \boldsymbol \lambda_l^\text{ref} \Vert_2^2,
\end{align*}
where $\mathbf P_{\mathbf u_l^\perp}$ is the projector on the hyperplane orthogonal to span($\mathbf u_l$).
We can show that, for every $\epsilon \in \left[\mathbf u_l^\top \boldsymbol \lambda_l^\text{prev}, \mathbf u_l^\top \boldsymbol \lambda_l^\text{opt} \right]$, $\frac{\epsilon^2}{\mathbf u_l^\top \mathbf u_l} \leq \alpha$. Indeed, $\forall \epsilon \in \left[\mathbf u_l^\top \boldsymbol \lambda_l^\text{prev}, \mathbf u_l^\top \boldsymbol \lambda_l^\text{opt} \right]$, $\exists \gamma \in [0, 1], \epsilon = \left((1 - \gamma) \boldsymbol \lambda_l^\text{prev} + \gamma \boldsymbol \lambda_l^\text{opt}\right)^\top \mathbf u_l$. Therefore,
\begin{align*}
    \frac{\epsilon^2}{\mathbf u_l^\top \mathbf u_l} &= \left((1 - \gamma) \boldsymbol \lambda_l^\text{prev} + \gamma \boldsymbol \lambda_l^\text{opt}\right)^\top \mathbf u_l (\mathbf u_l^\top \mathbf u_l)^{-1} \mathbf u_l^\top \left((1 - \gamma) \boldsymbol \lambda_l^\text{prev} + \gamma \boldsymbol \lambda_l^\text{opt}\right) \\
    &= \Vert \mathbf P_{\mathbf u_l} \left((1 - \gamma) \boldsymbol \lambda_l^\text{prev} + \gamma \boldsymbol \lambda_l^\text{opt}\right) \Vert_2^2 \text{ where } \mathbf P_{\mathbf u_l} \text{ is the orthogonal projector on span}(\mathbf u_l)\\
    &\leq \Vert \left((1 - \gamma) \boldsymbol \lambda_l^\text{prev} + \gamma \boldsymbol \lambda_l^\text{opt}\right) \Vert_2^2 \text{ since } \Vert \mathbf P_{\mathbf u_l} \Vert_2 \leq 1 \\
    &\leq (1 - \gamma) \Vert \boldsymbol \lambda_l^\text{prev} \Vert_2^2 + \gamma \Vert \boldsymbol \lambda_l^\text{opt} \Vert_2^2 \text{ using the norm convexity} \\
    &= \alpha.
\end{align*}
If $\frac{\epsilon^2}{\mathbf u_l^\top \mathbf u_l} = \alpha$, we have that $\mathbf P_{\mathbf u_l^\perp} \boldsymbol \lambda_l^\text{ref} = 0$ so we fall back to the case where $\boldsymbol \lambda_l^\text{ref}$ is collinear with $\mathbf u_l$. Otherwise we get 
\begin{equation}
\label{projection_s3}
    (1 + \hat{\mu}) = \pm \frac{\Vert \mathbf P_{\mathbf u_l^\perp} \boldsymbol \lambda_l^\text{ref} \Vert_2}{\sqrt{\alpha - \frac{\epsilon^2}{\mathbf u_l^\top \mathbf u_l}}}.
\end{equation}

Finally, if $\boldsymbol \lambda_l^\text{ref}$ is not collinear with $\mathbf u_l$, using equations \eqref{projection_s1}, \eqref{projection_s2} and \eqref{projection_s3}, we find
\begin{equation}
    \hat{\boldsymbol \lambda}_l = \frac{\epsilon}{\mathbf u_l^\top \mathbf u_l} \mathbf u_l + \sqrt{\alpha - \frac{\epsilon^2}{\mathbf u_l^\top \mathbf u_l}}\frac{\mathbf P_{\mathbf u_l^\perp} \boldsymbol \lambda_l^\text{ref}}{\Vert \mathbf P_{\mathbf u_l^\perp} \boldsymbol \lambda_l^\text{ref} \Vert_2},
\end{equation}
where the sign is determined by looking for the solution that minimizes \eqref{projection_optim}.
Thanks to this formulation, we managed to find an update for $\hat{\boldsymbol \lambda}_l$ that takes into account the structure of $\mathbf M_l$ through the contribution of $\boldsymbol \lambda_l^\text{ref}$.

\newpage
\section{Compactness of the feasible set}
\label{Appendix-Compactness}
The proof of Proposition 1 directly follows the demonstration proposed in \cite{Tenenhaus2017}. It relies on two key ingredients: (i) the compactness of the feasible set of the optimization problem and (ii) the uniqueness of the update. While the uniqueness of the update relies on rank conditions and cannot be a priori verified but can be monitored at runtime, the compactness of the feasible set can be proven as follows.

The feasible set $\Omega_l$ associated with both problems \eqref{separable_TGCCA_optim} and \eqref{non-separable_TGCCA_optim} can be defined as:
\begin{equation}
\Omega_l = \{ \mathbf{w}_l \in \mathbb{R}^{p_l}; ~ \mathbf{w}_l = \sum_{r = 1}^{R_l} \lambda_l^{(r)} \mathbf{w}_l^{(r)}; ~ \mathbf{w}_l^{(r)} = \mathbf{w}_{l, d_l}^{(r)} \otimes \dots \otimes \mathbf{w}_{l,1}^{(r)}; ~ \mathbf W_l^\top \mathbf W_l = \mathbf I_{R_l} ; ~ \boldsymbol{\lambda}_l^\top \boldsymbol{\lambda}_l \leq \alpha \}.
\label{Omega}
\end{equation}
In the case of \eqref{separable_TGCCA_optim}, we have $\boldsymbol{\lambda}_l^\top \boldsymbol{\lambda}_l = 1$ but this does not change the proof.

To show the result, we introduce the following sets:
\begin{align*}
    \Omega_\text{norm} &= \{ \mathbf{w}_l \in \mathbb{R}^{p_l}; \quad \mathbf{w}_l^\top \mathbf{w}_l = 1 \}, \\
    \Omega_\text{kron} &= \{ \mathbf{w}_l \in \mathbb{R}^{p_l}; \quad \mathbf{w}_l = \mathbf{w}_{l,d_l} \otimes \dots \otimes \mathbf{w}_{l,1} \}, \\
    \Omega_\text{mat} &= \{  \mathbf{W}_l \in \mathbb{R}^{p_l \times R_l}; \quad \mathbf{W}_l = \begin{bmatrix} \mathbf{w}_l^{(1)} & \dots & \mathbf{w}_l^{(R_l)} \end{bmatrix}; \quad \mathbf{w}_l^{(r)} \in \Omega_\text{norm} \cap \Omega_\text{kron} \}, \\
    \Omega_\text{orth} &= \{  \mathbf{W}_l \in \mathbb{R}^{p_l \times R_l}; \quad \mathbf{W}_l^\top \mathbf{W}_l = \mathbf{I}_{R_l} \}, \\
    \Lambda_l &= \{\boldsymbol{\lambda}_l \in \mathbb{R}^{R_l}; \quad \boldsymbol{\lambda}_l^\top \boldsymbol{\lambda}_l \leq \alpha \}.
\end{align*}
Using these sets, a new way to express $\Omega_l$ is derived:
\begin{equation*}
    \Omega_l = \{ \mathbf{w}_l \in \mathbb{R}^{p_l}; \quad \mathbf{w}_l = \mathbf{W}_l \boldsymbol{\lambda}_l; \quad \mathbf{W}_l \in \Omega_\text{mat} \cap \Omega_\text{orth}; \quad \boldsymbol{\lambda}_l \in \Lambda_l \}
\end{equation*}
Therefore, $\Omega_l$ is the image of the set $\Lambda_l \times (\Omega_\text{mat} \cap \Omega_\text{orth})$ by the continuous application $f:(\boldsymbol{\lambda}_l, \mathbf{W}_l) \mapsto \mathbf{W}_l \boldsymbol{\lambda}_l$. The proof that $\Omega_l$ is compact then reduces to prove that $\Lambda_l$ and $\Omega_\text{mat} \cap \Omega_\text{orth}$ are compact.

$\Lambda_l$ is compact as the norm-2 ball of radius $\alpha$ in $\mathbb{R}^{R_l}$ which is of finite dimension. In the case $\boldsymbol{\lambda}_l^\top \boldsymbol{\lambda}_l = \alpha$, $\Lambda_l$ becomes the boundary of this ball which remains a compact.

We will now show that $\Omega_\text{mat} \cap \Omega_\text{orth}$ is compact as the intersection of a compact set with a closed set.
\begin{itemize}
    \item $\Omega_\text{norm}$ is a compact set as the boundary of the norm-2 unit ball.
    \item $\Omega_\text{kron}$ is a closed set. It is a standard result in geometric algebra and a specificity of the set of rank-1 tensors. 
    \item $\Omega_\text{norm} \cap \Omega_\text{kron}$ is compact as the intersection of a closed and a compact set.
    \item $\Omega_\text{mat}$ is the image of $\times_{R_l}(\Omega_\text{norm} \cap \Omega_\text{kron})$ (the Cartesian product $R_l$ times) by the continuous operator that arranges vectors into a matrix. Hence, $\Omega_\text{mat}$ is compact. 
    \item $\Omega_\text{orth}$ is the set of semi-orthogonal matrices. This set is closed (and even compact but we only need it to be closed).
\end{itemize}
Consequently, $\Omega_\text{mat} \cap \Omega_\text{orth}$ is compact and we have shown that $\Omega_l$ is a compact set.

In the case where only one mode $m$ bears the orthogonality, similar arguments can be derived to show that $\Omega_l$ is indeed compact.

\newpage
\section{Rank versus number of canonical components}
\label{rank-vs-comp}
Introducing CCA with canonical vectors admitting a rank-$R$ CP decomposition may create a confusion between the rank of the decomposition and the number of extracted canonical components. This section describes the differences between the two concepts and between MGCCA with the deflation procedure imposing "orthogonality on the weight vectors" (see Section 2.4.2 of \cite{Gloaguen2020}) and TGCCA.

\subsection{Extracting K canonical components}
\label{Appendix:deflation}
In the main text, the described optimization problems aim to find the first canonical component for each block given by $\mathbf y_l^{[1]} = \mathbf X_l \mathbf w_l^{[1]}$. This first canonical component summarizes the information between and within the blocks but as the first component of Principal Component Analysis (PCA) alone does not always wholly explain the dataset, this first canonical component will not always be enough. We want to find other canonical components by finding new sets of canonical vectors in this context. One of the possible ways is to impose that the new canonical components are not correlated with the previous ones. This can be written:
\begin{align}
    \underset{\mathbf w_1^{[2]}, \ldots, \mathbf w_L^{[2]}}{\text{maximize}} \sum_{l,k=1}^L c_{lk} \text{g}\left(\mathbf w_l^{[2]\top} \mathbf \Sigma_{lk} \mathbf w_k^{[2]}\right) \\
    \text{s.t.} \quad \left\{
    \begin{array}{ll}
            \mathbf w_l^{[2]\top} \mathbf M_l \mathbf w_l^{[2]} = 1,  ~ l \in [L] \\
            \mathbf w_l^{[2]\top} \mathbf \Sigma_{lk} \mathbf w_k^{[1]} = 0,  ~ (l, k) \in [L]^2.
        \end{array}
    \right. \nonumber
\end{align}
This is the usual way of searching for the next canonical components, see for example \cite{10.1162/0899766042321814}. This can be implemented in practice by applying a deflation strategy to each block $\mathbf X_l$: $\mathbf X_l^{[1]} = \mathbf X_l - \mathbf y_l^{[1]} \left(\mathbf y_l^{[1] \top} \mathbf y_l^{[1]}\right)^{-1} \mathbf y_l^{[1] \top} \mathbf X_l$, consider $\boldsymbol \Sigma_{lk}^{[1]}$ the covariance between $\mathbf X_l^{[1]}$ and $\mathbf X_k^{[1]}$, and solve:
\begin{align}
    \underset{\mathbf w_1^{[2]}, \ldots, \mathbf w_L^{[2]}}{\text{maximize}} \sum_{l,k=1}^L c_{lk} \text{g}\left(\mathbf w_l^{[2]\top} \mathbf \Sigma_{lk}^{[1]} \mathbf w_k^{[2]}\right) \quad
    \text{s.t.} \quad \mathbf w_l^{[2]\top} \mathbf M_l \mathbf w_l^{[2]} = 1,  ~ l \in [L].
\end{align}
This is what is done for example in Section 2.4.1 of \cite{Gloaguen2020}.

Another possibility is to impose different constraints on the new set of canonical vectors with respect to the first ones. In Section 2.4.2 of \cite{Gloaguen2020}, the authors impose that $\mathbf w_l^{[2] \top} \mathbf w_l^{[1]} = 0$. Therefore the different canonical vectors, for a given block, are orthogonal. Using the rank-1 CP decompositions of the canonical vectors, it imposes that either $\mathbf w_{l,1}^{[2] \top} \mathbf w_{l,1}^{[1]} = 0$ or $\mathbf w_{l,2}^{[2] \top} \mathbf w_{l, 2}^{[1]} = 0$. They show that this leads to the proposition of a new deflation procedure that guarantees to get a new canonical vector satisfying the constraint.

Whatever the choices of constraints on the different sets of canonical vectors, the procedures can be iterated to extract $K$ sets of canonical vectors.

\subsection{Rank-R TGCCA vs MGCCA with R components}
Using the last presented approach and extracting R components, MGCCA generates $\mathbf W_l^\text{MGCCA} = \begin{bmatrix} \mathbf w_l^{[1]}, \dots \mathbf w_l^{[R]} \end{bmatrix}$ for each block. From this, we can construct $\mathbf w_l^\text{MGCCA} = \sum_{r=1}^R \mathbf w_l^{[r]}$ which results in a vector that admits an orthogonal rank-$R$ CP decomposition for which each factor of the decomposition has the same contribution ($\forall r \in [R], \boldsymbol \lambda_l^{[r]} = 1$). On the contrary, rank-$R$ TGCCA generates $\mathbf w_l^\text{TGCCA} = \sum_{r=1}^R \lambda_l^{(r)} \mathbf w_l^{(r)}$ so TGCCA has the flexibility to weight differently the different factors and can reduce the importance of the factors that are just modelling noise. Furthermore, the criteria they optimize are different:
\begin{align}
    \text{crit}^\text{MGCCA} &= \sum_{l, k = 1}^L c_{lk} \sum_{r = 1}^R \text{g} \left( \mathbf w_l^{[r] \top} \boldsymbol \Sigma_{lk}^{[r]} \mathbf w_k^{[r]} \right), \\
    \text{crit}^\text{TGCCA} &= \sum_{l, k = 1}^L c_{lk} \text{g} \left( \sum_{r, s = 1}^R \lambda_l^{(r)} \lambda_l^{(s)} \mathbf w_l^{(r) \top} \boldsymbol \Sigma_{lk} \mathbf w_k^{(s)} \right).
\end{align}
Ignoring the $\text{g}$ function and the modified $\boldsymbol \Sigma_{lk}^{[r]}$ matrices, we can see that MGCCA only takes into account the interactions between the same levels of factors, while TGCCA takes into account all the interactions.
\newpage
\section{Complexity analysis}
\label{Appendix-Complexity}
We propose in this section a time complexity analysis between RGCCA and the separable version of TGCCA. To simplify the computations and notations, we will say that all blocks have the same number of modes $d$, the same number of variables per mode $q$ and therefore the same total number of variables $p = q^d$. We note $n$ the number of observations and $L$ the number of blocks. We will give for the two methods the complexity of the update and the complexity of the initialization. We will also consider that the estimates of the regularization matrices $\hat{\mathbf M}_l$ are regularized estimates of the block covariances $\boldsymbol \Sigma_{ll}$.

\subsection{RGCCA initialization}
In the case of RGCCA, $\hat{\mathbf M}_l = \tau \mathbf I_p + (1 - \tau) \frac{\mathbf X_l^\top \mathbf X_l}{n}$. We can apply the same change of variable as in the separable version of TGCCA (see Section \ref{sample_sep_TGCCA}). In practice, we need to compute $\hat{\mathbf M}_l^{-\frac{1}{2}}$ and $\mathbf X_l \hat{\mathbf M}_l^{-\frac{1}{2}}$. Using the SVD of $\mathbf X_l$, we can write $\mathbf X_l = \mathbf S \boldsymbol \Delta \mathbf T^\top$, and we get $\hat{\mathbf M}_l^{-\frac{1}{2}} = \mathbf T (\tau + \frac{(1 - \tau)}{n} \boldsymbol \Delta^2)^{-\frac{1}{2}} \mathbf T^\top$. The complexity of the SVD of $\mathbf X_l$ is $\mathcal{O}(np\min(n, p))$ and the computation of $\mathbf M_l$ is $\mathcal{O}(p^2\min(n, p))$. The complexity of computing $\mathbf X_l \hat{\mathbf M}_l^{-\frac{1}{2}}$ is $\mathcal{O}(np^2)$.

We can show that this change of variable step is the bottleneck of RGCCA initialization and, as we need to do it for every block, the overall complexity is $\mathcal{O}(L(np(p + \min(n,p))) = \mathcal{O}(Lnp^2)$.

\subsection{Separable TGCCA initialization}
In the case of separable TGCCA, $\hat{\mathbf M}_l = \hat{\mathbf M}_{l, d} \otimes \dots \otimes \hat{\mathbf M}_{l, 1}= \left(\hat{\mathbf \Sigma}_{ll,d} + \sqrt[d]{\tau}\mathbf{I}_{p_{d}}\right) \otimes \dots \otimes \left(\hat{\mathbf{\Sigma}}_{ll,1} + \sqrt[d]{\tau}\mathbf{I}_{p_1}\right)$ and $\hat{\mathbf \Sigma}_{ll,m}$ are obtained as in Section 4.2.1 of \cite{Min}. Each of these estimates can be computed in $\mathcal{O}(npq)$. 
Then, we need to compute $\hat{\mathbf M}_{l, m}^{-\frac{1}{2}}$, which can be done in $\mathcal{O}(q^3)$. Finally, $\mathbf X_l \hat{\mathbf M}_{l}^{-\frac{1}{2}}$ must be computed, this can be done efficiently by folding $\mathbf X_l$ back to $\TT{X}_l \in \mathbb{R}^{n \times q \times \dots \times q}$ and computing $\TT{X}_l \times_2 \hat{\mathbf M}_{l, 1}^{-\frac{1}{2}} \dots \times_{d + 1} \hat{\mathbf M}_{l, d}^{-\frac{1}{2}}$, where $\times_m$ represents the mode-product between the $m^\text{th}$-mode of $\TT{X}_l$ and $\hat{\mathbf M}_{l, m - 1}^{-\frac{1}{2}}$. Each of these operations has a complexity of $\mathcal{O}(npq)$. 

Since there are $d$ operations per block and there are $L$ blocks, the total cost of the change of variables is $\mathcal{O}(Ldq(np + q^2))$.
As for RGCCA, this is also the bottleneck for the separable TGCCA initialization. Considering that there are at least 2 modes (i.e., $d \geq 2$) and that all modes have the same number of variables $q$, the complexity can be simplified to $\mathcal{O}(Lnpdq)$. We can see that we managed to trade a factor $dq$ against a factor $p = q^d$ in RGCCA.

\subsection{Updates}
The bottlenecks of both RGCCA and separable TGCCA are the computations of $\mathbf y_l = \mathbf X_l \mathbf w_l$ and $\mathbf z_l = \sum_{k = 1}^L c_{kl} \text{g}'\left( \mathbf y_l^\top \mathbf y_k \right) \mathbf y_k$. The associated complexities are respectively $\mathcal{O}(np)$ and $\mathcal{O}(Ln)$. For both methods, these computations have to be repeated for each block but for separable TGCCA, these computations also have to be repeated for each mode. Therefore we get that the complexities of the updates of RGCCA and separable TGCCA are respectively $\mathcal{O}(Ln( p + L))$ and $\mathcal{O}(Ldn (p + L))$. We can see that, compared to RGCCA, we lose a factor $d$ on the complexity of the update. This is due to the fact that we have added an extra layer of BCA compared to RGCCA.

If we note $T^\text{RGCCA}$ and $T^\text{TGCCA}$ the number of iterations needed for RGCCA and separable TGCCA to reach convergence, we have:
\begin{align*}
    \text{complexity}(\text{RGCCA}) &= \mathcal{O}(Lnp^2 + T^\text{RGCCA} Ln (p + L)) \\
    \text{complexity}(\text{TGCCA}) &= \mathcal{O}(Lnpdq + T^\text{TGCCA} Lnd (p + L)). 
\end{align*}
\newpage
\section{Retrieving the chemicals}
\label{Appendix-Chemicals}
\begin{figure}[ht]
    \centering
    \includegraphics[width=0.8\linewidth]{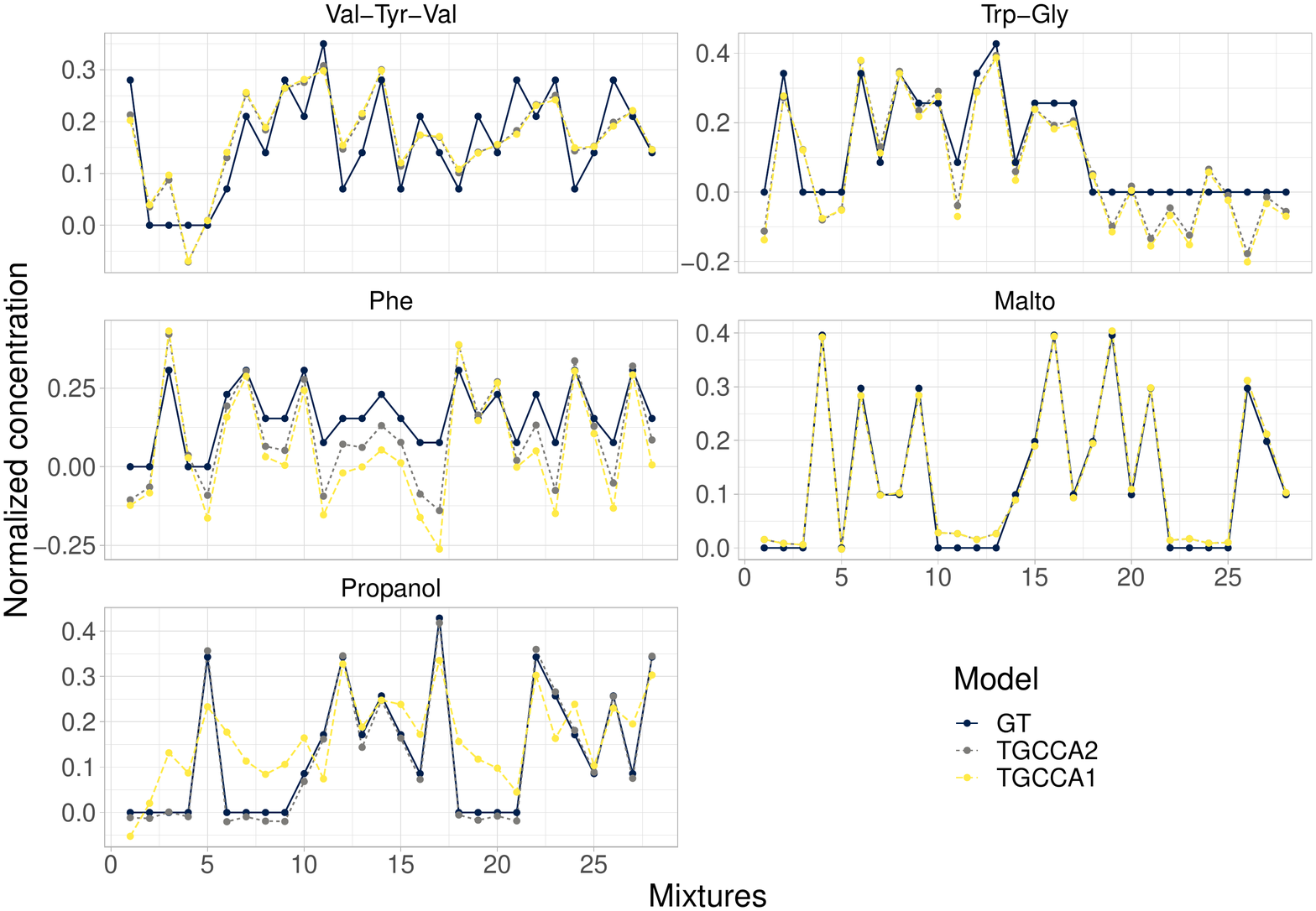}
    \caption{Normalized concentrations of the different chemicals for both rank-2 TGCCA and rank-1 TGCCA, we can see that rank-1 TGCCA does not capture Propanol as well as rank-2 TGCCA.}
    \label{fig:TGCCA1_vs_TGCCA2}
\end{figure}
In the study of the dataset from \cite{Acar2014}, we are interested in finding the concentrations of the five chemicals in the 28 available mixtures. This information is contained in the matrix $\mathbf A$ following the notation of \eqref{CMTF}. Nevertheless, we do not estimate this matrix using TGCCA. To avoid this problem, we suppose that, if $\mathbf B$ and $\mathbf C$ are well estimated, $\mathbf A$ can be deduced through the following optimisation problem:
\begin{equation}
    \argmin{\mathbf A} \Vert \mathbf X_{(1)} - \mathbf A \boldsymbol \Lambda (\mathbf B \odot \mathbf C)^\top \Vert_F^2,
\end{equation}
where $\boldsymbol \Lambda = \text{diag}(\boldsymbol \lambda)$. Hence we get
\begin{equation}
    \label{A}
    \mathbf A = \mathbf X_{(1)} \boldsymbol \Lambda (\mathbf B \odot \mathbf C) \left( \boldsymbol \Lambda (\mathbf B \odot \mathbf C)^\top (\mathbf B \odot \mathbf C) \boldsymbol \Lambda \right)^{-1}.
\end{equation}
In the case of TGCCA of rank 2, we extract for the first block $\mathbf W_{1, 1} \in \mathbb{R}^{13324 \times 2}$ and $\mathbf W_{1, 2} \in \mathbb{R}^{8 \times 2}$. We use these two matrices respectively as our matrices $\mathbf C$ and $\mathbf B$. As we imposed orthogonality constraints on columns of $\mathbf B$ and $\mathbf C$, equation \eqref{A} gives $\mathbf A = \mathbf X_{(1)} (\mathbf B \odot \mathbf C) \boldsymbol \Lambda^{-1}$. This gives us the first two columns of the matrix $\mathbf A$. For the next ones, we deflate the $\TT{X}$ tensor and repeat the procedure. As the next extracted components are of rank 1, 
the next columns of $\mathbf A$ are computed as $\mathbf X_{(1)}^{[k]} \mathbf w_1^{[k + 1]}$ for $k \in [3]$ where $\mathbf X_{(1)}^{[k]}$ is the mode-1 matricization of the tensor after its $k^\text{th}$ deflation, and $\mathbf w_1^{[k + 1]}$ is the associated canonical vector returned by TGCCA.

\subsection{Extracting Propanol}
According to \cite{Acar2014}, the concentration of Propanol cannot be inferred from the matrix block. Therefore, we do not expect to properly find it as a component extracted by TGCCA. Instead, we use the fact that overestimating the rank of the canonical vector leads to estimating factors that explain some extra variance of the block. This is why we look for a first component of rank 2 and expect to find Propanol as the second extracted factor. This is indeed the case, and we can see that $\lambda_1^{(2)}$, the weight associated with the second factor of the first block, is really low ($\approx 0.03$). This means that TGCCA remains robust in estimating the correlated component, even when the rank is overestimated. Figure \ref{fig:TGCCA1_vs_TGCCA2} shows that TGCCA with only rank-1 factors (i.e., MGCCA) cannot accurately estimate the true concentration of Propanol. Interestingly, Propanol is partially found as the fifth component extracted by rank-1 TGCCA. A better way to retrieve the Propanol concentration would be to have a specific way to identify unshared factors between blocks.

\subsection{Comparing best models}
\begin{figure}
    \centering
    \includegraphics[width=0.6\linewidth]{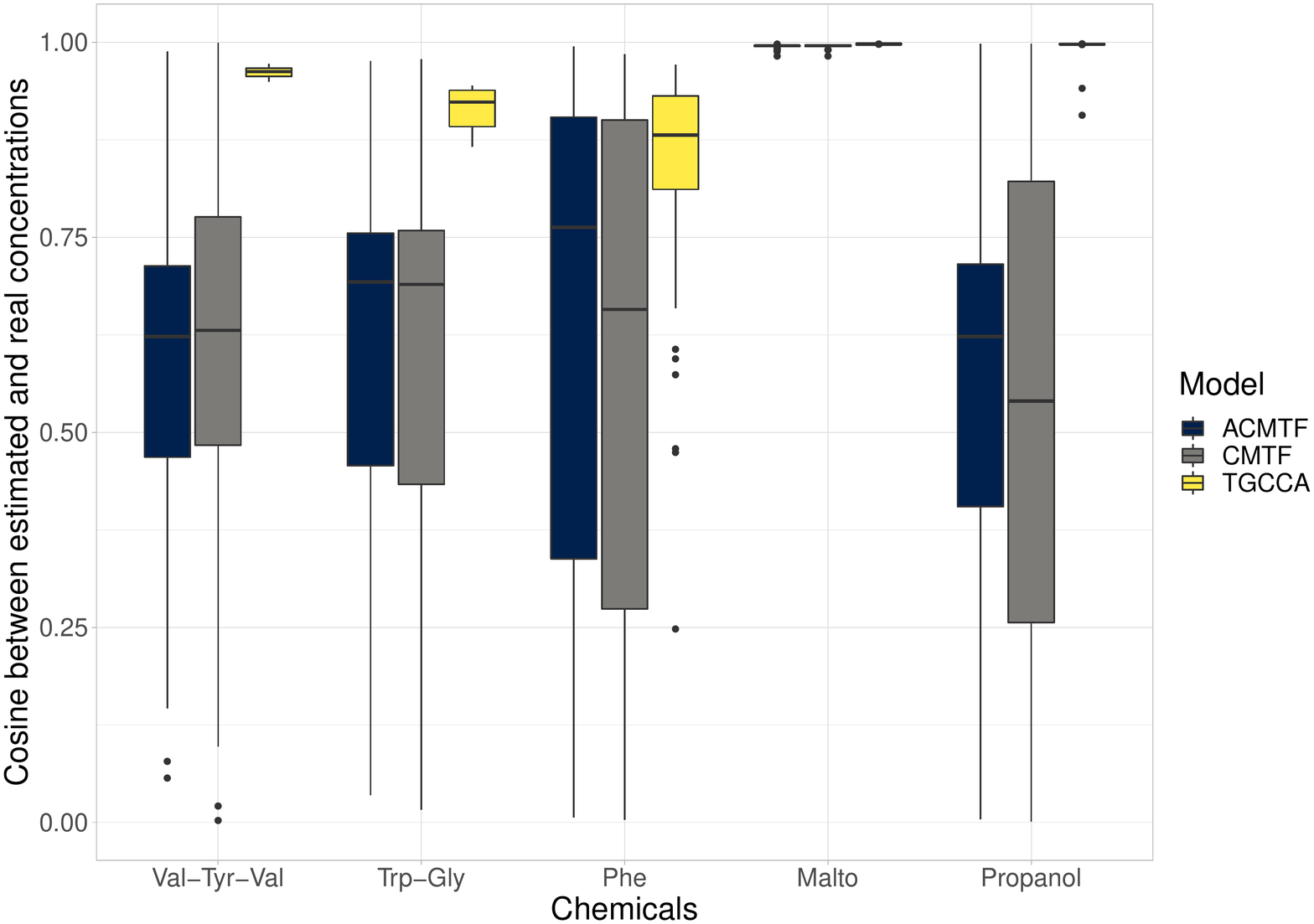}
    \caption{Boxplots of the cosines between the true concentrations and the estimated ones for TGCCA, CMTF and ACMTF.}
    \label{fig:acar-boxplots}
\end{figure}

\begin{figure}
    \centering
    \includegraphics[width=0.8\linewidth]{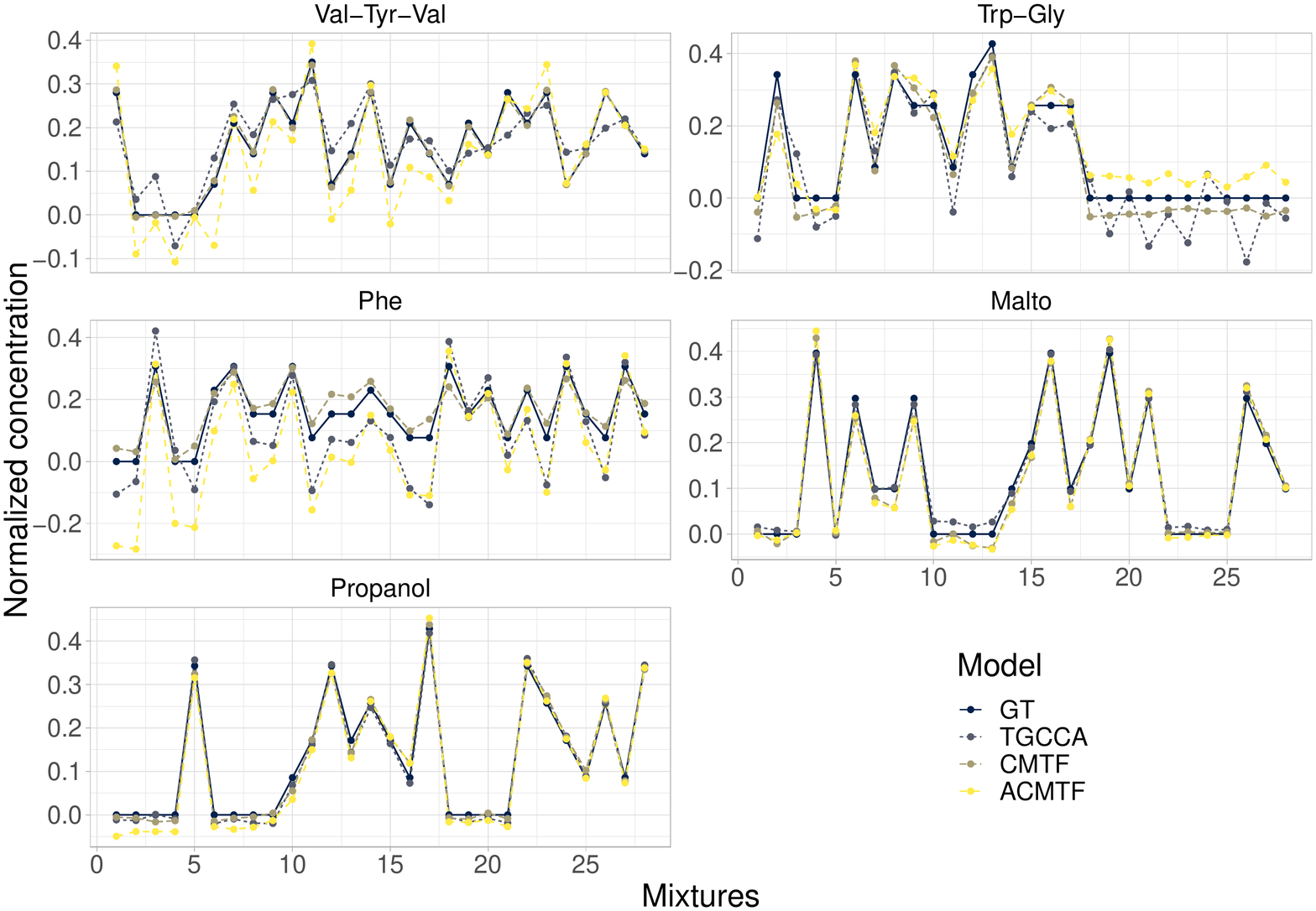}
    \caption{Normalized concentrations of the different chemicals for TGCCA, CMTF and ACMTF.}
    \label{fig:TGCCA_vs_CMTF_vs_ACMTF}
\end{figure}

As TGCCA, CMTF and ACMTF are all unsupervised methods, we select the best model for each method by keeping the one with the best criterion. For CMTF and ACMTF, we choose the model that minimizes equation \eqref{CMTF}. For TGCCA, we sum the values obtained for criterion \eqref{pop_TGCCA} for each component and keep the model with the highest sum. It is worth noting that, for both TGCCA and ACMTF, this best model does not correspond with the one that maximizes each of the cosines between the five estimated and real vectors of concentrations (see Figure \ref{fig:acar-boxplots}. We plot in Figure \ref{fig:TGCCA_vs_CMTF_vs_ACMTF} the vectors of concentrations reconstructed by the best models. We can witness that, for the columns of $\mathbf A$ found after deflation (i.e., Val-Tyr-Val, Trp-Gly, and Phe), TGCCA seems less accurate than the other methods. This advocates elaborating a global algorithm to extract the different canonical vectors jointly.

\newpage
\section{Matching faces from the Multi-Pie Face dataset}
\label{Appendix-matching faces}
\subsection{Presentation of the dataset}
\begin{figure}[bh]
    \centering
    \includegraphics[width=1.0\linewidth]{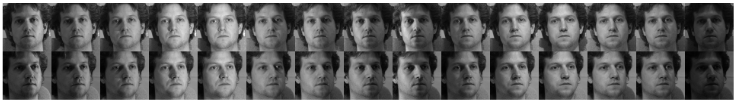}
    \caption{Images of the first subject in all illumination conditions. Each row corresponds to a different view.}
    \label{fig:subject1}
\end{figure}

The Multi-Pie Face dataset \citep{Multi-PIE} consists of images of people's faces. For each person, pictures are taken under 20 illumination conditions, 15 views, and different facial expressions. We take cropped images used in \cite{CR-GAN}, available on their \href{https://github.com/bluer555/CR-GAN/blob/master/README.md}{github repository}. This extraction consists of color images of size 128 $\times$ 128 from 250 subjects in two facial expressions (neutral and smile). We select the first 100 subjects to form our training set and the next 100 for the testing set. We use grayscale versions of the images and downsample them to size 64 $\times$ 64 using linear interpolation with the R package imager \citep{imager}. We select two views corresponding to cameras $05\_1$ and $05\_0$, which are positioned at angles 0° and -15° around the subject. We arbitrarily select 15 illumination conditions (2 to 6 and 10 to 19) and the neutral facial expression. The resulting images for the first subject are shown in Figure \ref{fig:subject1}. For each pose, we stack the images to make a tensor of dimensions 100 $\times$ 64 $\times$ 64 $\times$ 15.

\subsection{Pairing subjects}
Our goal is to use CCA methods to learn a common latent subspace between the two tensor blocks and use this learned representation in a classification task: given new subjects in the two views, pair the subjects across the two views. We compare RGCCA and spTGCCA with ranks 1 and 3. As the number of variables is much greater than the number of subjects, we use the identity matrix as the regularization matrix $\mathbf M_l$ in the RGCCA framework. The different canonical components are extracted using the deflation procedure for orthogonal components described in Section \ref{Appendix:deflation}. 

Once this subspace has been learned, it can be used to project new images. The projection is obtained by applying the preprocessing (centering and uniform scaling) used on the training set to the testing set and multiplying the image with the corresponding canonical vector. However, this last step is impossible since the canonical vectors have been learned using the 15 illumination conditions. Leveraging the structure of canonical vectors in TGCCA, we propose a workaround. Let $\TT{X} \in \mathbb{R}^{64 \times 64 \times 15}$ and $\TT{W} = \mathbf w_1 \circ \mathbf w_2 \circ \mathbf w_3 \in \mathbb{R}^{64 \times 64 \times 15}$,
\begin{equation}
\label{partial_projection}
    \mathbf x^\top \mathbf w = \TT{X} \times_1 \mathbf w_1 \times_2 \mathbf w_2 \times_3 \mathbf w_3
    = \text{Vec}\left( \TT{X} \times_1 \mathbf w_1 \times_2 \mathbf w_2 \right)^\top \mathbf w_3,
\end{equation}
where $\times_m$ denotes the mode-$m$ product.
If only slice $j$ is available on the third mode of $\TT{X}$, only the product $\text{Vec}\left( \TT{X} \times_1 \mathbf w_1 \times_2 \mathbf w_2 \right)_j w_{3,j}$ can be computed. This creates a partial projection equivalent to the full projection if the other tensor slices are filled with zeros. As training images have been centered, this zero-imputation reduces to imputing the missing slices to the means of the training subjects. If more slices are available (i.e., images of the same subjects are given in more than one illumination condition), \eqref{partial_projection} shows that they can just be added to the tensor while setting missing slices to zero. This reasoning works as-is for a rank-R tensor $\TT{W}$, so it can be applied even if the tensor rank is possibly high such as in the case of the folded version of a canonical vector obtained with RGCCA. 

For both projecting and centering the testing images, it is necessary to know in which illumination conditions the images are. We assume these illumination conditions are unknown and must be inferred from the data using a Linear Discriminant Analysis (LDA) classifier. This classifier is trained on the $100 \times 15 \times 2 = 3000$ images of the training set that were downsampled to size 16 $\times$ 16 using linear interpolation. This downsampling allows having more images than variables while leaving enough information to predict the illumination condition from the image. Cross-validation on the training set showed that gathering images from the two views was more interesting than training two classifiers.

As projections are partial when images are not present in all illumination conditions, we investigate the impact of the number of available illumination conditions by varying this number from 1 to 15. Therefore, we create 15 classification tasks where we aim to pair subjects across views. Each subject in each view is represented by a tensor of dimensions $64 \times 64 \times i$ with $i \in [15]$. The first step is to predict the illumination condition of each image by downsampling it and applying the LDA classifier. Then, the missing slices of the tensor are zero-imputed. Finally, the completed tensors are projected using the learned canonical vectors.

The pairing is then done based on the distances between the projections of the subjects in each view. As in \cite{10.5555/2540128.2540346}, we tried the $\ell_1$ and $\ell_2$ norms and the opposite of the cosine. Cross-validation on the training set showed that the cosine worked best for our task. Finding the best pairing is equivalent to finding the assignment that solves:
\begin{equation*}
    \underset{\mathbf P \in \{0, 1\}^{n \times n}}{\text{maximize}} \sum_{i,j = 1}^n p_{ij} d_{ij} \quad \text{s.t.} \quad \left\{
                \begin{array}{ll}
                    \forall i \in [n], \sum_{j = 1}^n p_{ij} = 1, \\
                    \forall j \in [n], \sum_{i = 1}^n p_{ij} = 1, 
                \end{array}
            \right.
\end{equation*}
where $d_{ij} = - \frac{\mathbf y_{1i}^\top \mathbf y_{2j}}{\Vert \mathbf y_{1i} \Vert_2 \Vert \mathbf y_{2j} \Vert_2}$, and $y_{li}$ is the projection of subject $i$ from view $l$. This problem can be efficiently solved using Integer Linear Programming. Finally, the accuracy of the matching can be measured and reported.

\subsection{Results}
Since the illumination conditions are randomly sampled, we repeat the experiments 100 times to get a better sense of the matching accuracy based on the latent subspaces learned by the different models. The results are shown in Figure \ref{fig:matching_accuracy}. Rank-3 TGCCA seems to be the best-performing method, followed by rank-1 TGCCA and RGCCA. While RGCCA learns "eigenfaces", TGCCA models remain more abstract and focus on face locations with higher variations (see Figure \ref{fig:canonical_vectors}). 

Figure \ref{fig:matching_accuracy} shows some dispersion even when the number of available illumination conditions is 15. This is explained by the fact that the classifier is applied to all test images at once without trying to predict 15 different illumination conditions for each subject. As a consequence, if an illumination condition is predicted twice, only one of the images will be used in the subject tensor, and the missing predictions will be set to zero. The variation is then due to the order in which the 15 illumination conditions are sampled.

\begin{figure}
\centering
\begin{subfigure}{1\textwidth}
  \centering
  \includegraphics[width=1\linewidth]{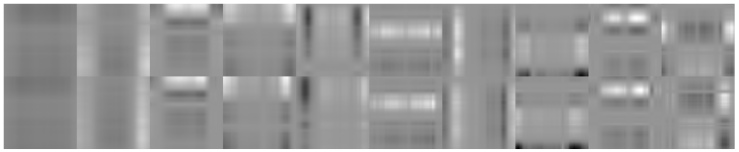}
  \caption{Rank-1 TGCCA}
\end{subfigure}
\begin{subfigure}{1\textwidth}
  \centering
  \includegraphics[width=1\linewidth]{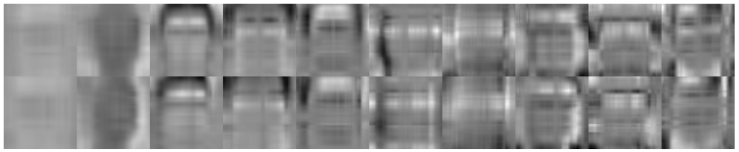}
  \caption{Rank-3 TGCCA}
\end{subfigure}
\begin{subfigure}{1\textwidth}
  \centering
  \includegraphics[width=1\linewidth]{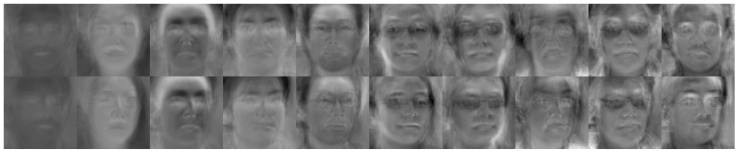}
  \caption{RGCCA}
\end{subfigure}
\caption{First 10 pairs of canonical vectors obtained with the three methods. Canonical vectors have been folded to shape $64 \times 64 \times 15$ and averaged over the last mode. Each row corresponds to a different view.}
\label{fig:eigenfaces}
\end{figure}

\newpage
\section{Simulations}
\label{simulation-appendix}
In this section, we present the data model used in the simulations presented in Section \ref{Simulations}, we detail the parameters of the numerical experiments and give more results with different numbers of samples, different levels of signal-to-noise ratio (SNR), and different number of blocks.

\subsection{Data model}
To evaluate the quality of the estimates provided by TGCCA, we extend the probabilistic TCCA model described in \cite{Min}. 
Let $\rho_{lk}$ for $l,k \in [L]$ be the pairwise correlations between blocks and $\mathbf{w}_l$ be given canonical vectors. We define:
\begin{itemize}
    \item the block covariance matrix $\boldsymbol{\Sigma}_{ll}^\eta$ as:
\begin{align}
\label{data_model}
    \boldsymbol{\Sigma}_{ll}^\eta = \mathbf{S}_l + \frac{\Vert \mathbf{S}_l \Vert_F}{\eta \Vert\mathbf{E}_l\Vert_F} \mathbf{E}_l, \quad
    \text{ with } \quad \mathbf{S}_l = \frac{\mathbf{w}_l \mathbf{w}_l^\top}{\Vert \mathbf{w}_l \Vert_2^4} \quad \text{and} \quad \mathbf{E}_l = \mathbf{P}_l \mathbf{T}_l \mathbf{T}_l^\top \mathbf{P}_l,
\end{align}
where $\mathbf{T}_l$ is a $p_l \times p_l$ arbitrary matrix, enabling noising $\mathbf{x}_l$; and $\mathbf{P}_l = \mathbf{I}_{p_l} - \frac{\mathbf{w}_l \mathbf{w}_l^\top}{\Vert \mathbf{w}_l \Vert_2^2}$ is the projector onto the orthogonal of span$(\mathbf{w}_l)$, ensuring that $\mathbf{w}_l^\top \boldsymbol{\Sigma}_{ll}^\eta \mathbf{w}_l = 1$. The SNR is controlled by the parameter $\eta$.
\item a linear transformation $\mathbf{a}_l = \rho_l\boldsymbol{\Sigma}_{ll}^\eta \mathbf{w}_l$, where $\rho_l \in [-1, 1]$ with $\rho_{lk} = \rho_l \rho_k$.
\end{itemize}

The simulated data is generated using the following latent factor model:
\begin{equation*}
    \mathbf{x}_l|z \sim \mathcal{N}(\mathbf{a}_l z, \boldsymbol{\Sigma}_{ll}^\eta - \mathbf{a}_l \mathbf{a}_l^\top) \quad \text{with} \quad z \sim \mathcal{N}(0, 1).
\end{equation*}
This allows the joint distribution of $(\mathbf{x}_1, \dots, \mathbf{x}_L)$ to be $\mathcal{N}(0, \boldsymbol{\Sigma}^\eta)$ where $\boldsymbol{\Sigma}^\eta = \begin{bmatrix} \boldsymbol{\Sigma}_{lk}^\eta
\end{bmatrix}_{\{l,k \in [L]\}}$
and $\boldsymbol{\Sigma}_{lk}^\eta = \boldsymbol{\Sigma}_{ll}^\eta \mathbf{w}_l \rho_{lk} \mathbf{w}_k^\top \boldsymbol{\Sigma}_{kk}^\eta$. Thanks to this model, the blocks are correlated through the linear transformation of the latent variable $z$.

\subsection{Data generation}
\begin{figure}[b]
\centering
\begin{subfigure}{0.2\columnwidth}
  \centering
  \includegraphics[width=1\linewidth]{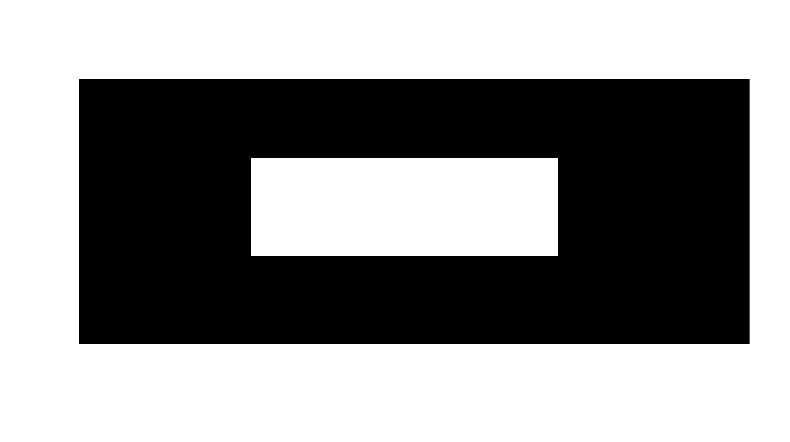}
  \caption{Square}
  \label{fig:sub1}
\end{subfigure}%
\begin{subfigure}{0.2\columnwidth}
  \centering
  \includegraphics[width=1\linewidth]{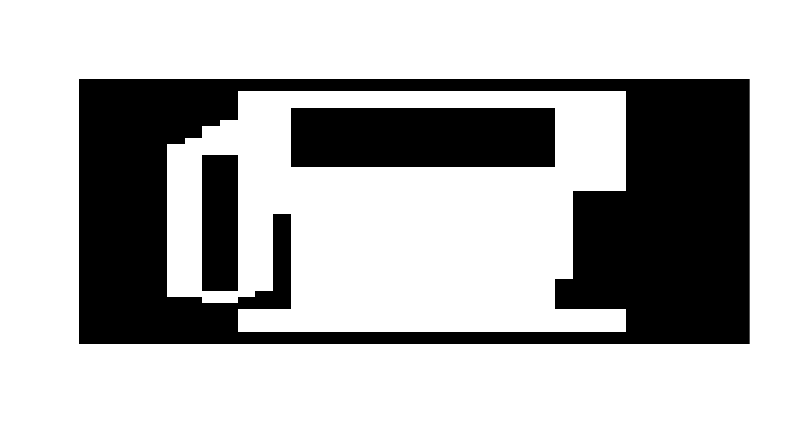}
  \caption{Gas}
  \label{fig:sub2}
\end{subfigure}%
\begin{subfigure}{0.2\columnwidth}
  \centering
  \includegraphics[width=1\linewidth]{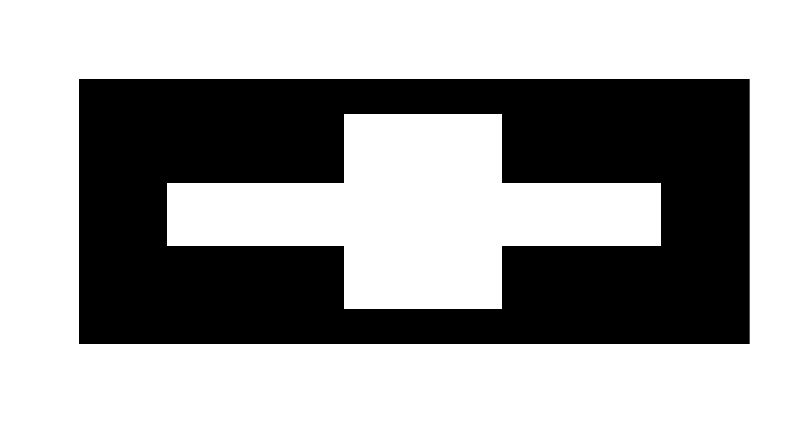}
  \caption{Cross}
  \label{fig:sub3}
\end{subfigure}%
\begin{subfigure}{0.2\columnwidth}
  \centering
  \includegraphics[width=1\linewidth]{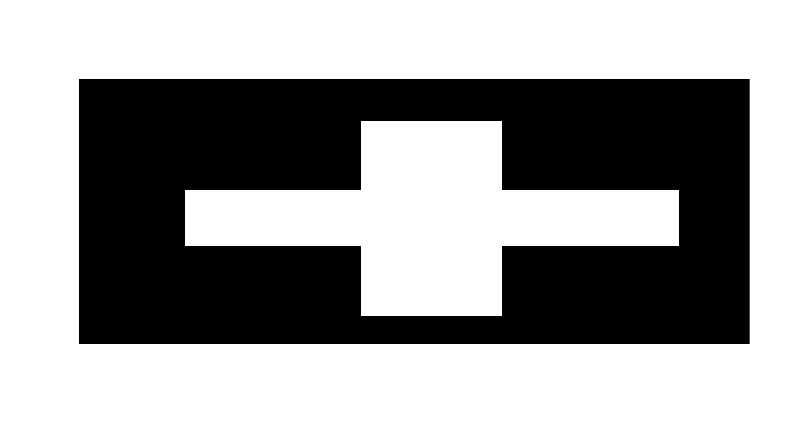}
  \caption{Cross (small)}
  \label{fig:sub4}
\end{subfigure}
\caption{Folded canonical vectors used to generate the data.}
\label{fig:canonical_vectors}
\end{figure}

In our numerical study, we generate data with $L = 5$ blocks. Information about the different blocks can be found in Table \ref{tab:L5blocks}. The folded shapes of the first 4 canonical vectors can be seen in Figure \ref{fig:canonical_vectors}. 10 folds of data are generated with $n = 1000$ samples per fold. $\rho_l = \sqrt{0.8}$ for $l \in [L]$ so every $\rho_{lk} = 0.8$. 

\begin{table}[b]
\caption{Description of the generated blocks.}
    \label{tab:L5blocks}
    \centering
    \resizebox{\columnwidth}{!}{%
\begin{tabular}{llllllll} 
\toprule 
Block & Name & Structure & Folded shape & Rank & Noise name & Noise rank & Used in $L = 2$ setting? \\ 
\midrule
1 & Square & matrix & 30 $\times$ 35 & 1 & Information & 8 & $\times$ \\
2 & Gas & matrix & 45 $\times$ 38 & 12 & Parking & 11 & $\checkmark$ \\
3 & Cross & matrix & 38 $\times$ 38 & 2 & Restaurant & 9 & $\times$ \\
4 & Cross (small) & matrix & 19 $\times$ 19 & 2 & Cup & 6 & $\checkmark$ \\
5 & Vector & vector & 100 & NA & NA & NA & $\times$ \\
\bottomrule
\end{tabular}
    }%
\end{table}

\begin{figure}
\centering
\begin{subfigure}{.2\textwidth}
  \centering
  \includegraphics[width=1\linewidth]{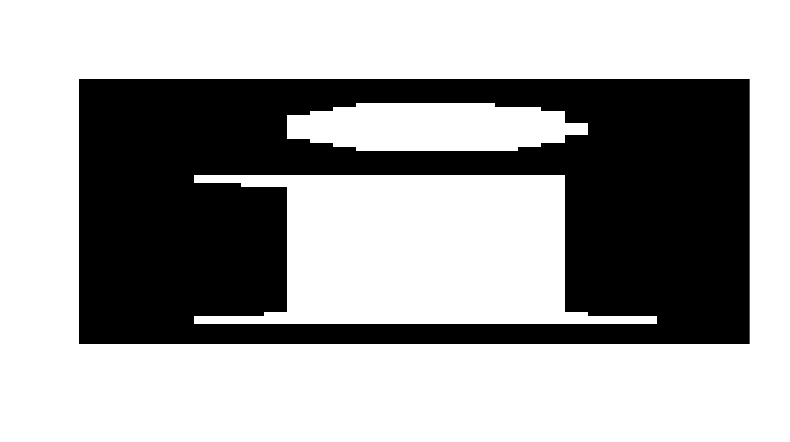}
  \caption{Information}
\end{subfigure}%
\begin{subfigure}{.2\textwidth}
  \centering
  \includegraphics[width=1\linewidth]{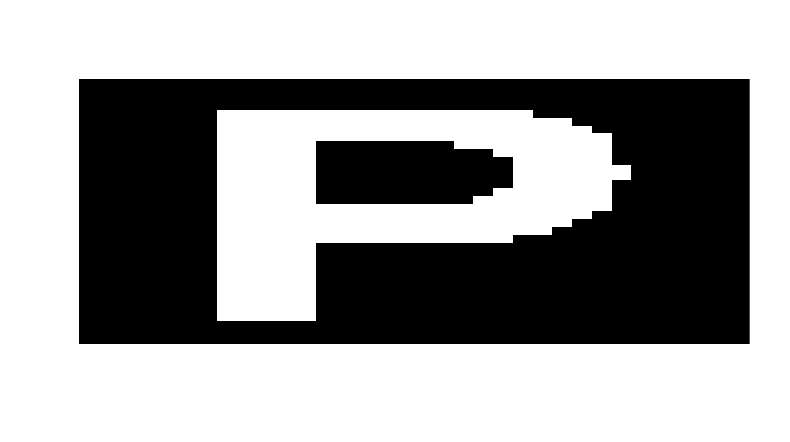}
  \caption{Parking}
\end{subfigure}%
\begin{subfigure}{.2\textwidth}
  \centering
  \includegraphics[width=1\linewidth]{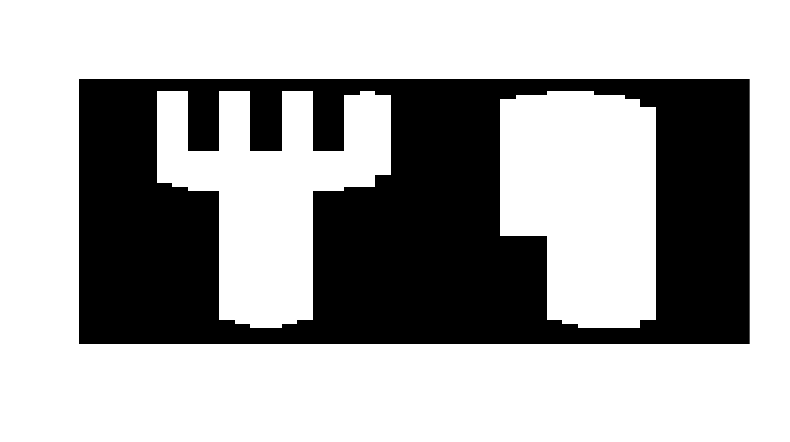}
  \caption{Restaurant}
\end{subfigure}%
\begin{subfigure}{.2\textwidth}
  \centering
  \includegraphics[width=1\linewidth]{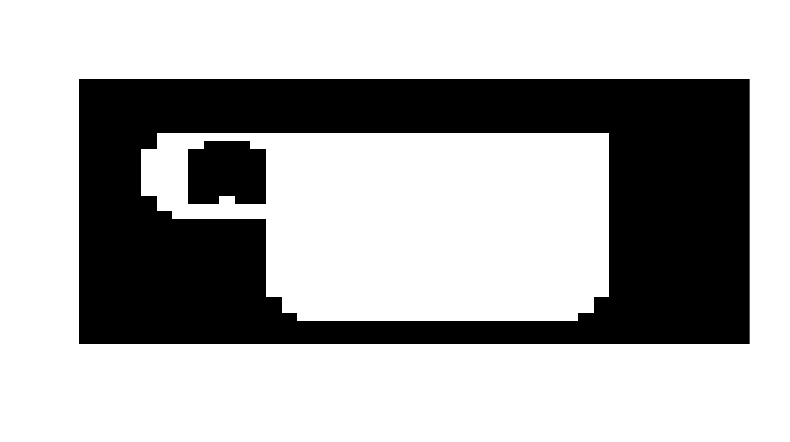}
  \caption{Cup}
\end{subfigure}
\caption{Folded shapes used to generate the structured noise.}
\label{fig:noise_shapes}
\end{figure}

Noise is added using the model described in \eqref{data_model}. For block $l$, $\mathbf{E}_l = \mathbf{P}_l \mathbf{T}_l \mathbf{T}_l^\top \mathbf{P}_l$, with $\mathbf{T}_l$ an arbitrary matrix in $\mathbb{R}^{p_l \times p_l}$. In order for $\boldsymbol{\Sigma}_{ll}^\eta$ to be positive-definite, $\mathbf{T}_l \mathbf{T}_l^\top$ has to be positive-definite.
We choose $\mathbf{T}_l$ such that $\mathbf{T}_l \mathbf{T}_l^\top = \mathbf{T}_l^{u} \mathbf{T}_l^{u^\top} + \mathbf{t}_l^{s} \mathbf{t}_l^{s^\top}$, where the first term defines unstructured noise and the second, structured one. These terms are defined as follows:
\begin{itemize}
    \item Unstructured noise is generated by sampling independent random normal variables and organizing them in a lower triangular matrix $\mathbf T_l^{u}$ of size $p_l \times p_l$. Therefore, using the Cholesky decomposition, the matrix $\mathbf{T}_l^{u} \mathbf{T}_l^{u^\top}$ is a symmetric positive-definite matrix.
    \item Structured noise is added using 2D shapes shown in Figure \ref{fig:noise_shapes} and described in Table \ref{tab:L5blocks}. 
    To create noise from these shapes, we just vectorize them in vectors $\mathbf{t}_l^s$. 
\end{itemize}
Unstructured noise is added to every block, but no structured noise is added to the last block ("Vector"). Both types of noises are normalized by their Frobenius norms before being added together. Data has been generated for 4 levels of SNR: -20dB, -10.5dB, -6dB and 0dB ($\eta$ equals respectively 0.1, 0.3, 0.5 and 1). 

It is possible to split the folds from the 10 folds of $n = 1000$ samples each to generate different experiments. 
In this section, some results are given for 
\begin{itemize}
    \item 10 folds with $n = 1000$ samples per fold,
    \item 20 folds with $n = 500$ samples per fold,
    \item 33 folds with $n = 300$ samples per fold,
    \item 50 folds with $n = 200$ samples per fold,
    \item 100 folds with $n = 100$ samples per fold.
\end{itemize}

Results in the main text are presented for $n = 1000$, a SNR level of -20dB, and a selection of 2 blocks ("Gas" and "Cross (small)") among the 5 that were created, resulting in $L = 2$.

\subsection{Methods}
The models included in the comparison are, in the $L = 2$ settings, TGCCA, MGCCA \citep{Gloaguen2020}, TCCA \citep{Min}, 2DCCA \citep{Chen}, RGCCA \citep{Tenenhaus2017} and the per block SVD. In the $L = 5$ settings, only TGCCA, MGCCA, RGCCA and the per block SVD are included. If relevant, the method's rank is added as a suffix and the separable assumption as a prefix with the letters "sp". 

We apply small changes to the codes of \cite{Chen} and \cite{Min} to harmonize the experiments. We add the convergence criterion from the latter to the former to compare computation times, and we set the shrinkage parameter as a parameter of TCCA to have the same shrinkage parameter for all models. 
The shrinkage parameter $\tau$ is set to 0.001. 
As \cite{Chen} proposes a so-called "effective" initialization strategy, we use it to run 2DCCA1. TGCCA, MGCCA and TCCA are run with 5 different starts. 

We use TGCCA and MGCCA in a CCA settings, i.e.:
\begin{itemize}
    \item The function $\text{g}$ is the identity function,
    \item The elements of the design matrix $\mathbf{C}$ are $c_{lk} = 1 - \delta_{lk}$ where $\delta$ is the Kronecker delta,
    \item For the constraint matrix $\mathbf{M}_l$, two cases are considered depending on the separable assumption made on these matrices (only the second case applies for MGCCA):
    \begin{itemize}
        \item When $\mathbf{M}_l$ is not assumed to be separable, $\hat{\mathbf{M}}_l = \hat{\mathbf{\Sigma}}_{ll} + \tau_l\mathbf{I}_{p_l}$ which is a regularized version of the empirical covariance $\hat{\mathbf{\Sigma}}_{ll}$. We choose $\tau_l = 0.001$ for $l \in [L]$.
        \item When $\mathbf{M}_l$ is assumed to be separable, a separable estimate of the covariance \citep{Hoff2011, Min} is used. Without going into too much details, this estimator can be written as $\hat{\mathbf{\Sigma}}_{ll} = \hat{\mathbf{\Sigma}}_{ll,d_l} \otimes \dots \otimes \hat{\mathbf{\Sigma}}_{ll,1}$. We propose here a regularized version of it, where $\hat{\mathbf{M}}_l = \left(\hat{\mathbf{\Sigma}}_{ll,d_l} + \sqrt[d_l]{\tau_l}\mathbf{I}_{p_{l, d_l}}\right) \otimes \dots \otimes \left(\hat{\mathbf{\Sigma}}_{ll,1} + \sqrt[d_l]{\tau_l}\mathbf{I}_{p_{l, 1}}\right)$. Here, the regularization term is multiplied by $\sqrt[d_l]{\tau_l}$ instead of $\tau_l$ so that, when the Kronecker products are developed, the term in front of $\mathbf{I}_{p_l}$ is $\tau_l$, which is a way to have a similar level of regularization between the separable and non-separable cases. As before, for all blocks, $\tau_l$ is set to $0.001$.
    \end{itemize}
\end{itemize}
Concerning the normalization procedure, for all the methods, variables of each block were centered and scaled by $s_l = \sqrt{\frac{p_l}{n}} \| \mathbf{X}_l \|_F$, where $\| . \|_F$ is the Frobenius norm.

\subsection{Results}
All experiments were run on a 
personal computer using the R language \citep{R}. 

Results are given in  \crefrange{tab:2B_1000_all}{tab:2B_vs_5B_100_all}. Cosines (with median and 2.5\% and 97.5\% quantiles over the different folds) between the canonical vectors used to generate the data and the estimated ones are reported. The computation time (with median and 2.5\% and 97.5\% quantiles over the different folds) is reported in seconds. For models run with multiple starts (TCCA, MGCCA and TGCCA), the computation time includes the 5 runs.

From \crefrange{tab:2B_1000_all}{tab:2B_100_all}, results are shown for the $L = 2$ settings and compare 2DCCA, TCCA, MGCCA, TGCCA, RGCCA and per-block SVD. From \crefrange{tab:5B_1000_all}{tab:5B_100_all}, results are shown for the $L = 5$ settings and compare MGCCA, TGCCA, RGCCA and per-block SVD. Finally, from \crefrange{tab:2B_vs_5B_1000_all}{tab:2B_vs_5B_100_all} a comparison between the $L = 2$ and $L = 5$ settings is proposed. Each of the 15 tables is split into 4 smaller tables, one per SNR. Among one group of 5 tables (\crefrange{tab:2B_1000_all}{tab:2B_100_all}, \crefrange{tab:5B_1000_all}{tab:5B_100_all} and \crefrange{tab:2B_vs_5B_1000_all}{tab:2B_vs_5B_100_all}), the number of folds is increasing and thus the number of samples per fold is decreasing.

Firstly, for all models but 2DCCA3, the accuracy increases with the SNR and the number of samples per fold. 
2DCCA3 totally fails to retrieve the canonical vectors. We think that this is due to our experimental settings. Indeed, 2DCCA3 tries to find 3 canonical vectors of rank 1 such that the canonical components $y_l = \mathbf{w}_l^\top \mathbf{x}_l$ are uncorrelated while the data is simulated from only one canonical component per block with the associated canonical vectors of ranks greater than 1. This illustrates the differences between rank and number of components discussed in Section \ref{rank-vs-comp}.

We can see that the accuracy is better with a much lower standard deviation when the SNR is greater than -20dB. When the SNR is low (-20dB), the different methods are more sensitive to the choice of the starting point. To highlight this last point, an additional experiment was held with a SNR of -20dB for MGCCA, TCCA and TGCCA. This time, instead of keeping the results associated with the best random initialization only, we display the median and 2.5\% and 97.5\% quantiles for each block ($L = 2$) through 100 random starts on a given fold (results are reported in Table \ref{tab:init}). Even if the median coincides with the higher quantile, the lower quantile is extremely low. On the other hand, the "effective" strategy of 2DCCA1 does not always lead to a good initial point either (see, for example, the high standard deviation reported in Table \ref{tab:2B_1000_all} for 2DCCA1). We do not provide guidelines for choosing a good initial point. Still, if possible, we advise running MGCCA, TCCA and TGCCA multiple times with initial points chosen randomly and keeping the models with the highest correlation between blocks.

\begin{table}[b]
  \caption{Cosine between the true canonical vectors and the estimated ones for different models on block "Gas" and "Cross (little)" for SNR of -20dB for n = 1000, fold 1, with 100 random initial points. Median and quantiles (2.5\% and 97.5\%) are reported. The medians coincide with the higher quantile but the lower quantile is very low.}
    \label{tab:init}
  \centering
\begin{tabular}{llll} 
\toprule 
Model & Gas & Cross (small) \\ 
\midrule 
TCCA1 & 0.89 (0.00, 0.89) & 0.86 (0.23, 0.86) \\ 
TGCCA1 & 0.89 (0.00, 0.89) & 0.86 (0.23, 0.86) \\ 
spTCCA1 & 0.89 (0.00, 0.89) & 0.86 (0.23, 0.86) \\ 
MGCCA & 0.89 (0.00, 0.89) & 0.86 (0.23, 0.86) \\ 
TCCA3 & 0.89 (0.00, 0.89) & 0.86 (0.23, 0.86) \\ 
TGCCA3 & 0.95 (0.01, 0.95) & 0.96 (0.08, 0.96) \\ 
spTCCA3 & 0.89 (0.00, 0.89) & 0.86 (0.23, 0.87) \\ 
spTGCCA3 & 0.94 (0.01, 0.94) & 0.95 (0.07, 0.95) \\ 
\bottomrule 
\end{tabular}
\end{table}

\begin{figure}
    \centering
    \includegraphics[width=0.7\linewidth]{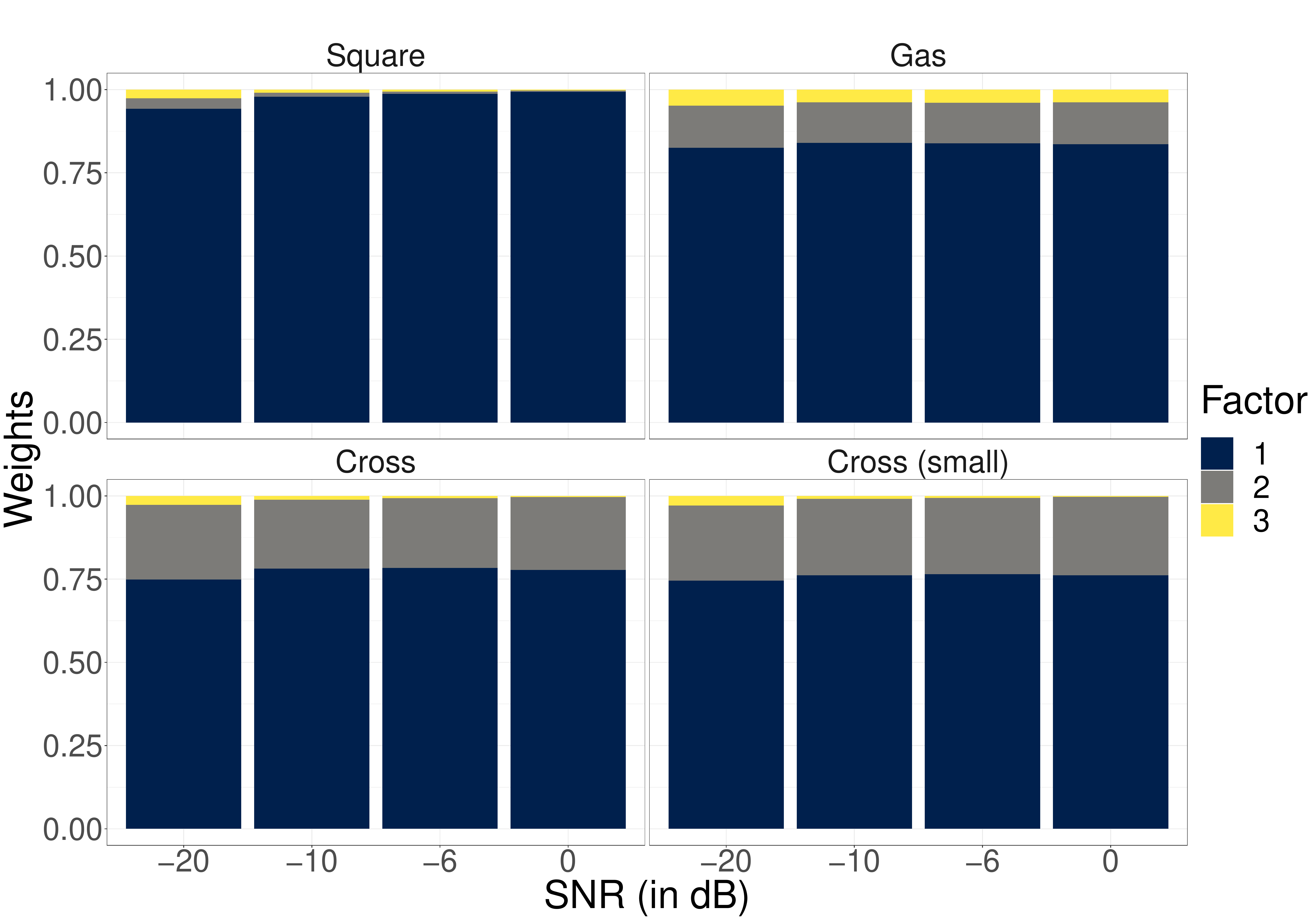}
    \caption{Contributions ($\lambda_l$) of the different rank-1 factors for spTGCC3 for $n = 1000$ and 10 folds. We can observe that the extra factors are cancelling out for the low rank canonical vectors when the SNR increases.}
    \label{fig:contrib}
\end{figure}

TGCCA3 and spTGCCA3 perform better than rank-1 models when the SNR is high. It is expected as the rank of the underlying canonical vectors is greater than 1 for every block except for the first block in the $L = 5$ settings. 
The opposite trend is observed for this block, even if TGCCA3 and spTGCCA3 remain very good.
It can be explained by the fact that the weights of the different rank-1 factors are not null, but only one of them is far from zero (see Figure \ref{fig:contrib}). On the other hand, when the SNR is low, TGCCA3 and spTGCCA3 tend to perform worse than the rank-1 methods. It is also expected as rank-3 models have more degrees of freedom and are more flexible. Therefore, when the SNR is too low, rank-3 models can describe both the relevant information and the noise (see factors 2 and 3 for "Square" and factor 3 for "Cross" and "Cross (small)" on Figure \ref{fig:contrib}). 

In the $L = 5$ settings, we see that the models take profit from the
redundancy between blocks to estimate the canonical vectors more accurately. We can also point out that spTGCCA
scales well with the number of blocks and remains fast when evaluated on the 5 blocks.
To investigate the interest of analyzing more than 2 blocks jointly, we compared the same models both applied on 2 blocks and on 5 blocks (respectively denoted with suffixes "b2" and "b5"). It shows that having more correlated blocks acts like virtually increasing the SNR or the number of observations $n$. Hence the accuracy of the models applied on 5 blocks is much higher for SNR of -20dB and slightly better for higher SNR. See Tables \ref{tab:2B_vs_5B_1000_all}-\ref{tab:2B_vs_5B_100_all}. 


Finally, spTGCCA appears to be the fastest method (considering that reported computation times correspond to 5 runs). spTGCCA is faster because it needs only to work with much smaller matrices ($\hat{\mathbf{M}}_{l, m} \in \mathbb{R}^{p_{l,m} \times p_{l, m}}$), compute them once, make a change of variable and then work without regularization matrices. However, one must be cautious when comparing to TCCA because the Matlab code of TCCA was called from R, leading to some slight overestimation of the reported computation time. 

\subsection{Additional experiments with 3D canonical vectors}
\label{Appendix-4D}
\begin{figure}
\centering
\begin{subfigure}{.3\textwidth}
  \centering
  \includegraphics[width=1\linewidth]{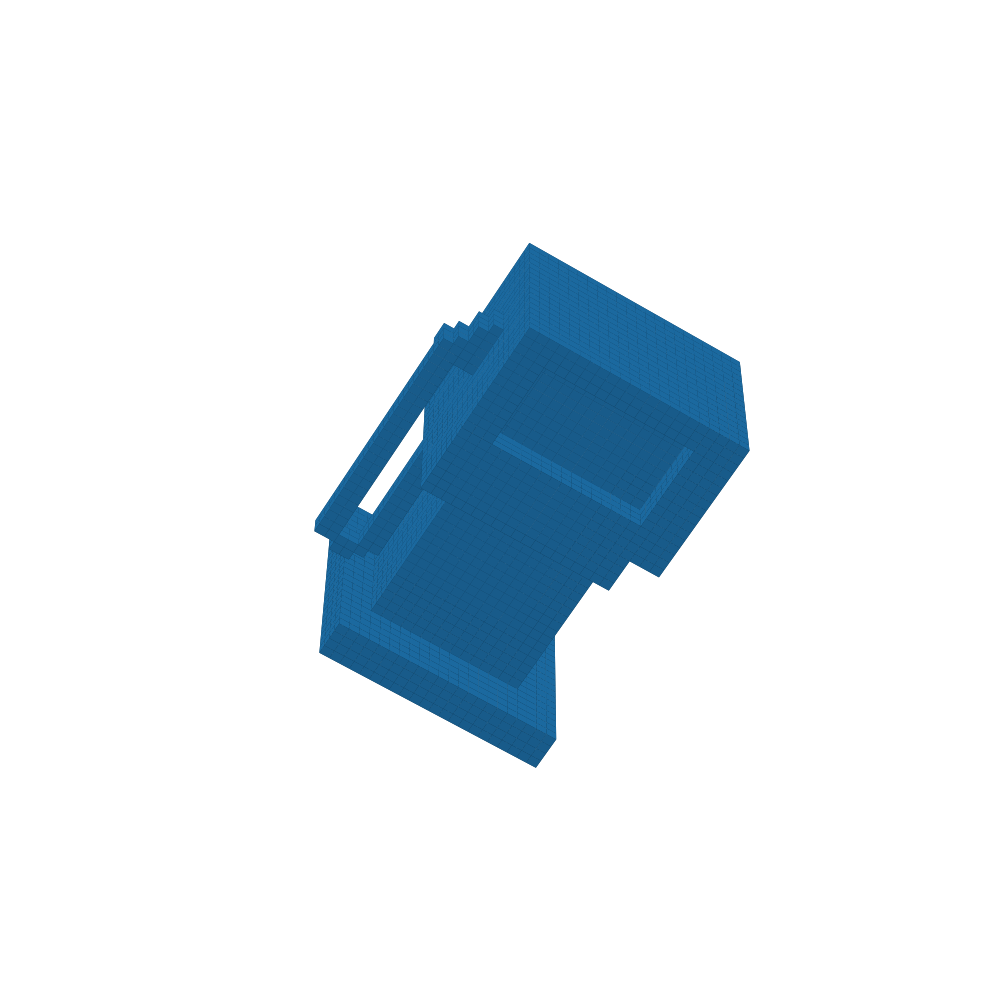}
  \caption{Gas 3D}
\end{subfigure}%
\begin{subfigure}{.3\textwidth}
  \centering
  \includegraphics[width=1\linewidth]{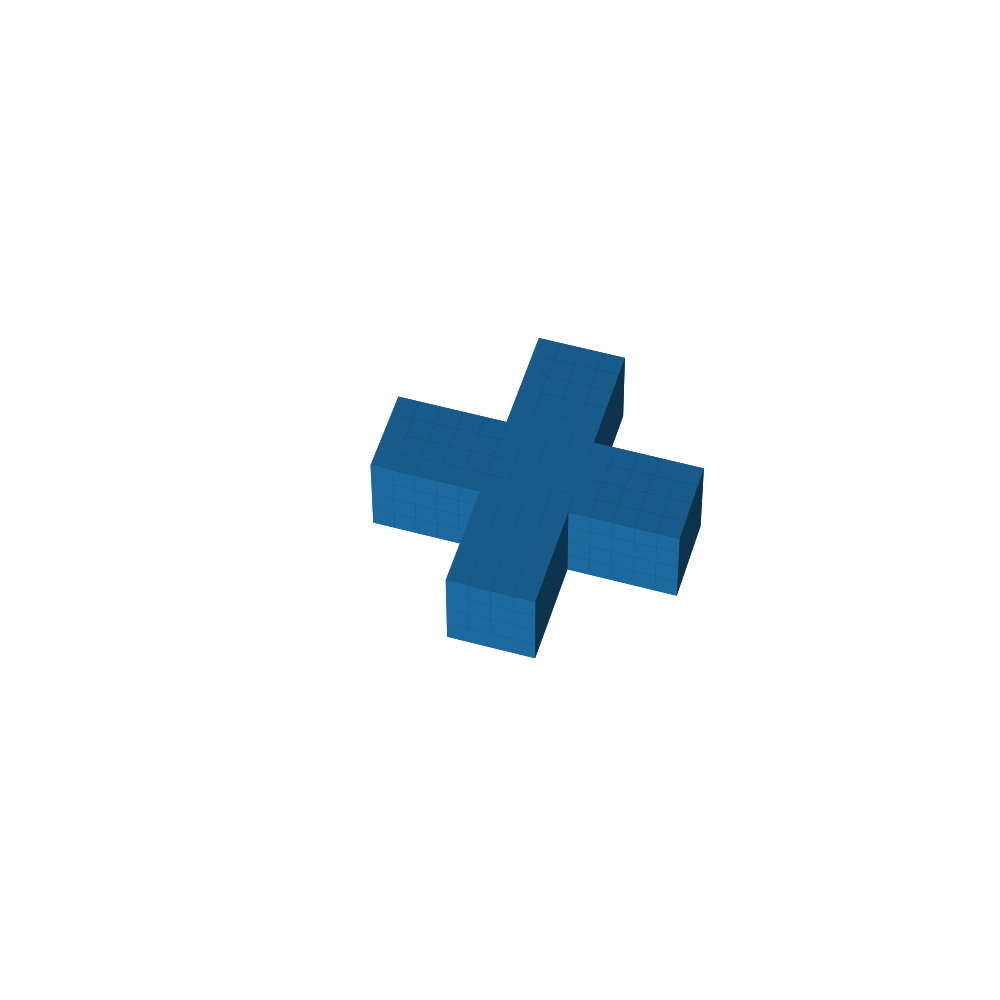}
  \caption{Cross (small) 3D}
\end{subfigure}
\caption{Folded shapes used in the 4D settings.}
\label{fig:3D_shapes}
\end{figure}
We repeat most of the previous experiments with 3D versions of the "Gas" and "Cross (small)" shapes shown in Figure \ref{fig:3D_shapes}. Shapes "Cross" and "Cross (small) 3D" are used in the $L = 2$ settings. Due to the size of the "Gas 3D" shape, the unstructured noise matrices $\mathbf T_l^u$ have been replaced with homothety matrices. Results in the main text are presented for n = 1000, and a SNR level of -10dB in the $L = 2$ settings.

2DCCA is removed from these experiments since we did not find an available implementation for higher-order tensors. As "Gas 3D" is of high dimension, the regularization matrices $\mathbf M_l$ are set to the identity matrices for RGCCA and TGCCA in the $L = 5$ settings. Therefore, only spTGCCA is used and is reported as TGCCA. Orthogonality is imposed on the first mode for TGCCA models. All models are run with 10 different starting points. \Crefrange{tab:4D-2B_1000_all}{tab:4D-2B_100_all} show the results for the $L = 2$ settings, and \crefrange{tab:4D-5B_1000_all}{tab:4D-5B_100_all} show the results for the $L = 5$ settings.

Conclusions are similar to the previous experiments but RGCCA perform much better than before. This is probably due to the unstructured noise being simpler in these new experiments.

\begin{table}
  \caption{Cosine between the true and the estimated canonical vectors for different models on blocks "Gas" and "Cross (small)", for levels of SNR -20dB, -10dB, -6dB and 0dB from top to bottom, and computation times, for n = 1000 and 10 folds. Median and quantiles (2.5\% and 97.5\%) are reported.}
    \label{tab:2B_1000_all}
  \centering
  \resizebox{0.6\columnwidth}{!}{%
\begin{tabular}{llll} 
\toprule 
Model & Gas & Cross (small) & Computation time \\ 
\midrule 
2DCCA1 & 0.30 (0.01, 0.89) & 0.43 (0.16, 0.85) & 3.09 (2.76, 4.50) \\ 
TCCA1 & 0.89 (0.22, 0.90) & 0.85 (0.32, 0.86) & 7.72 (7.38, 9.17) \\ 
TGCCA1 & 0.89 (0.87, 0.90) & 0.85 (0.83, 0.86) & 8.60 (8.36, 9.12) \\ 
spTCCA1 & 0.89 (0.22, 0.90) & 0.85 (0.32, 0.86) & 7.43 (7.24, 8.04) \\ 
MGCCA & 0.89 (0.87, 0.90) & 0.86 (0.83, 0.86) & 5.08 (4.85, 5.32) \\ 
2DCCA3 & 0.04 (0.01, 0.21) & 0.13 (0.05, 0.31) & 1.50 (1.41, 3.09) \\ 
TCCA3 & 0.89 (0.87, 0.90) & 0.85 (0.83, 0.86) & 7.93 (7.73, 8.92) \\ 
TGCCA3 & 0.91 (0.78, 0.94) & 0.92 (0.79, 0.96) & 11.32 (10.12, 15.94) \\ 
spTCCA3 & 0.89 (0.87, 0.90) & 0.85 (0.83, 0.86) & 7.32 (7.22, 7.51) \\ 
spTGCCA3 & 0.92 (0.82, 0.94) & 0.93 (0.83, 0.96) & 5.66 (5.52, 6.39) \\ 
RGCCA & 0.17 (0.05, 0.26) & 0.11 (0.06, 0.20) & 13.12 (12.67, 14.07) \\ 
SVD & 0.00 (0.00, 0.01) & 0.01 (0.00, 0.03) & 5.78 (5.44, 6.07) \\ 
\bottomrule 
\end{tabular}

  }
  \resizebox{0.6\columnwidth}{!}{%
  \begin{tabular}{llll} 
\toprule 
Model & Gas & Cross (small) & Computation time \\ 
\midrule 
2DCCA1 & 0.90 (0.89, 0.91) & 0.88 (0.87, 0.88) & 2.89 (2.60, 4.57) \\ 
TCCA1 & 0.90 (0.89, 0.91) & 0.88 (0.87, 0.88) & 7.42 (7.38, 7.52) \\ 
TGCCA1 & 0.90 (0.89, 0.91) & 0.88 (0.87, 0.88) & 8.17 (8.02, 8.53) \\ 
spTCCA1 & 0.90 (0.89, 0.91) & 0.88 (0.87, 0.88) & 7.19 (7.11, 7.33) \\ 
MGCCA & 0.90 (0.89, 0.91) & 0.88 (0.87, 0.88) & 4.84 (4.68, 5.35) \\ 
2DCCA3 & 0.13 (0.05, 0.23) & 0.16 (0.03, 0.36) & 1.51 (1.41, 3.07) \\ 
TCCA3 & 0.90 (0.89, 0.91) & 0.88 (0.87, 0.88) & 7.61 (7.58, 7.68) \\ 
TGCCA3 & 0.97 (0.96, 0.98) & 0.99 (0.98, 0.99) & 9.65 (9.23, 10.28) \\ 
spTCCA3 & 0.90 (0.89, 0.91) & 0.88 (0.87, 0.88) & 7.34 (7.13, 7.89) \\ 
spTGCCA3 & 0.97 (0.96, 0.98) & 0.99 (0.98, 0.99) & 5.10 (4.90, 5.48) \\ 
RGCCA & 0.93 (0.91, 0.94) & 0.97 (0.96, 0.98) & 11.97 (11.38, 12.74) \\ 
SVD & 0.02 (0.01, 0.06) & 0.01 (0.00, 0.04) & 5.45 (5.16, 5.94) \\ 
\bottomrule 
\end{tabular}

  }
  \resizebox{0.6\columnwidth}{!}{%
  \begin{tabular}{llll} 
\toprule 
Model & Gas & Cross (small) & Computation time \\ 
\midrule 
2DCCA1 & 0.90 (0.90, 0.91) & 0.88 (0.87, 0.88) & 3.03 (2.81, 4.57) \\ 
TCCA1 & 0.90 (0.90, 0.91) & 0.88 (0.87, 0.88) & 7.40 (7.27, 7.49) \\ 
TGCCA1 & 0.90 (0.90, 0.91) & 0.88 (0.87, 0.88) & 8.07 (7.97, 8.25) \\ 
spTCCA1 & 0.90 (0.90, 0.91) & 0.88 (0.87, 0.88) & 7.19 (7.14, 7.35) \\ 
MGCCA & 0.90 (0.90, 0.91) & 0.88 (0.87, 0.88) & 4.87 (4.56, 5.19) \\ 
2DCCA3 & 0.12 (0.03, 0.28) & 0.06 (0.02, 0.34) & 1.51 (1.40, 3.00) \\ 
TCCA3 & 0.90 (0.90, 0.91) & 0.88 (0.87, 0.88) & 7.67 (7.55, 7.84) \\ 
TGCCA3 & 0.98 (0.97, 0.98) & 0.99 (0.98, 0.99) & 9.51 (9.25, 10.76) \\ 
spTCCA3 & 0.90 (0.90, 0.91) & 0.88 (0.87, 0.88) & 7.42 (7.22, 7.69) \\ 
spTGCCA3 & 0.98 (0.97, 0.98) & 0.99 (0.98, 0.99) & 5.03 (4.89, 5.35) \\ 
RGCCA & 0.95 (0.95, 0.96) & 0.98 (0.97, 0.98) & 11.78 (11.13, 12.13) \\ 
SVD & 0.07 (0.01, 0.13) & 0.05 (0.01, 0.21) & 5.36 (5.18, 5.66) \\ 
\bottomrule 
\end{tabular}

  }
  \resizebox{0.6\columnwidth}{!}{%
  \begin{tabular}{llll} 
\toprule 
Model & Gas & Cross (small) & Computation time \\ 
\midrule 
2DCCA1 & 0.91 (0.91, 0.91) & 0.88 (0.88, 0.88) & 3.24 (2.66, 4.78) \\ 
TCCA1 & 0.91 (0.91, 0.91) & 0.88 (0.88, 0.88) & 7.40 (7.32, 7.52) \\ 
TGCCA1 & 0.91 (0.91, 0.91) & 0.88 (0.88, 0.88) & 8.12 (7.88, 8.29) \\ 
spTCCA1 & 0.91 (0.91, 0.91) & 0.88 (0.88, 0.88) & 7.22 (7.12, 7.31) \\ 
MGCCA & 0.91 (0.91, 0.91) & 0.88 (0.88, 0.88) & 4.79 (4.68, 5.11) \\ 
2DCCA3 & 0.09 (0.01, 0.21) & 0.12 (0.07, 0.22) & 1.54 (1.44, 3.24) \\ 
TCCA3 & 0.91 (0.91, 0.91) & 0.88 (0.88, 0.88) & 7.67 (7.54, 7.85) \\ 
TGCCA3 & 0.98 (0.98, 0.98) & 1.00 (0.99, 1.00) & 9.69 (9.35, 11.20) \\ 
spTCCA3 & 0.91 (0.91, 0.91) & 0.88 (0.88, 0.88) & 7.49 (7.28, 7.91) \\ 
spTGCCA3 & 0.98 (0.98, 0.98) & 0.99 (0.99, 1.00) & 4.97 (4.84, 5.33) \\ 
RGCCA & 0.98 (0.97, 0.98) & 0.99 (0.99, 0.99) & 11.56 (11.04, 12.17) \\ 
SVD & 0.99 (0.97, 0.99) & 0.99 (0.96, 1.00) & 5.32 (5.10, 5.59) \\ 
\bottomrule 
\end{tabular}

  }
\end{table}
\begin{table}
  \caption{Cosine between the true and the estimated canonical vectors for different models on blocks "Gas" and "Cross (small)", for levels of SNR -20dB, -10dB, -6dB and 0dB from top to bottom, and computation times, for n = 500 and 20 folds. Median and quantiles (2.5\% and 97.5\%) are reported.}
    \label{tab:2B_500_all}
  \centering
  \resizebox{0.6\columnwidth}{!}{%
  \begin{tabular}{llll} 
\toprule 
Model & Gas & Cross (small) & Computation time \\ 
\midrule 
2DCCA1 & 0.03 (0.00, 0.85) & 0.23 (0.05, 0.79) & 2.32 (2.17, 5.39) \\ 
TCCA1 & 0.81 (0.00, 0.89) & 0.77 (0.09, 0.86) & 8.63 (7.56, 10.45) \\ 
TGCCA1 & 0.86 (0.68, 0.89) & 0.82 (0.63, 0.86) & 3.16 (3.02, 4.50) \\ 
spTCCA1 & 0.72 (0.00, 0.89) & 0.62 (0.10, 0.86) & 7.82 (7.25, 8.84) \\ 
MGCCA & 0.86 (0.72, 0.89) & 0.82 (0.69, 0.86) & 2.81 (2.50, 3.18) \\ 
2DCCA3 & 0.03 (0.00, 0.12) & 0.06 (0.01, 0.25) & 0.95 (0.91, 2.14) \\ 
TCCA3 & 0.86 (0.36, 0.89) & 0.82 (0.43, 0.86) & 9.45 (7.81, 11.98) \\ 
TGCCA3 & 0.16 (0.01, 0.68) & 0.19 (0.01, 0.64) & 4.73 (3.49, 7.97) \\ 
spTCCA3 & 0.86 (0.14, 0.89) & 0.83 (0.31, 0.86) & 8.05 (7.19, 8.65) \\ 
spTGCCA3 & 0.20 (0.01, 0.70) & 0.21 (0.02, 0.67) & 2.79 (2.51, 3.53) \\ 
RGCCA & 0.11 (0.00, 0.22) & 0.07 (0.01, 0.15) & 3.27 (3.06, 3.57) \\ 
SVD & 0.01 (0.00, 0.03) & 0.01 (0.00, 0.04) & 1.19 (1.18, 1.28) \\ 
\bottomrule 
\end{tabular}

  }
  \resizebox{0.6\columnwidth}{!}{%
  \begin{tabular}{llll} 
\toprule 
Model & Gas & Cross (small) & Computation time \\ 
\midrule 
2DCCA1 & 0.90 (0.87, 0.90) & 0.87 (0.85, 0.88) & 2.27 (2.11, 3.49) \\ 
TCCA1 & 0.90 (0.87, 0.90) & 0.87 (0.85, 0.88) & 7.32 (7.22, 7.47) \\ 
TGCCA1 & 0.90 (0.87, 0.90) & 0.87 (0.85, 0.88) & 2.80 (2.65, 3.04) \\ 
spTCCA1 & 0.90 (0.87, 0.90) & 0.87 (0.85, 0.88) & 7.08 (6.99, 7.31) \\ 
MGCCA & 0.90 (0.87, 0.90) & 0.87 (0.85, 0.88) & 2.73 (2.49, 3.13) \\ 
2DCCA3 & 0.05 (0.01, 0.13) & 0.06 (0.01, 0.23) & 0.94 (0.91, 2.11) \\ 
TCCA3 & 0.90 (0.87, 0.90) & 0.87 (0.85, 0.88) & 7.58 (7.40, 7.87) \\ 
TGCCA3 & 0.96 (0.93, 0.96) & 0.97 (0.94, 0.98) & 3.68 (3.55, 3.95) \\ 
spTCCA3 & 0.90 (0.87, 0.90) & 0.87 (0.85, 0.88) & 7.09 (7.02, 7.31) \\ 
spTGCCA3 & 0.96 (0.92, 0.96) & 0.97 (0.95, 0.98) & 2.53 (2.45, 2.86) \\ 
RGCCA & 0.86 (0.83, 0.89) & 0.94 (0.88, 0.96) & 3.00 (2.77, 3.46) \\ 
SVD & 0.04 (0.00, 0.09) & 0.02 (0.00, 0.08) & 1.18 (1.17, 1.23) \\ 
\bottomrule 
\end{tabular}

  }
  \resizebox{0.6\columnwidth}{!}{%
  \begin{tabular}{llll} 
\toprule 
Model & Gas & Cross (small) & Computation time \\ 
\midrule 
2DCCA1 & 0.90 (0.89, 0.91) & 0.87 (0.86, 0.88) & 2.26 (2.13, 3.34) \\ 
TCCA1 & 0.90 (0.89, 0.91) & 0.87 (0.86, 0.88) & 7.26 (7.16, 7.45) \\ 
TGCCA1 & 0.90 (0.89, 0.91) & 0.87 (0.86, 0.88) & 2.82 (2.67, 3.09) \\ 
spTCCA1 & 0.90 (0.89, 0.91) & 0.87 (0.86, 0.88) & 7.07 (6.98, 7.18) \\ 
MGCCA & 0.90 (0.89, 0.91) & 0.87 (0.86, 0.88) & 2.51 (2.35, 2.91) \\ 
2DCCA3 & 0.05 (0.01, 0.16) & 0.07 (0.00, 0.25) & 0.92 (0.91, 2.06) \\ 
TCCA3 & 0.90 (0.89, 0.91) & 0.87 (0.86, 0.88) & 7.63 (7.48, 7.81) \\ 
TGCCA3 & 0.97 (0.96, 0.98) & 0.98 (0.97, 0.99) & 3.62 (3.39, 3.93) \\ 
spTCCA3 & 0.90 (0.89, 0.91) & 0.87 (0.86, 0.88) & 7.09 (7.03, 7.22) \\ 
spTGCCA3 & 0.97 (0.96, 0.97) & 0.98 (0.97, 0.98) & 2.60 (2.45, 2.93) \\ 
RGCCA & 0.91 (0.89, 0.93) & 0.96 (0.95, 0.97) & 2.67 (2.54, 2.83) \\ 
SVD & 0.09 (0.02, 0.18) & 0.10 (0.01, 0.24) & 1.27 (1.22, 1.37) \\ 
\bottomrule 
\end{tabular}

  }
  \resizebox{0.6\columnwidth}{!}{%
  \begin{tabular}{llll} 
\toprule 
Model & Gas & Cross (small) & Computation time \\ 
\midrule 
2DCCA1 & 0.91 (0.90, 0.91) & 0.88 (0.87, 0.88) & 2.53 (2.31, 3.91) \\ 
TCCA1 & 0.91 (0.90, 0.91) & 0.88 (0.87, 0.88) & 7.33 (7.22, 8.20) \\ 
TGCCA1 & 0.91 (0.90, 0.91) & 0.88 (0.87, 0.88) & 2.69 (2.56, 2.97) \\ 
spTCCA1 & 0.91 (0.90, 0.91) & 0.88 (0.87, 0.88) & 7.13 (7.06, 7.25) \\ 
MGCCA & 0.91 (0.90, 0.91) & 0.88 (0.87, 0.88) & 2.51 (2.34, 2.93) \\ 
2DCCA3 & 0.07 (0.01, 0.14) & 0.07 (0.01, 0.21) & 0.95 (0.90, 2.09) \\ 
TCCA3 & 0.91 (0.90, 0.91) & 0.88 (0.87, 0.88) & 7.63 (7.48, 7.80) \\ 
TGCCA3 & 0.98 (0.97, 0.98) & 0.99 (0.98, 0.99) & 3.58 (3.49, 3.96) \\ 
spTCCA3 & 0.91 (0.90, 0.91) & 0.88 (0.87, 0.88) & 7.11 (7.00, 7.43) \\ 
spTGCCA3 & 0.98 (0.97, 0.98) & 0.99 (0.98, 0.99) & 2.51 (2.36, 2.90) \\ 
RGCCA & 0.94 (0.93, 0.95) & 0.98 (0.97, 0.98) & 2.48 (2.41, 2.58) \\ 
SVD & 0.98 (0.95, 0.98) & 0.99 (0.91, 0.99) & 1.33 (1.28, 1.41) \\ 
\bottomrule 
\end{tabular}

  }
\end{table}
\begin{table}
  \caption{Cosine between the true and the estimated canonical vectors for different models on blocks "Gas" and "Cross (small)", for levels of SNR -20dB, -10dB, -6dB and 0dB from top to bottom, and computation times, for n = 300 and 33 folds. Median and quantiles (2.5\% and 97.5\%) are reported.}
    \label{tab:2B_300_all}
  \centering
  \resizebox{0.6\columnwidth}{!}{%
  \begin{tabular}{llll} 
\toprule 
Model & Gas & Cross (small) & Computation time \\ 
\midrule 
2DCCA1 & 0.03 (0.00, 0.38) & 0.22 (0.13, 0.49) & 2.45 (2.01, 4.13) \\ 
TCCA1 & 0.04 (0.00, 0.80) & 0.23 (0.15, 0.71) & 9.21 (7.98, 10.50) \\ 
TGCCA1 & 0.25 (0.00, 0.85) & 0.28 (0.05, 0.81) & 1.78 (1.51, 2.65) \\ 
spTCCA1 & 0.04 (0.00, 0.75) & 0.23 (0.15, 0.65) & 8.10 (7.48, 8.65) \\ 
MGCCA & 0.59 (0.00, 0.88) & 0.44 (0.04, 0.84) & 1.69 (1.59, 2.02) \\ 
2DCCA3 & 0.02 (0.00, 0.09) & 0.04 (0.00, 0.22) & 0.80 (0.72, 1.32) \\ 
TCCA3 & 0.19 (0.00, 0.88) & 0.28 (0.15, 0.85) & 10.45 (8.27, 13.22) \\ 
TGCCA3 & 0.07 (0.00, 0.35) & 0.06 (0.00, 0.36) & 2.27 (1.95, 3.44) \\ 
spTCCA3 & 0.07 (0.00, 0.85) & 0.23 (0.16, 0.84) & 8.05 (7.33, 9.52) \\ 
spTGCCA3 & 0.06 (0.00, 0.33) & 0.05 (0.01, 0.33) & 1.79 (1.60, 2.17) \\ 
RGCCA & 0.07 (0.01, 0.17) & 0.05 (0.00, 0.12) & 1.03 (0.96, 1.11) \\ 
SVD & 0.01 (0.00, 0.04) & 0.02 (0.00, 0.05) & 0.43 (0.43, 0.49) \\ 
\bottomrule 
\end{tabular}

  }
  \resizebox{0.6\columnwidth}{!}{%
  \begin{tabular}{llll} 
\toprule 
Model & Gas & Cross (small) & Computation time \\ 
\midrule 
2DCCA1 & 0.89 (0.71, 0.90) & 0.86 (0.68, 0.87) & 2.38 (2.00, 3.50) \\ 
TCCA1 & 0.89 (0.86, 0.90) & 0.86 (0.83, 0.87) & 7.35 (7.25, 7.71) \\ 
TGCCA1 & 0.89 (0.85, 0.90) & 0.86 (0.82, 0.87) & 1.46 (1.37, 1.65) \\ 
spTCCA1 & 0.89 (0.85, 0.90) & 0.86 (0.82, 0.87) & 7.15 (7.06, 7.42) \\ 
MGCCA & 0.89 (0.86, 0.90) & 0.86 (0.83, 0.87) & 1.59 (1.47, 1.72) \\ 
2DCCA3 & 0.03 (0.00, 0.13) & 0.05 (0.00, 0.20) & 0.74 (0.71, 1.24) \\ 
TCCA3 & 0.89 (0.86, 0.90) & 0.86 (0.83, 0.87) & 7.70 (7.53, 7.88) \\ 
TGCCA3 & 0.93 (0.88, 0.95) & 0.95 (0.91, 0.97) & 2.12 (1.94, 2.25) \\ 
spTCCA3 & 0.89 (0.85, 0.90) & 0.86 (0.82, 0.87) & 7.11 (7.01, 7.20) \\ 
spTGCCA3 & 0.93 (0.88, 0.95) & 0.94 (0.91, 0.96) & 1.64 (1.52, 1.82) \\ 
RGCCA & 0.77 (0.35, 0.83) & 0.88 (0.42, 0.93) & 1.23 (1.06, 1.64) \\ 
SVD & 0.05 (0.01, 0.13) & 0.03 (0.00, 0.09) & 0.46 (0.46, 0.50) \\ 
\bottomrule 
\end{tabular}

  }
  \resizebox{0.6\columnwidth}{!}{%
  \begin{tabular}{llll} 
\toprule 
Model & Gas & Cross (small) & Computation time \\ 
\midrule 
2DCCA1 & 0.89 (0.88, 0.90) & 0.87 (0.86, 0.88) & 2.09 (1.95, 2.66) \\ 
TCCA1 & 0.90 (0.88, 0.90) & 0.87 (0.86, 0.88) & 7.32 (7.21, 7.48) \\ 
TGCCA1 & 0.90 (0.88, 0.90) & 0.87 (0.85, 0.88) & 1.47 (1.37, 1.68) \\ 
spTCCA1 & 0.90 (0.88, 0.90) & 0.87 (0.85, 0.88) & 7.15 (7.05, 7.28) \\ 
MGCCA & 0.90 (0.88, 0.90) & 0.87 (0.85, 0.87) & 1.53 (1.45, 1.78) \\ 
2DCCA3 & 0.03 (0.01, 0.11) & 0.08 (0.01, 0.21) & 0.77 (0.70, 1.27) \\ 
TCCA3 & 0.90 (0.88, 0.90) & 0.87 (0.86, 0.88) & 7.70 (7.51, 7.88) \\ 
TGCCA3 & 0.96 (0.93, 0.97) & 0.97 (0.95, 0.98) & 2.11 (1.99, 2.30) \\ 
spTCCA3 & 0.90 (0.88, 0.90) & 0.87 (0.85, 0.88) & 7.07 (6.98, 7.22) \\ 
spTGCCA3 & 0.95 (0.93, 0.96) & 0.97 (0.95, 0.98) & 1.55 (1.45, 1.78) \\ 
RGCCA & 0.83 (0.79, 0.86) & 0.94 (0.91, 0.96) & 0.99 (0.91, 1.09) \\ 
SVD & 0.08 (0.00, 0.26) & 0.11 (0.01, 0.51) & 0.43 (0.42, 0.47) \\ 
\bottomrule 
\end{tabular}

  }
  \resizebox{0.6\columnwidth}{!}{%
  \begin{tabular}{llll} 
\toprule 
Model & Gas & Cross (small) & Computation time \\ 
\midrule 
2DCCA1 & 0.90 (0.89, 0.90) & 0.87 (0.86, 0.88) & 2.76 (2.29, 3.92) \\ 
TCCA1 & 0.90 (0.89, 0.91) & 0.88 (0.86, 0.88) & 7.30 (7.17, 7.42) \\ 
TGCCA1 & 0.90 (0.89, 0.91) & 0.88 (0.86, 0.88) & 1.51 (1.35, 1.75) \\ 
spTCCA1 & 0.90 (0.89, 0.91) & 0.88 (0.86, 0.88) & 7.16 (7.07, 7.30) \\ 
MGCCA & 0.90 (0.89, 0.91) & 0.87 (0.86, 0.88) & 1.52 (1.43, 1.70) \\ 
2DCCA3 & 0.05 (0.01, 0.13) & 0.06 (0.00, 0.19) & 0.74 (0.71, 1.21) \\ 
TCCA3 & 0.90 (0.89, 0.91) & 0.88 (0.86, 0.88) & 7.69 (7.48, 7.83) \\ 
TGCCA3 & 0.97 (0.96, 0.98) & 0.99 (0.97, 0.99) & 2.03 (1.93, 2.19) \\ 
spTCCA3 & 0.90 (0.89, 0.91) & 0.88 (0.86, 0.88) & 7.07 (6.94, 7.17) \\ 
spTGCCA3 & 0.96 (0.95, 0.97) & 0.98 (0.96, 0.98) & 1.55 (1.47, 1.75) \\ 
RGCCA & 0.87 (0.84, 0.89) & 0.96 (0.94, 0.97) & 0.91 (0.86, 1.01) \\ 
SVD & 0.97 (0.89, 0.97) & 0.98 (0.87, 0.99) & 0.42 (0.42, 0.47) \\ 
\bottomrule 
\end{tabular}

  }
\end{table}
\begin{table}
  \caption{Cosine between the true and the estimated canonical vectors for different models on blocks "Gas" and "Cross (small)", for levels of SNR -20dB, -10dB, -6dB and 0dB from top to bottom, and computation times, for n = 200 and 50 folds. Median and quantiles (2.5\% and 97.5\%) are reported.}
    \label{tab:2B_200_all}
  \centering
  \resizebox{0.6\columnwidth}{!}{%
  \begin{tabular}{llll} 
\toprule 
Model & Gas & Cross (small) & Computation time \\ 
\midrule 
2DCCA1 & 0.02 (0.00, 0.28) & 0.19 (0.11, 0.41) & 1.93 (1.83, 2.99) \\ 
TCCA1 & 0.02 (0.00, 0.77) & 0.21 (0.10, 0.71) & 9.93 (8.33, 11.22) \\ 
TGCCA1 & 0.08 (0.00, 0.71) & 0.24 (0.10, 0.62) & 1.10 (0.94, 1.87) \\ 
spTCCA1 & 0.02 (0.00, 0.68) & 0.21 (0.10, 0.61) & 8.54 (7.77, 9.40) \\ 
MGCCA & 0.18 (0.00, 0.79) & 0.23 (0.04, 0.73) & 1.15 (1.06, 1.61) \\ 
2DCCA3 & 0.01 (0.00, 0.06) & 0.03 (0.00, 0.16) & 0.63 (0.60, 0.73) \\ 
TCCA3 & 0.03 (0.00, 0.82) & 0.22 (0.10, 0.76) & 10.93 (8.80, 13.37) \\ 
TGCCA3 & 0.07 (0.00, 0.30) & 0.08 (0.00, 0.26) & 1.56 (1.34, 2.27) \\ 
spTCCA3 & 0.02 (0.00, 0.67) & 0.22 (0.10, 0.59) & 8.26 (7.46, 9.85) \\ 
spTGCCA3 & 0.05 (0.00, 0.32) & 0.07 (0.01, 0.31) & 1.19 (1.12, 1.44) \\ 
RGCCA & 0.05 (0.01, 0.19) & 0.06 (0.00, 0.13) & 0.46 (0.41, 0.56) \\ 
SVD & 0.02 (0.00, 0.06) & 0.02 (0.00, 0.06) & 0.19 (0.19, 0.23) \\ 
\bottomrule 
\end{tabular}

  }
  \resizebox{0.6\columnwidth}{!}{%
  \begin{tabular}{llll} 
\toprule 
Model & Gas & Cross (small) & Computation time \\ 
\midrule 
2DCCA1 & 0.87 (0.03, 0.89) & 0.85 (0.18, 0.87) & 1.93 (1.83, 2.13) \\ 
TCCA1 & 0.88 (0.83, 0.89) & 0.85 (0.81, 0.87) & 7.59 (7.42, 8.49) \\ 
TGCCA1 & 0.88 (0.82, 0.89) & 0.85 (0.80, 0.87) & 0.99 (0.91, 1.14) \\ 
spTCCA1 & 0.88 (0.81, 0.89) & 0.85 (0.79, 0.87) & 7.70 (7.23, 8.14) \\ 
MGCCA & 0.88 (0.83, 0.89) & 0.85 (0.81, 0.86) & 1.11 (1.02, 1.23) \\ 
2DCCA3 & 0.03 (0.00, 0.08) & 0.06 (0.00, 0.23) & 0.64 (0.61, 0.73) \\ 
TCCA3 & 0.88 (0.84, 0.89) & 0.85 (0.81, 0.87) & 7.76 (7.59, 8.16) \\ 
TGCCA3 & 0.90 (0.83, 0.93) & 0.93 (0.85, 0.96) & 1.52 (1.39, 1.81) \\ 
spTCCA3 & 0.88 (0.82, 0.89) & 0.85 (0.80, 0.87) & 7.13 (7.01, 7.40) \\ 
spTGCCA3 & 0.89 (0.83, 0.92) & 0.92 (0.85, 0.94) & 1.15 (1.08, 1.24) \\ 
RGCCA & 0.59 (0.09, 0.75) & 0.76 (0.17, 0.92) & 0.66 (0.53, 1.80) \\ 
SVD & 0.05 (0.00, 0.15) & 0.05 (0.01, 0.15) & 0.20 (0.20, 0.24) \\ 
\bottomrule 
\end{tabular}

  }
  \resizebox{0.6\columnwidth}{!}{%
  \begin{tabular}{llll} 
\toprule 
Model & Gas & Cross (small) & Computation time \\ 
\midrule 
2DCCA1 & 0.88 (0.86, 0.89) & 0.86 (0.83, 0.87) & 1.93 (1.83, 2.96) \\ 
TCCA1 & 0.89 (0.87, 0.90) & 0.86 (0.83, 0.87) & 7.52 (7.30, 7.97) \\ 
TGCCA1 & 0.89 (0.87, 0.90) & 0.86 (0.83, 0.87) & 0.95 (0.86, 1.08) \\ 
spTCCA1 & 0.89 (0.85, 0.90) & 0.86 (0.82, 0.87) & 7.48 (7.18, 7.91) \\ 
MGCCA & 0.89 (0.86, 0.90) & 0.86 (0.83, 0.87) & 1.09 (1.01, 1.21) \\ 
2DCCA3 & 0.02 (0.00, 0.10) & 0.06 (0.00, 0.17) & 0.66 (0.62, 0.78) \\ 
TCCA3 & 0.89 (0.87, 0.90) & 0.86 (0.83, 0.87) & 7.74 (7.55, 8.02) \\ 
TGCCA3 & 0.94 (0.90, 0.96) & 0.96 (0.92, 0.97) & 1.47 (1.36, 1.60) \\ 
spTCCA3 & 0.89 (0.87, 0.90) & 0.86 (0.83, 0.87) & 7.09 (6.99, 7.35) \\ 
spTGCCA3 & 0.93 (0.88, 0.94) & 0.94 (0.91, 0.96) & 1.12 (1.05, 1.22) \\ 
RGCCA & 0.72 (0.62, 0.76) & 0.91 (0.85, 0.94) & 0.54 (0.45, 0.74) \\ 
SVD & 0.14 (0.01, 0.45) & 0.17 (0.01, 0.54) & 0.18 (0.18, 0.22) \\ 
\bottomrule 
\end{tabular}

  }
  \resizebox{0.6\columnwidth}{!}{%
  \begin{tabular}{llll} 
\toprule 
Model & Gas & Cross (small) & Computation time \\ 
\midrule 
2DCCA1 & 0.88 (0.86, 0.89) & 0.87 (0.85, 0.87) & 1.93 (1.85, 2.38) \\ 
TCCA1 & 0.90 (0.88, 0.90) & 0.87 (0.86, 0.87) & 7.39 (7.22, 8.03) \\ 
TGCCA1 & 0.90 (0.89, 0.91) & 0.87 (0.86, 0.88) & 0.87 (0.80, 1.01) \\ 
spTCCA1 & 0.90 (0.88, 0.91) & 0.87 (0.86, 0.88) & 7.66 (7.27, 8.03) \\ 
MGCCA & 0.90 (0.87, 0.90) & 0.87 (0.85, 0.87) & 1.07 (0.99, 1.25) \\ 
2DCCA3 & 0.03 (0.01, 0.09) & 0.06 (0.01, 0.15) & 0.66 (0.62, 0.80) \\ 
TCCA3 & 0.90 (0.88, 0.90) & 0.87 (0.86, 0.87) & 7.69 (7.54, 7.93) \\ 
TGCCA3 & 0.97 (0.94, 0.98) & 0.98 (0.96, 0.99) & 1.40 (1.35, 1.56) \\ 
spTCCA3 & 0.90 (0.89, 0.91) & 0.87 (0.86, 0.88) & 7.09 (6.97, 7.21) \\ 
spTGCCA3 & 0.94 (0.91, 0.95) & 0.96 (0.94, 0.97) & 1.08 (1.00, 1.28) \\ 
RGCCA & 0.73 (0.67, 0.79) & 0.94 (0.92, 0.96) & 0.49 (0.43, 0.56) \\ 
SVD & 0.95 (0.81, 0.96) & 0.96 (0.66, 0.98) & 0.19 (0.18, 0.22) \\ 
\bottomrule 
\end{tabular}

  }
\end{table}
\begin{table}
  \caption{Cosine between the true and the estimated canonical vectors for different models on blocks "Gas" and "Cross (small)", for levels of SNR -20dB, -10dB, -6dB and 0dB from top to bottom, and computation times, for n = 100 and 100 folds. Median and quantiles (2.5\% and 97.5\%) are reported.}
    \label{tab:2B_100_all}
  \centering
  \resizebox{0.6\columnwidth}{!}{%
  \begin{tabular}{llll} 
\toprule 
Model & Gas & Cross (small) & Computation time \\ 
\midrule 
2DCCA1 & 0.02 (0.00, 0.14) & 0.17 (0.04, 0.35) & 2.28 (1.93, 3.22) \\ 
TCCA1 & 0.02 (0.00, 0.21) & 0.19 (0.09, 0.39) & 10.69 (9.35, 11.81) \\ 
TGCCA1 & 0.04 (0.00, 0.52) & 0.20 (0.01, 0.45) & 0.62 (0.51, 0.91) \\ 
spTCCA1 & 0.02 (0.00, 0.40) & 0.20 (0.05, 0.38) & 9.32 (8.48, 10.05) \\ 
MGCCA & 0.07 (0.00, 0.56) & 0.18 (0.01, 0.50) & 0.72 (0.64, 0.87) \\ 
2DCCA3 & 0.01 (0.00, 0.05) & 0.03 (0.00, 0.13) & 0.53 (0.51, 0.61) \\ 
TCCA3 & 0.02 (0.00, 0.57) & 0.19 (0.09, 0.48) & 13.11 (10.14, 13.55) \\ 
TGCCA3 & 0.06 (0.00, 0.35) & 0.07 (0.01, 0.24) & 1.01 (0.90, 1.72) \\ 
spTCCA3 & 0.02 (0.00, 0.52) & 0.20 (0.05, 0.42) & 9.63 (7.77, 11.28) \\ 
spTGCCA3 & 0.05 (0.00, 0.30) & 0.06 (0.00, 0.27) & 0.73 (0.63, 0.90) \\ 
RGCCA & 0.05 (0.00, 0.17) & 0.04 (0.00, 0.14) & 0.22 (0.16, 0.29) \\ 
SVD & 0.03 (0.00, 0.08) & 0.03 (0.00, 0.08) & 0.04 (0.04, 0.08) \\ 
\bottomrule 
\end{tabular}

  }
  \resizebox{0.6\columnwidth}{!}{%
  \begin{tabular}{llll} 
\toprule 
Model & Gas & Cross (small) & Computation time \\ 
\midrule 
2DCCA1 & 0.27 (0.01, 0.80) & 0.56 (0.03, 0.84) & 1.90 (1.81, 2.65) \\ 
TCCA1 & 0.80 (0.00, 0.86) & 0.81 (0.10, 0.85) & 8.74 (8.08, 11.07) \\ 
TGCCA1 & 0.84 (0.62, 0.88) & 0.81 (0.36, 0.86) & 0.58 (0.49, 0.86) \\ 
spTCCA1 & 0.83 (0.00, 0.88) & 0.79 (0.15, 0.86) & 8.27 (7.84, 9.34) \\ 
MGCCA & 0.83 (0.66, 0.87) & 0.80 (0.64, 0.85) & 0.68 (0.61, 0.82) \\ 
2DCCA3 & 0.01 (0.00, 0.05) & 0.04 (0.00, 0.12) & 0.53 (0.51, 0.68) \\ 
TCCA3 & 0.80 (0.01, 0.86) & 0.81 (0.12, 0.85) & 8.22 (7.81, 11.55) \\ 
TGCCA3 & 0.78 (0.19, 0.88) & 0.81 (0.07, 0.93) & 1.04 (0.89, 2.11) \\ 
spTCCA3 & 0.83 (0.01, 0.88) & 0.80 (0.16, 0.86) & 7.42 (7.10, 9.28) \\ 
spTGCCA3 & 0.76 (0.12, 0.83) & 0.80 (0.15, 0.89) & 0.76 (0.65, 1.01) \\ 
RGCCA & 0.17 (0.01, 0.53) & 0.31 (0.04, 0.79) & 0.28 (0.19, 0.75) \\ 
SVD & 0.06 (0.00, 0.19) & 0.07 (0.01, 0.24) & 0.05 (0.04, 0.07) \\ 
\bottomrule 
\end{tabular}

  }
  \resizebox{0.6\columnwidth}{!}{%
  \begin{tabular}{llll} 
\toprule 
Model & Gas & Cross (small) & Computation time \\ 
\midrule 
2DCCA1 & 0.70 (0.10, 0.78) & 0.82 (0.31, 0.84) & 1.91 (1.82, 2.12) \\ 
TCCA1 & 0.82 (0.02, 0.85) & 0.83 (0.35, 0.85) & 8.32 (7.94, 9.10) \\ 
TGCCA1 & 0.88 (0.81, 0.89) & 0.85 (0.75, 0.87) & 0.53 (0.45, 0.66) \\ 
spTCCA1 & 0.87 (0.01, 0.89) & 0.84 (0.18, 0.86) & 7.27 (7.11, 8.26) \\ 
MGCCA & 0.86 (0.81, 0.88) & 0.83 (0.77, 0.86) & 0.67 (0.60, 0.75) \\ 
2DCCA3 & 0.02 (0.00, 0.05) & 0.04 (0.00, 0.11) & 0.54 (0.51, 0.62) \\ 
TCCA3 & 0.82 (0.33, 0.85) & 0.83 (0.48, 0.85) & 8.02 (7.77, 8.79) \\ 
TGCCA3 & 0.89 (0.78, 0.93) & 0.92 (0.82, 0.95) & 0.93 (0.85, 1.06) \\ 
spTCCA3 & 0.88 (0.39, 0.89) & 0.85 (0.47, 0.87) & 7.81 (7.66, 8.12) \\ 
spTGCCA3 & 0.84 (0.72, 0.88) & 0.87 (0.80, 0.91) & 0.69 (0.62, 0.80) \\ 
RGCCA & 0.47 (0.14, 0.57) & 0.84 (0.34, 0.90) & 0.34 (0.22, 0.93) \\ 
SVD & 0.17 (0.01, 0.67) & 0.19 (0.01, 0.87) & 0.05 (0.04, 0.08) \\ 
\bottomrule 
\end{tabular}

  }
  \resizebox{0.6\columnwidth}{!}{%
  \begin{tabular}{llll} 
\toprule 
Model & Gas & Cross (small) & Computation time \\ 
\midrule 
2DCCA1 & 0.63 (0.50, 0.75) & 0.81 (0.77, 0.85) & 1.92 (1.83, 2.14) \\ 
TCCA1 & 0.81 (0.73, 0.86) & 0.84 (0.80, 0.86) & 8.38 (8.03, 8.88) \\ 
TGCCA1 & 0.89 (0.86, 0.90) & 0.86 (0.84, 0.88) & 0.48 (0.42, 0.55) \\ 
spTCCA1 & 0.89 (0.86, 0.90) & 0.86 (0.84, 0.87) & 7.25 (7.13, 7.55) \\ 
MGCCA & 0.88 (0.84, 0.89) & 0.85 (0.82, 0.87) & 0.64 (0.59, 0.72) \\ 
2DCCA3 & 0.02 (0.00, 0.05) & 0.04 (0.00, 0.14) & 0.53 (0.51, 0.60) \\ 
TCCA3 & 0.81 (0.73, 0.86) & 0.84 (0.80, 0.86) & 8.13 (7.91, 9.43) \\ 
TGCCA3 & 0.94 (0.88, 0.97) & 0.96 (0.93, 0.98) & 0.92 (0.84, 1.02) \\ 
spTCCA3 & 0.89 (0.86, 0.90) & 0.86 (0.84, 0.88) & 7.13 (7.00, 7.33) \\ 
spTGCCA3 & 0.86 (0.79, 0.90) & 0.90 (0.87, 0.92) & 0.67 (0.60, 0.76) \\ 
RGCCA & 0.46 (0.38, 0.57) & 0.90 (0.87, 0.93) & 0.22 (0.15, 0.33) \\ 
SVD & 0.90 (0.39, 0.93) & 0.94 (0.53, 0.97) & 0.05 (0.04, 0.08) \\ 
\bottomrule 
\end{tabular}

  }
\end{table}
\begin{table}
  \caption{Cosine between the true and the estimated canonical vectors for different models on blocks "Square", "Gas", "Cross", "Cross (small)" and "Vector", for levels of SNR -20dB, -10dB, -6dB, and 0dB from top to bottom, and computation times, for n = 1000 and 10 folds. Median and quantiles (2.5\% and 97.5\%) are reported.}
    \label{tab:5B_1000_all}
  \centering
  \resizebox{\columnwidth}{!}{%
  \begin{tabular}{lllllll} 
\toprule 
Model & Square & Gas & Cross & Cross (small) & Vector & Computation time \\ 
\midrule 
TGCCA1 & 0.99 (0.97, 0.99) & 0.90 (0.89, 0.90) & 0.88 (0.87, 0.88) & 0.86 (0.85, 0.87) & 0.94 (0.92, 0.96) & 20.35 (19.73, 21.34) \\ 
MGCCA & 0.99 (0.97, 0.99) & 0.90 (0.89, 0.90) & 0.88 (0.87, 0.88) & 0.87 (0.85, 0.87) & 0.94 (0.92, 0.96) & 9.88 (9.61, 10.20) \\ 
TGCCA3 & 0.96 (0.94, 0.97) & 0.96 (0.94, 0.97) & 0.97 (0.94, 0.97) & 0.96 (0.95, 0.97) & 0.94 (0.93, 0.96) & 23.55 (22.51, 27.70) \\ 
spTGCCA3 & 0.96 (0.94, 0.97) & 0.96 (0.94, 0.97) & 0.97 (0.94, 0.97) & 0.96 (0.95, 0.97) & 0.94 (0.93, 0.96) & 10.15 (9.91, 10.61) \\ 
RGCCA & 0.88 (0.84, 0.89) & 0.86 (0.79, 0.88) & 0.87 (0.82, 0.88) & 0.91 (0.87, 0.93) & 0.93 (0.92, 0.96) & 32.44 (28.48, 36.78) \\ 
SVD & 0.01 (0.00, 0.02) & 0.00 (0.00, 0.01) & 0.01 (0.00, 0.02) & 0.01 (0.00, 0.03) & 0.07 (0.01, 0.13) & 12.85 (12.65, 13.01) \\ 
\bottomrule 
\end{tabular}

  }
  \resizebox{\columnwidth}{!}{%
  \begin{tabular}{lllllll} 
\toprule 
Model & Square & Gas & Cross & Cross (small) & Vector & Computation time \\ 
\midrule 
TGCCA1 & 1.00 (0.99, 1.00) & 0.91 (0.90, 0.91) & 0.89 (0.89, 0.89) & 0.88 (0.87, 0.88) & 0.98 (0.98, 0.99) & 19.02 (18.88, 19.45) \\ 
MGCCA & 1.00 (0.99, 1.00) & 0.91 (0.90, 0.91) & 0.89 (0.89, 0.89) & 0.88 (0.87, 0.88) & 0.98 (0.98, 0.99) & 9.63 (9.46, 10.26) \\ 
TGCCA3 & 0.99 (0.98, 0.99) & 0.98 (0.97, 0.98) & 0.99 (0.99, 0.99) & 0.99 (0.99, 0.99) & 0.98 (0.98, 0.99) & 24.19 (22.21, 25.08) \\ 
spTGCCA3 & 0.99 (0.98, 0.99) & 0.98 (0.97, 0.98) & 0.99 (0.99, 0.99) & 0.99 (0.99, 0.99) & 0.98 (0.98, 0.99) & 9.77 (9.62, 10.19) \\ 
RGCCA & 0.96 (0.95, 0.96) & 0.95 (0.94, 0.95) & 0.95 (0.95, 0.96) & 0.98 (0.97, 0.98) & 0.98 (0.98, 0.99) & 29.80 (28.21, 31.24) \\ 
SVD & 0.02 (0.00, 0.11) & 0.02 (0.01, 0.06) & 0.01 (0.00, 0.06) & 0.01 (0.00, 0.04) & 0.98 (0.97, 0.99) & 12.93 (12.86, 13.74) \\ 
\bottomrule 
\end{tabular}

  }
  \resizebox{\columnwidth}{!}{%
  \begin{tabular}{lllllll} 
\toprule 
Model & Square & Gas & Cross & Cross (small) & Vector & Computation time \\ 
\midrule 
TGCCA1 & 1.00 (1.00, 1.00) & 0.91 (0.91, 0.91) & 0.89 (0.89, 0.89) & 0.88 (0.88, 0.88) & 0.99 (0.98, 0.99) & 18.97 (18.66, 19.40) \\ 
MGCCA & 1.00 (1.00, 1.00) & 0.91 (0.91, 0.91) & 0.89 (0.89, 0.89) & 0.88 (0.88, 0.88) & 0.99 (0.98, 0.99) & 9.83 (9.61, 10.37) \\ 
TGCCA3 & 0.99 (0.99, 0.99) & 0.98 (0.98, 0.98) & 0.99 (0.99, 0.99) & 0.99 (0.99, 1.00) & 0.99 (0.98, 0.99) & 22.76 (22.00, 24.48) \\ 
spTGCCA3 & 0.99 (0.99, 0.99) & 0.98 (0.98, 0.98) & 0.99 (0.99, 0.99) & 0.99 (0.99, 0.99) & 0.99 (0.98, 0.99) & 9.91 (9.72, 10.28) \\ 
RGCCA & 0.97 (0.97, 0.98) & 0.97 (0.96, 0.97) & 0.97 (0.96, 0.97) & 0.98 (0.98, 0.99) & 0.99 (0.98, 0.99) & 28.99 (27.72, 30.75) \\ 
SVD & 0.06 (0.02, 0.19) & 0.07 (0.01, 0.13) & 0.03 (0.00, 0.15) & 0.05 (0.01, 0.21) & 0.99 (0.99, 0.99) & 11.88 (11.75, 11.92) \\ 
\bottomrule 
\end{tabular}

  }
  \resizebox{\columnwidth}{!}{%
  \begin{tabular}{lllllll} 
\toprule 
Model & Square & Gas & Cross & Cross (small) & Vector & Computation time \\ 
\midrule 
TGCCA1 & 1.00 (1.00, 1.00) & 0.91 (0.91, 0.91) & 0.89 (0.89, 0.89) & 0.88 (0.88, 0.88) & 0.99 (0.99, 1.00) & 18.24 (18.08, 18.48) \\ 
MGCCA & 1.00 (1.00, 1.00) & 0.91 (0.91, 0.91) & 0.89 (0.89, 0.89) & 0.88 (0.88, 0.88) & 0.99 (0.99, 1.00) & 9.66 (9.47, 10.26) \\ 
TGCCA3 & 1.00 (1.00, 1.00) & 0.98 (0.98, 0.98) & 1.00 (0.99, 1.00) & 1.00 (0.99, 1.00) & 0.99 (0.99, 1.00) & 22.02 (21.74, 22.83) \\ 
spTGCCA3 & 1.00 (0.99, 1.00) & 0.98 (0.98, 0.98) & 1.00 (0.99, 1.00) & 1.00 (0.99, 1.00) & 0.99 (0.99, 1.00) & 9.72 (9.48, 10.25) \\ 
RGCCA & 0.99 (0.98, 0.99) & 0.98 (0.98, 0.99) & 0.98 (0.98, 0.99) & 0.99 (0.99, 0.99) & 0.99 (0.99, 1.00) & 30.16 (29.33, 31.02) \\ 
SVD & 0.99 (0.99, 0.99) & 0.99 (0.97, 0.99) & 0.99 (0.98, 0.99) & 0.99 (0.96, 1.00) & 1.00 (1.00, 1.00) & 12.80 (12.68, 13.13) \\ 
\bottomrule 
\end{tabular}

  }

\end{table}
\begin{table}
  \caption{Cosine between the true and the estimated canonical vectors for different models on blocks "Square", "Gas", "Cross", "Cross (small)" and "Vector", for levels of SNR -20dB, -10dB, -6dB, and 0dB from top to bottom, and computation times, for n = 500 and 20 folds. Median and quantiles (2.5\% and 97.5\%) are reported.}
    \label{tab:5B_500_all}
  \centering
  \resizebox{\columnwidth}{!}{%
  \begin{tabular}{lllllll} 
\toprule 
Model & Square & Gas & Cross & Cross (small) & Vector & Computation time \\ 
\midrule 
TGCCA1 & 0.98 (0.91, 0.99) & 0.89 (0.82, 0.89) & 0.87 (0.73, 0.88) & 0.86 (0.83, 0.87) & 0.90 (0.85, 0.93) & 6.91 (6.33, 8.42) \\ 
MGCCA & 0.98 (0.94, 0.99) & 0.89 (0.84, 0.89) & 0.87 (0.78, 0.88) & 0.86 (0.82, 0.87) & 0.90 (0.86, 0.93) & 5.34 (4.95, 5.77) \\ 
TGCCA3 & 0.92 (0.78, 0.94) & 0.91 (0.68, 0.94) & 0.93 (0.76, 0.95) & 0.93 (0.83, 0.95) & 0.89 (0.81, 0.92) & 9.56 (8.90, 10.66) \\ 
spTGCCA3 & 0.92 (0.79, 0.93) & 0.91 (0.69, 0.94) & 0.93 (0.79, 0.95) & 0.93 (0.84, 0.94) & 0.89 (0.81, 0.93) & 5.60 (5.38, 6.32) \\ 
RGCCA & 0.78 (0.41, 0.82) & 0.74 (0.41, 0.80) & 0.76 (0.47, 0.80) & 0.81 (0.50, 0.87) & 0.87 (0.66, 0.92) & 9.16 (7.81, 11.87) \\ 
SVD & 0.01 (0.00, 0.04) & 0.01 (0.00, 0.03) & 0.02 (0.00, 0.03) & 0.01 (0.00, 0.04) & 0.03 (0.00, 0.17) & 3.31 (2.90, 3.53) \\ 
\bottomrule 
\end{tabular}

  }
  \resizebox{\columnwidth}{!}{%
  \begin{tabular}{lllllll} 
\toprule 
Model & Square & Gas & Cross & Cross (small) & Vector & Computation time \\ 
\midrule 
TGCCA1 & 0.99 (0.98, 1.00) & 0.90 (0.89, 0.90) & 0.88 (0.87, 0.89) & 0.87 (0.86, 0.88) & 0.97 (0.95, 0.97) & 6.01 (5.77, 6.60) \\ 
MGCCA & 0.99 (0.98, 1.00) & 0.90 (0.88, 0.90) & 0.88 (0.87, 0.89) & 0.87 (0.86, 0.88) & 0.97 (0.95, 0.97) & 5.10 (4.96, 5.52) \\ 
TGCCA3 & 0.97 (0.96, 0.98) & 0.97 (0.95, 0.97) & 0.98 (0.96, 0.98) & 0.98 (0.96, 0.98) & 0.97 (0.95, 0.98) & 8.79 (8.46, 9.60) \\ 
spTGCCA3 & 0.97 (0.96, 0.97) & 0.96 (0.95, 0.97) & 0.98 (0.96, 0.98) & 0.98 (0.96, 0.98) & 0.97 (0.95, 0.98) & 5.30 (5.10, 5.74) \\ 
RGCCA & 0.92 (0.90, 0.93) & 0.90 (0.88, 0.91) & 0.91 (0.89, 0.92) & 0.95 (0.93, 0.96) & 0.97 (0.95, 0.98) & 6.43 (6.22, 7.21) \\ 
SVD & 0.03 (0.00, 0.13) & 0.04 (0.00, 0.09) & 0.03 (0.01, 0.07) & 0.02 (0.00, 0.08) & 0.95 (0.92, 0.97) & 3.50 (3.29, 3.57) \\ 
\bottomrule 
\end{tabular}

  }
  \resizebox{\columnwidth}{!}{%
  \begin{tabular}{lllllll} 
\toprule 
Model & Square & Gas & Cross & Cross (small) & Vector & Computation time \\ 
\midrule 
TGCCA1 & 1.00 (0.99, 1.00) & 0.91 (0.90, 0.91) & 0.89 (0.88, 0.89) & 0.88 (0.87, 0.88) & 0.98 (0.97, 0.98) & 6.00 (5.80, 6.24) \\ 
MGCCA & 1.00 (0.99, 1.00) & 0.90 (0.90, 0.91) & 0.89 (0.88, 0.89) & 0.88 (0.86, 0.88) & 0.98 (0.97, 0.98) & 5.21 (5.05, 5.59) \\ 
TGCCA3 & 0.98 (0.98, 0.99) & 0.98 (0.97, 0.98) & 0.99 (0.98, 0.99) & 0.99 (0.98, 0.99) & 0.98 (0.97, 0.98) & 8.87 (8.60, 9.45) \\ 
spTGCCA3 & 0.98 (0.97, 0.98) & 0.97 (0.97, 0.98) & 0.98 (0.98, 0.99) & 0.99 (0.97, 0.99) & 0.98 (0.97, 0.98) & 5.36 (5.09, 5.73) \\ 
RGCCA & 0.95 (0.94, 0.96) & 0.93 (0.92, 0.94) & 0.94 (0.93, 0.95) & 0.97 (0.96, 0.97) & 0.98 (0.97, 0.98) & 6.79 (6.60, 7.09) \\ 
SVD & 0.08 (0.01, 0.25) & 0.09 (0.02, 0.18) & 0.06 (0.01, 0.24) & 0.10 (0.01, 0.24) & 0.98 (0.97, 0.99) & 3.50 (2.87, 3.56) \\ 
\bottomrule 
\end{tabular}

  }
  \resizebox{\columnwidth}{!}{%
  \begin{tabular}{lllllll} 
\toprule 
Model & Square & Gas & Cross & Cross (small) & Vector & Computation time \\ 
\midrule 
TGCCA1 & 1.00 (1.00, 1.00) & 0.91 (0.90, 0.91) & 0.89 (0.88, 0.89) & 0.88 (0.87, 0.88) & 0.99 (0.99, 0.99) & 5.87 (5.65, 6.12) \\ 
MGCCA & 1.00 (0.99, 1.00) & 0.91 (0.90, 0.91) & 0.89 (0.88, 0.89) & 0.88 (0.87, 0.88) & 0.99 (0.99, 0.99) & 5.10 (4.91, 5.49) \\ 
TGCCA3 & 0.99 (0.99, 0.99) & 0.98 (0.98, 0.98) & 0.99 (0.99, 0.99) & 0.99 (0.99, 1.00) & 0.99 (0.99, 0.99) & 8.90 (8.34, 9.83) \\ 
spTGCCA3 & 0.99 (0.99, 0.99) & 0.98 (0.97, 0.98) & 0.99 (0.98, 0.99) & 0.99 (0.99, 0.99) & 0.99 (0.99, 0.99) & 5.34 (5.00, 5.77) \\ 
RGCCA & 0.97 (0.96, 0.98) & 0.96 (0.95, 0.97) & 0.96 (0.95, 0.97) & 0.98 (0.98, 0.99) & 0.99 (0.99, 0.99) & 6.46 (6.33, 6.91) \\ 
SVD & 0.98 (0.96, 0.99) & 0.98 (0.95, 0.98) & 0.98 (0.94, 0.98) & 0.99 (0.91, 0.99) & 0.99 (0.99, 0.99) & 3.16 (2.92, 3.33) \\ 
\bottomrule 
\end{tabular}

  }

\end{table}
\begin{table}
  \caption{Cosine between the true and the estimated canonical vectors for different models on blocks "Square", "Gas", "Cross", "Cross (small)" and "Vector", for levels of SNR -20dB, -10dB, -6dB, and 0dB from top to bottom, and computation times, for n = 300 and 33 folds. Median and quantiles (2.5\% and 97.5\%) are reported.}
    \label{tab:5B_300_all}
  \centering
  \resizebox{\columnwidth}{!}{%
  \begin{tabular}{lllllll} 
\toprule 
Model & Square & Gas & Cross & Cross (small) & Vector & Computation time \\ 
\midrule 
TGCCA1 & 0.95 (0.44, 0.98) & 0.87 (0.50, 0.89) & 0.85 (0.33, 0.87) & 0.82 (0.52, 0.86) & 0.83 (0.51, 0.89) & 4.09 (3.55, 5.89) \\ 
MGCCA & 0.95 (0.83, 0.97) & 0.87 (0.75, 0.89) & 0.85 (0.69, 0.87) & 0.83 (0.76, 0.86) & 0.84 (0.73, 0.89) & 3.55 (3.27, 4.01) \\ 
TGCCA3 & 0.86 (0.31, 0.90) & 0.82 (0.39, 0.89) & 0.87 (0.39, 0.91) & 0.86 (0.47, 0.91) & 0.82 (0.51, 0.89) & 6.44 (5.64, 10.08) \\ 
spTGCCA3 & 0.85 (0.35, 0.89) & 0.82 (0.41, 0.88) & 0.87 (0.46, 0.91) & 0.87 (0.53, 0.91) & 0.82 (0.56, 0.89) & 3.77 (3.37, 4.40) \\ 
RGCCA & 0.67 (0.21, 0.72) & 0.62 (0.27, 0.69) & 0.64 (0.31, 0.70) & 0.69 (0.33, 0.78) & 0.79 (0.42, 0.84) & 3.10 (2.64, 4.01) \\ 
SVD & 0.02 (0.00, 0.06) & 0.01 (0.00, 0.04) & 0.02 (0.00, 0.06) & 0.02 (0.00, 0.05) & 0.04 (0.01, 0.15) & 1.12 (1.08, 1.16) \\ 
\bottomrule 
\end{tabular}

  }
  \resizebox{\columnwidth}{!}{%
  \begin{tabular}{lllllll} 
\toprule 
Model & Square & Gas & Cross & Cross (small) & Vector & Computation time \\ 
\midrule 
TGCCA1 & 0.99 (0.97, 0.99) & 0.90 (0.88, 0.90) & 0.88 (0.85, 0.88) & 0.87 (0.85, 0.87) & 0.94 (0.92, 0.96) & 3.18 (3.03, 3.38) \\ 
MGCCA & 0.99 (0.98, 0.99) & 0.90 (0.87, 0.90) & 0.88 (0.85, 0.89) & 0.86 (0.84, 0.87) & 0.94 (0.92, 0.96) & 3.27 (3.04, 3.49) \\ 
TGCCA3 & 0.96 (0.94, 0.97) & 0.95 (0.92, 0.96) & 0.96 (0.93, 0.97) & 0.96 (0.94, 0.98) & 0.94 (0.92, 0.96) & 5.21 (5.00, 5.43) \\ 
spTGCCA3 & 0.95 (0.93, 0.96) & 0.95 (0.92, 0.96) & 0.96 (0.93, 0.97) & 0.96 (0.94, 0.97) & 0.95 (0.92, 0.96) & 3.34 (3.07, 3.57) \\ 
RGCCA & 0.87 (0.85, 0.89) & 0.83 (0.80, 0.87) & 0.85 (0.82, 0.88) & 0.91 (0.89, 0.94) & 0.94 (0.92, 0.96) & 2.22 (2.05, 2.43) \\ 
SVD & 0.04 (0.00, 0.13) & 0.05 (0.01, 0.13) & 0.03 (0.00, 0.12) & 0.03 (0.00, 0.09) & 0.90 (0.69, 0.95) & 0.96 (0.93, 1.00) \\ 
\bottomrule 
\end{tabular}

  }
  \resizebox{\columnwidth}{!}{%
  \begin{tabular}{lllllll} 
\toprule 
Model & Square & Gas & Cross & Cross (small) & Vector & Computation time \\ 
\midrule 
TGCCA1 & 0.99 (0.97, 1.00) & 0.90 (0.89, 0.91) & 0.88 (0.88, 0.89) & 0.87 (0.86, 0.88) & 0.96 (0.95, 0.98) & 3.14 (2.94, 3.33) \\ 
MGCCA & 0.99 (0.98, 1.00) & 0.90 (0.89, 0.90) & 0.88 (0.87, 0.89) & 0.87 (0.86, 0.88) & 0.96 (0.95, 0.98) & 3.26 (3.06, 3.42) \\ 
TGCCA3 & 0.97 (0.95, 0.98) & 0.97 (0.96, 0.97) & 0.98 (0.97, 0.98) & 0.98 (0.97, 0.98) & 0.96 (0.95, 0.98) & 5.04 (4.86, 5.34) \\ 
spTGCCA3 & 0.96 (0.95, 0.97) & 0.96 (0.95, 0.97) & 0.97 (0.96, 0.97) & 0.97 (0.96, 0.98) & 0.96 (0.95, 0.98) & 3.39 (3.09, 3.66) \\ 
RGCCA & 0.90 (0.88, 0.93) & 0.88 (0.85, 0.89) & 0.89 (0.86, 0.90) & 0.95 (0.93, 0.96) & 0.97 (0.95, 0.98) & 2.18 (2.09, 2.36) \\ 
SVD & 0.11 (0.00, 0.29) & 0.08 (0.00, 0.26) & 0.05 (0.01, 0.21) & 0.11 (0.01, 0.51) & 0.97 (0.95, 0.98) & 1.02 (0.93, 1.09) \\ 
\bottomrule 
\end{tabular}

  }
  \resizebox{\columnwidth}{!}{%
  \begin{tabular}{lllllll} 
\toprule 
Model & Square & Gas & Cross & Cross (small) & Vector & Computation time \\ 
\midrule 
TGCCA1 & 1.00 (0.99, 1.00) & 0.91 (0.90, 0.91) & 0.89 (0.88, 0.89) & 0.88 (0.87, 0.88) & 0.98 (0.97, 0.99) & 3.08 (2.95, 3.33) \\ 
MGCCA & 1.00 (0.99, 1.00) & 0.90 (0.90, 0.91) & 0.89 (0.88, 0.89) & 0.88 (0.87, 0.88) & 0.98 (0.97, 0.99) & 3.34 (3.03, 3.49) \\ 
TGCCA3 & 0.99 (0.98, 0.99) & 0.98 (0.97, 0.98) & 0.99 (0.98, 0.99) & 0.99 (0.98, 0.99) & 0.98 (0.97, 0.99) & 5.03 (4.86, 5.31) \\ 
spTGCCA3 & 0.98 (0.97, 0.98) & 0.97 (0.96, 0.97) & 0.98 (0.97, 0.98) & 0.98 (0.97, 0.99) & 0.98 (0.97, 0.99) & 3.38 (3.15, 3.81) \\ 
RGCCA & 0.94 (0.91, 0.95) & 0.91 (0.89, 0.92) & 0.92 (0.89, 0.93) & 0.97 (0.96, 0.98) & 0.98 (0.97, 0.99) & 2.03 (1.96, 2.18) \\ 
SVD & 0.96 (0.88, 0.98) & 0.97 (0.89, 0.97) & 0.97 (0.92, 0.98) & 0.98 (0.87, 0.99) & 0.99 (0.98, 0.99) & 1.07 (1.06, 1.20) \\ 
\bottomrule 
\end{tabular}

  }

\end{table}
\begin{table}
  \caption{Cosine between the true and the estimated canonical vectors for different models on blocks "Square", "Gas", "Cross", "Cross (small)" and "Vector", for levels of SNR -20dB, -10dB, -6dB, and 0dB from top to bottom, and computation times, for n = 200 and 50 folds. Median and quantiles (2.5\% and 97.5\%) are reported.}
    \label{tab:5B_200_all}
  \centering
  \resizebox{\columnwidth}{!}{%
  \begin{tabular}{lllllll} 
\toprule 
Model & Square & Gas & Cross & Cross (small) & Vector & Computation time \\ 
\midrule 
TGCCA1 & 0.91 (0.57, 0.96) & 0.84 (0.49, 0.87) & 0.83 (0.23, 0.86) & 0.79 (0.61, 0.85) & 0.77 (0.50, 0.85) & 3.23 (2.59, 5.80) \\ 
MGCCA & 0.92 (0.62, 0.96) & 0.85 (0.50, 0.87) & 0.83 (0.32, 0.86) & 0.80 (0.70, 0.84) & 0.79 (0.52, 0.85) & 2.59 (2.39, 3.13) \\ 
TGCCA3 & 0.70 (0.18, 0.85) & 0.68 (0.07, 0.83) & 0.68 (0.09, 0.88) & 0.74 (0.10, 0.88) & 0.69 (0.14, 0.84) & 5.38 (4.37, 9.74) \\ 
spTGCCA3 & 0.72 (0.06, 0.82) & 0.70 (0.05, 0.82) & 0.75 (0.05, 0.87) & 0.76 (0.07, 0.88) & 0.72 (0.07, 0.84) & 2.97 (2.62, 3.96) \\ 
RGCCA & 0.57 (0.24, 0.67) & 0.52 (0.20, 0.63) & 0.55 (0.22, 0.67) & 0.63 (0.29, 0.78) & 0.67 (0.36, 0.81) & 2.07 (1.52, 3.81) \\ 
SVD & 0.02 (0.00, 0.06) & 0.02 (0.00, 0.06) & 0.02 (0.00, 0.06) & 0.02 (0.00, 0.06) & 0.05 (0.01, 0.20) & 0.49 (0.42, 0.53) \\ 
\bottomrule 
\end{tabular}

  }
  \resizebox{\columnwidth}{!}{%
  \begin{tabular}{lllllll} 
\toprule 
Model & Square & Gas & Cross & Cross (small) & Vector & Computation time \\ 
\midrule 
TGCCA1 & 0.98 (0.96, 0.99) & 0.89 (0.86, 0.90) & 0.87 (0.84, 0.88) & 0.86 (0.84, 0.87) & 0.92 (0.87, 0.94) & 2.20 (2.10, 2.39) \\ 
MGCCA & 0.98 (0.96, 0.99) & 0.89 (0.86, 0.90) & 0.87 (0.85, 0.88) & 0.86 (0.83, 0.87) & 0.92 (0.87, 0.94) & 2.36 (2.22, 2.60) \\ 
TGCCA3 & 0.94 (0.90, 0.96) & 0.93 (0.89, 0.95) & 0.94 (0.91, 0.96) & 0.95 (0.92, 0.96) & 0.92 (0.87, 0.94) & 3.87 (3.65, 4.21) \\ 
spTGCCA3 & 0.91 (0.88, 0.94) & 0.92 (0.88, 0.94) & 0.93 (0.90, 0.95) & 0.94 (0.90, 0.95) & 0.92 (0.87, 0.95) & 2.42 (2.29, 2.67) \\ 
RGCCA & 0.80 (0.74, 0.85) & 0.76 (0.70, 0.80) & 0.77 (0.72, 0.81) & 0.88 (0.85, 0.91) & 0.92 (0.87, 0.95) & 1.18 (1.05, 1.37) \\ 
SVD & 0.05 (0.00, 0.20) & 0.05 (0.00, 0.15) & 0.07 (0.02, 0.13) & 0.05 (0.01, 0.15) & 0.78 (0.10, 0.93) & 0.46 (0.44, 0.48) \\ 
\bottomrule 
\end{tabular}

  }
  \resizebox{\columnwidth}{!}{%
  \begin{tabular}{lllllll} 
\toprule 
Model & Square & Gas & Cross & Cross (small) & Vector & Computation time \\ 
\midrule 
TGCCA1 & 0.99 (0.97, 1.00) & 0.90 (0.88, 0.90) & 0.88 (0.87, 0.88) & 0.87 (0.86, 0.87) & 0.95 (0.89, 0.96) & 2.12 (2.00, 2.33) \\ 
MGCCA & 0.99 (0.97, 0.99) & 0.89 (0.89, 0.90) & 0.88 (0.87, 0.88) & 0.86 (0.85, 0.87) & 0.95 (0.90, 0.96) & 2.39 (2.22, 2.67) \\ 
TGCCA3 & 0.96 (0.94, 0.97) & 0.95 (0.93, 0.96) & 0.97 (0.95, 0.97) & 0.97 (0.95, 0.98) & 0.95 (0.90, 0.96) & 3.99 (3.81, 4.25) \\ 
spTGCCA3 & 0.94 (0.91, 0.95) & 0.94 (0.92, 0.95) & 0.95 (0.94, 0.96) & 0.95 (0.93, 0.97) & 0.95 (0.90, 0.96) & 2.40 (2.26, 2.56) \\ 
RGCCA & 0.84 (0.80, 0.88) & 0.79 (0.74, 0.81) & 0.80 (0.76, 0.84) & 0.92 (0.88, 0.93) & 0.95 (0.90, 0.97) & 1.10 (1.04, 1.32) \\ 
SVD & 0.10 (0.01, 0.50) & 0.14 (0.01, 0.45) & 0.08 (0.01, 0.38) & 0.17 (0.01, 0.54) & 0.95 (0.89, 0.97) & 0.45 (0.43, 0.49) \\ 
\bottomrule 
\end{tabular}

  }
  \resizebox{\columnwidth}{!}{%
  \begin{tabular}{lllllll} 
\toprule 
Model & Square & Gas & Cross & Cross (small) & Vector & Computation time \\ 
\midrule 
TGCCA1 & 1.00 (0.99, 1.00) & 0.90 (0.89, 0.91) & 0.89 (0.87, 0.89) & 0.88 (0.86, 0.88) & 0.97 (0.95, 0.98) & 1.99 (1.89, 2.23) \\ 
MGCCA & 0.99 (0.99, 1.00) & 0.90 (0.89, 0.90) & 0.88 (0.87, 0.89) & 0.87 (0.86, 0.88) & 0.97 (0.95, 0.98) & 2.30 (2.17, 2.46) \\ 
TGCCA3 & 0.98 (0.97, 0.99) & 0.97 (0.96, 0.98) & 0.98 (0.97, 0.99) & 0.98 (0.97, 0.99) & 0.97 (0.95, 0.98) & 3.88 (3.74, 4.07) \\ 
spTGCCA3 & 0.95 (0.94, 0.96) & 0.95 (0.94, 0.96) & 0.96 (0.95, 0.97) & 0.97 (0.95, 0.97) & 0.97 (0.95, 0.98) & 2.37 (2.23, 2.61) \\ 
RGCCA & 0.86 (0.82, 0.89) & 0.80 (0.76, 0.84) & 0.82 (0.76, 0.85) & 0.94 (0.92, 0.95) & 0.97 (0.96, 0.98) & 1.13 (1.01, 1.22) \\ 
SVD & 0.95 (0.86, 0.97) & 0.95 (0.81, 0.96) & 0.95 (0.73, 0.97) & 0.96 (0.66, 0.98) & 0.98 (0.97, 0.99) & 0.45 (0.44, 0.49) \\ 
\bottomrule 
\end{tabular}

  }
\end{table}
\begin{table}
  \caption{Cosine between the true and the estimated canonical vectors for different models on blocks "Square", "Gas", "Cross", "Cross (small)" and "Vector", for levels of SNR -20dB, -10dB, -6dB, and 0dB from top to bottom, and computation times, for n = 100 and 100 folds. Median and quantiles (2.5\% and 97.5\%) are reported.}
    \label{tab:5B_100_all}
  \centering
  \resizebox{\columnwidth}{!}{%
  \begin{tabular}{lllllll} 
\toprule 
Model & Square & Gas & Cross & Cross (small) & Vector & Computation time \\ 
\midrule 
TGCCA1 & 0.51 (0.00, 0.92) & 0.53 (0.00, 0.84) & 0.34 (0.01, 0.81) & 0.40 (0.02, 0.81) & 0.44 (0.01, 0.80) & 2.38 (1.78, 4.78) \\ 
MGCCA & 0.78 (0.00, 0.92) & 0.66 (0.00, 0.83) & 0.70 (0.01, 0.81) & 0.65 (0.00, 0.81) & 0.63 (0.01, 0.80) & 1.72 (1.58, 2.05) \\ 
TGCCA3 & 0.27 (0.01, 0.73) & 0.27 (0.01, 0.72) & 0.24 (0.00, 0.74) & 0.24 (0.01, 0.77) & 0.29 (0.04, 0.74) & 4.58 (3.18, 7.75) \\ 
spTGCCA3 & 0.24 (0.00, 0.71) & 0.20 (0.00, 0.69) & 0.25 (0.00, 0.70) & 0.22 (0.00, 0.74) & 0.30 (0.03, 0.74) & 2.01 (1.70, 3.17) \\ 
RGCCA & 0.25 (0.03, 0.51) & 0.24 (0.02, 0.44) & 0.25 (0.01, 0.49) & 0.33 (0.02, 0.62) & 0.32 (0.03, 0.62) & 1.38 (0.74, 3.78) \\ 
SVD & 0.03 (0.00, 0.09) & 0.03 (0.00, 0.08) & 0.03 (0.00, 0.09) & 0.03 (0.00, 0.08) & 0.06 (0.00, 0.18) & 0.10 (0.09, 0.14) \\ 
\bottomrule 
\end{tabular}

  }
  \resizebox{\columnwidth}{!}{%
  \begin{tabular}{lllllll} 
\toprule 
Model & Square & Gas & Cross & Cross (small) & Vector & Computation time \\ 
\midrule 
TGCCA1 & 0.97 (0.89, 0.98) & 0.87 (0.81, 0.89) & 0.85 (0.75, 0.87) & 0.84 (0.76, 0.86) & 0.85 (0.75, 0.91) & 1.39 (1.25, 1.66) \\ 
MGCCA & 0.95 (0.92, 0.97) & 0.86 (0.80, 0.88) & 0.84 (0.79, 0.87) & 0.83 (0.76, 0.85) & 0.85 (0.75, 0.92) & 1.48 (1.38, 1.64) \\ 
TGCCA3 & 0.88 (0.79, 0.92) & 0.86 (0.76, 0.91) & 0.89 (0.78, 0.93) & 0.89 (0.78, 0.94) & 0.85 (0.74, 0.91) & 2.79 (2.62, 3.13) \\ 
spTGCCA3 & 0.82 (0.74, 0.86) & 0.82 (0.73, 0.87) & 0.85 (0.79, 0.89) & 0.86 (0.78, 0.91) & 0.85 (0.74, 0.91) & 1.57 (1.48, 1.71) \\ 
RGCCA & 0.61 (0.49, 0.70) & 0.56 (0.45, 0.63) & 0.56 (0.46, 0.65) & 0.76 (0.64, 0.83) & 0.85 (0.72, 0.90) & 0.58 (0.46, 0.89) \\ 
SVD & 0.07 (0.00, 0.22) & 0.06 (0.00, 0.19) & 0.07 (0.00, 0.26) & 0.07 (0.01, 0.24) & 0.23 (0.01, 0.79) & 0.10 (0.09, 0.15) \\ 
\bottomrule 
\end{tabular}

  }
  \resizebox{\columnwidth}{!}{%
  \begin{tabular}{lllllll} 
\toprule 
Model & Square & Gas & Cross & Cross (small) & Vector & Computation time \\ 
\midrule 
TGCCA1 & 0.98 (0.93, 0.99) & 0.89 (0.85, 0.90) & 0.87 (0.83, 0.88) & 0.86 (0.83, 0.87) & 0.88 (0.81, 0.93) & 1.29 (1.19, 1.43) \\ 
MGCCA & 0.97 (0.93, 0.98) & 0.87 (0.84, 0.89) & 0.85 (0.83, 0.87) & 0.85 (0.81, 0.86) & 0.88 (0.81, 0.93) & 1.48 (1.38, 1.61) \\ 
TGCCA3 & 0.93 (0.87, 0.95) & 0.91 (0.86, 0.94) & 0.93 (0.88, 0.96) & 0.94 (0.89, 0.96) & 0.89 (0.81, 0.93) & 2.85 (2.71, 3.04) \\ 
spTGCCA3 & 0.84 (0.80, 0.89) & 0.87 (0.80, 0.90) & 0.88 (0.84, 0.91) & 0.89 (0.85, 0.92) & 0.88 (0.81, 0.93) & 1.52 (1.43, 1.63) \\ 
RGCCA & 0.63 (0.53, 0.71) & 0.57 (0.47, 0.62) & 0.57 (0.49, 0.67) & 0.80 (0.73, 0.87) & 0.89 (0.82, 0.94) & 0.50 (0.41, 0.63) \\ 
SVD & 0.18 (0.01, 0.74) & 0.17 (0.01, 0.67) & 0.14 (0.01, 0.71) & 0.19 (0.01, 0.87) & 0.86 (0.46, 0.94) & 0.11 (0.10, 0.15) \\ 
\bottomrule 
\end{tabular}

  }
  \resizebox{\columnwidth}{!}{%
  \begin{tabular}{lllllll} 
\toprule 
Model & Square & Gas & Cross & Cross (small) & Vector & Computation time \\ 
\midrule 
TGCCA1 & 0.99 (0.97, 0.99) & 0.90 (0.87, 0.90) & 0.88 (0.86, 0.89) & 0.87 (0.84, 0.88) & 0.90 (0.83, 0.94) & 1.22 (1.15, 1.33) \\ 
MGCCA & 0.98 (0.96, 0.99) & 0.89 (0.86, 0.90) & 0.87 (0.84, 0.88) & 0.86 (0.83, 0.87) & 0.90 (0.83, 0.94) & 1.46 (1.37, 1.65) \\ 
TGCCA3 & 0.96 (0.93, 0.98) & 0.95 (0.92, 0.97) & 0.97 (0.94, 0.98) & 0.97 (0.94, 0.98) & 0.91 (0.83, 0.94) & 2.77 (2.61, 2.94) \\ 
spTGCCA3 & 0.86 (0.81, 0.88) & 0.88 (0.85, 0.90) & 0.90 (0.88, 0.92) & 0.91 (0.88, 0.93) & 0.91 (0.83, 0.94) & 1.44 (1.37, 1.59) \\ 
RGCCA & 0.60 (0.52, 0.69) & 0.52 (0.46, 0.62) & 0.56 (0.47, 0.62) & 0.80 (0.73, 0.86) & 0.93 (0.87, 0.95) & 0.42 (0.36, 0.54) \\ 
SVD & 0.92 (0.47, 0.94) & 0.90 (0.39, 0.93) & 0.90 (0.46, 0.94) & 0.94 (0.53, 0.97) & 0.96 (0.93, 0.98) & 0.11 (0.10, 0.14) \\ 
\bottomrule 
\end{tabular}

  }
\end{table}
\begin{table}
  \caption{Cosine between the true and the estimated canonical vectors for different models on blocks "Gas" and "Cross (small)", for levels of SNR -20dB, -10dB, -6dB and 0dB from top to bottom, and computation times, for n = 1000 and 10 folds. Median and quantiles (2.5\% and 97.5\%) are reported. Models that saw only 2 blocks (suffixed with "b2") are compare with models that saw the 5 blocks (suffixed with "b5").}
    \label{tab:2B_vs_5B_1000_all}
  \centering
  \resizebox{0.6\columnwidth}{!}{%
  %

  }
\end{table}
\begin{table}
  \caption{Cosine between the true and the estimated canonical vectors for different models on blocks "Gas" and "Cross (small)", for levels of SNR -20dB, -10dB, -6dB and 0dB from top to bottom, and computation times, for n = 500 and 20 folds. Median and quantiles (2.5\% and 97.5\%) are reported. Models that saw only 2 blocks (suffixed with "b2") are compare with models that saw the 5 blocks (suffixed with "b5").}
    \label{tab:2B_vs_5B_500_all}
  \centering
  \resizebox{0.6\columnwidth}{!}{%
  %

  }
\end{table}
\begin{table}
  \caption{Cosine between the true and the estimated canonical vectors for different models on blocks "Gas" and "Cross (small)", for levels of SNR -20dB, -10dB, -6dB and 0dB from top to bottom, and computation times, for n = 300 and 33 folds. Median and quantiles (2.5\% and 97.5\%) are reported. Models that saw only 2 blocks (suffixed with "b2") are compare with models that saw the 5 blocks (suffixed with "b5").}
    \label{tab:2B_vs_5B_300_all}
  \centering
  \resizebox{0.6\columnwidth}{!}{%
  %

  }
\end{table}

\begin{table}
  \caption{Cosine between the true and the estimated canonical vectors for different models on blocks "Cross" and "Cross (small) 3D", for levels of SNR -20dB, -10dB, -6dB and 0dB from top to bottom, and computation times, for n = 1000 and 10 folds. Median and quantiles (2.5\% and 97.5\%) are reported (3D settings).}
    \label{tab:4D-2B_1000_all}
  \centering
  \resizebox{0.6\columnwidth}{!}{%
  %

  }
\end{table}
\begin{table}
  \caption{Cosine between the true and the estimated canonical vectors for different models on blocks "Cross" and "Cross (small) 3D", for levels of SNR -20dB, -10dB, -6dB and 0dB from top to bottom, and computation times, for n = 500 and 20 folds. Median and quantiles (2.5\% and 97.5\%) are reported (3D settings).}
    \label{tab:4D-2B_500_all}
  \centering
  \resizebox{0.6\columnwidth}{!}{%
  %

  }
\end{table}
\begin{table}
  \caption{Cosine between the true and the estimated canonical vectors for different models on blocks "Cross" and "Cross (small) 3D", for levels of SNR -20dB, -10dB, -6dB and 0dB from top to bottom, and computation times, for n = 100 and 100 folds. Median and quantiles (2.5\% and 97.5\%) are reported (3D settings).}
    \label{tab:4D-2B_100_all}
  \centering
  \resizebox{0.6\columnwidth}{!}{%
  %

  }
\end{table}
\begin{table}
  \caption{Cosine between the true and the estimated canonical vectors for different models on blocks "Square", "Gas 3D", "Cross", "Cross (small) 3D" and "Vector", for levels of SNR -20dB, -10dB, -6dB, and 0dB from top to bottom, and computation times, for n = 1000 and 10 folds. Median and quantiles (2.5\% and 97.5\%) are reported (3D settings).}
    \label{tab:4D-5B_1000_all}
  \centering
  \resizebox{\columnwidth}{!}{%
  %

  }
\end{table}
\begin{table}
  \caption{Cosine between the true and the estimated canonical vectors for different models on blocks "Square", "Gas 3D", "Cross", "Cross (small) 3D" and "Vector", for levels of SNR -20dB, -10dB, -6dB, and 0dB from top to bottom, and computation times, for n = 500 and 20 folds. Median and quantiles (2.5\% and 97.5\%) are reported (3D settings).}
    \label{tab:4D-5B_500_all}
  \centering
  \resizebox{\columnwidth}{!}{%
  %

  }
\end{table}
\begin{table}
  \caption{Cosine between the true and the estimated canonical vectors for different models on blocks "Square", "Gas 3D", "Cross", "Cross (small) 3D" and "Vector", for levels of SNR -20dB, -10dB, -6dB, and 0dB from top to bottom, and computation times, for n = 100 and 100 folds. Median and quantiles (2.5\% and 97.5\%) are reported (3D settings).}
    \label{tab:4D-5B_100_all}
  \centering
  \resizebox{\columnwidth}{!}{%
  %

  }
\end{table}

\end{document}